\documentclass[runningheads]{llncs}

\usepackage{eccv}

\usepackage{eccvabbrv}

\usepackage{graphicx}
\usepackage{booktabs}
\usepackage{times}
\usepackage{soul}
\usepackage{url}
\usepackage[utf8]{inputenc}
\usepackage{amssymb}
\usepackage{booktabs}
\usepackage[accsupp]{axessibility} 
\usepackage{multirow}
\usepackage{mathrsfs}
\usepackage{multirow} 
\usepackage{colortbl} 
\usepackage{xcolor} 
\usepackage{caption}
\usepackage{xspace}

\usepackage[ruled, vlined, linesnumbered]{algorithm2e}
\usepackage{algpseudocode}
\usepackage{blindtext}
\usepackage{titletoc}

\newcommand{\topsequence}[4]{\{ #1^{#2} \}_{#2=#3}^{#4}}
\newcommand{\botsequence}[4]{\{ #1_{#2} \}_{#2=#3}^{#4}}
\newcommand{\objpair}{$\left(o_{i}^{t}, o_{j}^{t} \right) $\xspace}

\newcommand{\method}{SceneSayer\xspace}
\newcommand{\orpu}{object representation processing unit\xspace}
\newcommand{\scpu}{spatial context processing unit\xspace}
\newcommand{\ldpu}{latent dynamics processing unit\xspace}

\newcommand{\ORPU}{Object Representation Processing Unit\xspace}
\newcommand{\SCPU}{Spatial Context Processing Unit\xspace}
\newcommand{\LDPU}{Latent Dynamics Processing Unit\xspace}
\newcommand{\RNum}[1]{\uppercase\expandafter{\romannumeral #1\relax}}

\definecolor{highlightColor}{RGB}{194, 240, 151}

\usepackage[accsupp]{axessibility}  

\usepackage[pagebackref,breaklinks,colorlinks,citecolor=eccvblue]{hyperref}
\usepackage{hyperref}

\usepackage{orcidlink}

\begin{document}

\title{Towards Scene Graph Anticipation}

\titlerunning{SceneSayer for Scene Graph Anticipation}


\author{Rohith Peddi\inst{1}\orcidlink{0009-0007-4705-8129} \and
Saksham Singh\inst{2}\orcidlink{0009-0009-5320-393X}$^*$ \and
Saurabh \inst{2}\orcidlink{0009-0002-4763-8872}$^*$ \and
Parag Singla\inst{2}\orcidlink{0009-0000-9190-9794} \and
Vibhav Gogate\inst{1}\orcidlink{0000-0002-6459-7358}}

\authorrunning{R.Peddi et al.}

\institute{The University of Texas at Dallas \and
    Indian Institute of Technology Delhi
}

\maketitle

\begin{abstract}
Spatio-temporal scene graphs represent interactions in a video by decomposing scenes into individual objects and their pair-wise temporal relationships. Long-term anticipation of the fine-grained pair-wise relationships between objects is a challenging problem. To this end, we introduce the task of Scene Graph Anticipation (SGA). We adapt state-of-the-art scene graph generation methods as baselines to anticipate future pair-wise relationships between objects and propose a novel approach SceneSayer. In SceneSayer, we leverage object-centric representations of relationships to reason about the observed video frames and model the evolution of relationships between objects. We take a continuous time perspective and model the latent dynamics of the evolution of object interactions using concepts of NeuralODE and NeuralSDE, respectively. We infer representations of future relationships by solving an Ordinary Differential Equation and a Stochastic Differential Equation, respectively. Extensive experimentation on the Action Genome dataset validates the efficacy of the proposed methods.\footnotetext{$^*$ denotes equal contribution with names in alphabetical order.} 

\end{abstract}

\section{Introduction} \label{sec:intro}

We focus on spatio-temporal scene graphs \cite{Ji_2019}, which is a widely used framework for representing the evolving spatial and temporal relationships among objects. These graphs contain information about the objects present in a video, including their categories, positions, sizes, and spatial dependencies. Simultaneously, they illustrate how these relationships evolve over time, revealing objects' movement, interactions, and configuration changes across consecutive frames in a video sequence. They facilitate our understanding of dynamic scenes and serve as a valuable tool for addressing downstream tasks in applications such as action recognition and video analysis, where the temporal dynamics of object interactions play a crucial role.

\begin{figure*}[!t]
    \centering
    \includegraphics[width=\textwidth]{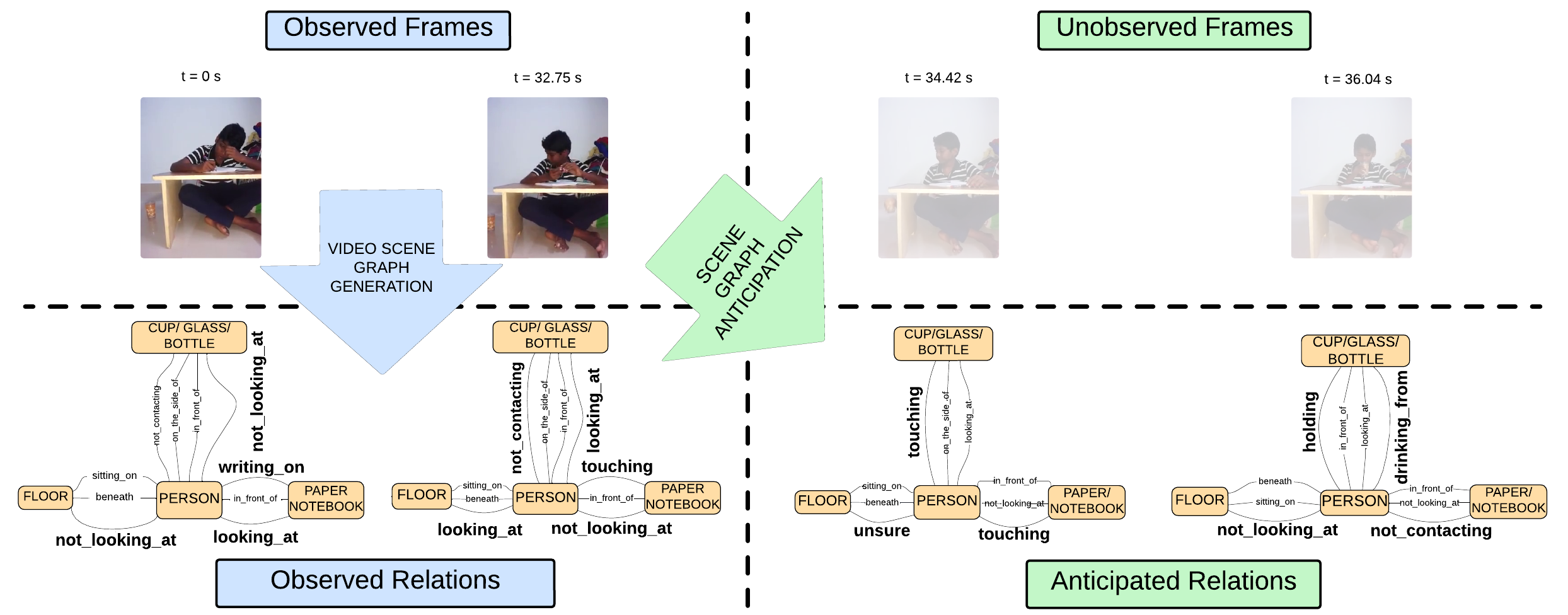}
    \caption[Caption for LOF]{\textbf{Task Description.} We contrast the task of Video Scene Graph Generation (VidSGG) on the left with the proposed task of Scene Graph Anticipation (SGA) on the right. VidSGG entails the identification of relationships from the observed data, such as (\textit{Person}, \textbf{looking\_at}, \textit{Floor}) and (\textit{Person}, \textbf{not\_contacting}, \textit{Cup}). SGA aims to anticipate the evolution of these relationships to   (\textit{Person}, \textbf{touching}, \textit{Cup}), and eventually, (\textit{Person}, \textbf{drinking\_from}, \textit{Cup}).\protect\footnotemark}
    \label{fig:task_picture}
\end{figure*}

We introduce a novel task known as \textit{Scene Graph Anticipation (SGA)}, which, given a video stream, aims to forecast future interactions between objects, as shown in Figure \ref{fig:task_picture}. The Scene Graph Anticipation (SGA) task holds significance across diverse domains due to its potential applications and relevance to several downstream tasks. For instance, it contributes to \textit{enhanced video understanding} by predicting spatiotemporal relationships within video scenes, facilitating improved video analysis and interpreting complex object interactions over time. SGA plays a crucial role in \textit{activity recognition}, enabling systems to predict future object interactions for more accurate classification and advanced surveillance. Anticipation aids in \textit{anomaly detection} by identifying deviations from expected object relationships, thereby enhancing the detection of abnormal events in video sequences. \textit{Intelligent surveillance systems} can also benefit from SGA, allowing systems to predict and respond to security threats by understanding evolving object relationships in monitored environments. Finally, SGA is essential for predicting object movements and interactions for applications in \textit{robotics and autonomous systems}, contributing to safer and more efficient navigation and decision-making processes.

To tackle the challenges of SGA\footnotetext{The relationships presented are taken from the ground truth annotations of the Action Genome}, we introduce two novel approaches that extend a state-of-the-art scene graph generation method \cite{cong_et_al_sttran_2021}. These approaches utilize object-centric representations of relationships, allowing us to analyze observed video frames and effectively model the dynamic evolution of interactions between objects. This detailed understanding of temporal dynamics is the basis for our proposed methods.

In a departure from traditional sequential modelling, our two approaches adopt a continuous-time perspective. Drawing inspiration from Neural Ordinary Differential Equations (NeuralODE) \cite{Chen_2018} and Neural Stochastic Differential Equations (NeuralSDE) \cite{kidger2021efficient}, respectively, we develop methods that capture the latent dynamics governing the evolution of object interactions. By formulating the anticipation problem as solving Ordinary Differential Equations (ODE) and Stochastic Differential Equations (SDE) and thus using a continuous representation of the anticipated relationships, we hope to significantly expand the fidelity of our predictions. We rigorously validated our proposed methods and strong generation-based baselines on the Action Genome dataset \cite{Ji_2019}, a benchmark for spatio-temporal scene understanding. Our experimental results demonstrate the superior performance of our approaches in accurately anticipating fine-grained pair-wise relationships between objects.


\section{Related Work} 




\textbf{Spatial Graphs.} Learning to represent visual content in static data such as 2D/3D images as a spatial graph where the objects act as nodes and edges to describe the visual relation between the objects is called \textit{Image Scene Graph Generation} (ImgSGG). There has been extensive research in 2D ImgSGG direction following the seminal work of Visual Genome  \cite{krishna_et_al_visual_2017}. Following this \cite{kim_et_al_3dsgg_2019} extended the task to static 3D (RGB \& depth) scene data. Recently, there has been a surge in methods that explored the role of foundation models in ImgSGG task-variants such as open-vocabulary ImgSGG \cite{chen_et_al_expanding_scene_graph_2023}, weakly supervised ImgSGG \cite{kim_et_al_llm4sgg_2023}, panoptic ImgSGG \cite{zhou_et_al_vlprompt_2023} and zero-shot ImgSGG \cite{zhao_et_al_less_2023,li2023zeroshotvisualrelationdetection}. 

\textbf{Spatio-Temporal Graphs.} Dynamic visual content, such as videos, provides a more natural set of contextual features describing dynamic interaction between objects. Encoding such content into a structured spatiotemporal graphical representation for frames where nodes describe objects and edges describe the temporal relations is called Video Scene Graph Generation (VidSGG). Early approaches on VidSGG extended previously proposed ImSGG-based methods to the temporal domain. We refer to \cite{zhu_et_al_sgg_2022} for a comprehensive survey on earlier work. Recent work explored learning better representations using architectures like Transformers \cite{cong_et_al_sttran_2021,shit_et_al_relformer_2022} and unbiased representations \cite{nag_et_al_tempura_2023,khandelwal_correlation_2023} owing to the long-tailed datasets Action Genome \cite{Ji_2019}, VidVRD \cite{xindi_et_al_vid_vrd_2017}. 


\textbf{Applications of Structured Representations.} Structured representation of visual content has been used in various downstream tasks, including task planning \cite{agia_et_al_taskography_2022}, image manipulation \cite{dhamo_et_al_image_2020}, visual question answering\cite{cherian_et_al_vqa_2022}, video synthesis using action graphs \cite{bar_et_al_action_2021} and using scene graph as a knowledge base \cite{kurenkov_et_al_sgm_2023}. To the best of our knowledge, we are the first to formally introduce the SGA task and propose baseline methodologies\footnote{SGA is related to \cite{Wu_Zhao_Wang_2021}, wherein they use relation forecasting as an intermediate step for action prediction. However, we aim for precise anticipation of future relationships between interacting objects. SGA is also closely related to \cite{Mi_2021}, which constructed a dataset from Action Genome by sampling frames and truncating the videos. However, we differentiate our method by utilizing the complete dataset to train our models \textbf{without any pre-processing}. Furthermore, we acknowledge that the approach in \cite{Mi_2021} resembles the proposed baseline (Variant-1).}.



\textbf{Video Prediction.} Early video prediction methods treat dynamics modelling as a sequential prediction problem in pixel space using image-level features \cite{lee_et_al_stochastic_2018}. Later approaches proposed include using external knowledge as priors \cite{walker2016uncertain,finn2016unsupervised,lu_et_al_mm_prior_kdd}, better architectural design to model contextual information \cite{yu2022modular,qi2021learning}, and focusing on object-centric representations \cite{wu2023slotformer}. Recently, there has been a surge in methods proposed that use diffusion models to estimate the distribution of a short future video clip \cite{höppe2022diffusion,voleti2022mcvd}. Contrary to the conventional dense pixel-based generation often seen in video prediction techniques, SGA emphasizes the importance of learning relationship representations that facilitate the prediction of the evolution of observed interactions between objects. Our experiments demonstrate that the proposed method can effectively estimate these relationship representations for periods extending beyond 30 seconds into the future.

\textbf{Neural Differential Equations.} Several previous works leveraged the frameworks of NeuralODE \cite{Chen_2018} and NeuralSDE \cite{kidger2021efficient} to explore the learning of latent dynamics models for various tasks of interest such as trajectory prediction \cite{liang_et_al_node_2021}, traffic forecasting \cite{liu2023graphbased}, video generation \cite{park2021vidode}, and simulation of multi-agent dynamical systems \cite{huang2020learning,poli2021graph,zijie_graph_ode_2021}.

\section{Background}


\noindent \textbf{Ordinary Differential Equations (ODEs)} The initial value problem (IVP) is given by:
\begin{equation}
    \frac{\mathbf{d}\mathbf{z}(t)}{\mathbf{d}t} = \mathbf{f}( \mathbf{z}(t), t), \quad \mathbf{z}(t_0) = z_0
\end{equation}
Here, \(\mathbf{f}: \mathbb{R}^d \times \mathbb{R} \rightarrow \mathbb{R}^d\) represents a time-varying smooth vector field, and \(\mathbf{z}(t)\) is the solution to the IVP. Chen et al.~\cite{Chen_2018} introduced the framework of NeuralODEs wherein they relaxed the time-variance of the vector field $\mathbf{f}$ and 
parameterized it through neural networks, thus enabling efficient learning of dynamical systems from data. This approach has paved the way for modelling the dynamics of latent states through LatentODEs. Analytical solutions for complex ODEs are typically infeasible; hence, we resort to numerical methods that discretize the time domain into finite intervals, approximating the solution at each step. Thus, they trade off precision with computation time. On the faster, less accurate side, we have single-step methods, such as Euler, while on the slower, more accurate side, we have multistep methods, such as Adams-Bashforth.

\noindent \textbf{Stochastic Differential Equations (SDEs)} An initial value problem is formulated as:
\begin{equation}
    \mathbf{d}\mathbf{z}(t) = \mu(\mathbf{z}(t), t) \mathbf{d}t + \sigma(\mathbf{z}(t), t) \mathbf{d}\mathbf{W}(t), \quad \mathbf{z}(t_0) = z_0
\end{equation}

\noindent Here, $\mu(\mathbf{z}(t), t): \mathbb{R}^d \times \mathbb{R} \rightarrow \mathbb{R}^d$, $\sigma(\mathbf{z}(t), t): \mathbb{R}^d \times \mathbb{R} \rightarrow \mathbb{R}^{(d \times m)}$ represent drift and diffusion terms, respectively and $\mathbf{W}(t)$ denotes an $m$-dimensional Wiener process. NeuralSDEs employ neural networks to parameterize both drift and diffusion terms thus enabling the learning of stochastic processes from data \cite{Li_2021,kidger2021efficient}. Solving SDEs analytically is often challenging and infeasible. Numerical solutions depend on the choice of interpretation of the SDE, which is often connected to the conceptual model underlying the equation. There are many interpretations of an SDE, of which Stratonovich is one of the most popular \cite{oksendal03}. Stratonovich's interpretation of an SDE is commonly used to model physical systems subjected to noise. SDE solvers use discretization methods to approximate the continuous dynamics of the system over small time intervals. The most common numerical methods for solving SDEs include Euler-Maruyama, Milstein, and Runge-Kutta schemes, each varying in accuracy and computational complexity.


\section{Notation \& Problem Description}

\subsubsection{Notation.}  Given an input video segment $V_{1}^{T}$, we represent it using a set of frames $V_{1}^{T} = \topsequence{I}{t}{1}{T}$ defined on discretized time steps $t = \{1, 2, \cdots, T\}$, where the total number of observed frames $T$ varies across video segments. A scene graph is a symbolic representation of the objects present in the frame and their pair-wise relationships. In each frame $I^t$, we represent the set of objects observed in it using $O^t = \botsequence{o^t}{k}{1}{N(t)}$, where $N(t)$ denotes the total number of objects observed in $I^t$. Let $\mathcal{C}$ be the set comprising all object categories, then each instance of an object $o_{k}^{t}$ is defined by its bounding box information $\mathbf{b}^t_{k}$ and object category $\mathbf{c}^t_{k}$, where $\mathbf{b}^t_{k} \in [0,1]^{4}$ and $\mathbf{c}^t_{k} \in \mathcal{C}$. Let $\mathcal{P}$ be the set comprising all predicate classes that spatio-temporally describe pair-wise relationships between two objects; then each pair of objects $\left(o_{i}^{t}, o_{j}^{t} \right)$ may exhibit multiple relationships, defined through predicates $\{p^{t}_{ijk}\}_{k}$ where $p^{t}_{ijk} \in \mathcal{P}$. We define a relationship instance $r_{ijk}^{t}$ as a triplet $\left(o_{i}^{t}, p^{t}_{ijk}, o_{j}^{t} \right)$ that combines two distinct objects $\left(o_{i}^{t}, o_{j}^{t} \right)$ and a predicate $p^{t}_{ijk}$. Thus, the scene graph $\mathcal{G}^{t}$ is the set of all relationship triplets $\mathcal{G}^{t} = \{r_{ijk}^{t}\}_{ijk}$. Additionally, for each observed object $o_{i}^{t}$ and a pair of objects $\left(o_{i}^{t}, o_{j}^{t} \right)$, we use $\hat{\mathbf{c}}_{i}^{t} \in \left[ 0,1 \right]^{|\mathcal{C}|}$ and $\hat{\mathbf{p}}_{ij}^{t} \in \left[ 0,1 \right]^{|\mathcal{P}|}$ to represent the distributions over object categories and predicate classes respectively. Here $\sum_{k} \hat{{c}}_{ik}^{t} = 1$, $\sum_k \hat{{p}}_{ijk}^{t} = 1$.

\subsubsection{Problem Description.} We formally define and contrast the tasks of Video Scene Graph Generation (VidSGG) and Scene Graph Anticipation (SGA) as follows: 

\begin{enumerate}
    \item The goal of \textbf{VidSGG} is to build scene graphs $\topsequence{\mathcal{G}}{t}{1}{T}$ for the observed video segment $V_{1}^{T} = \topsequence{I}{t}{1}{T}$. It entails the detection of objects $\botsequence{o^t}{k}{0}{N(t)}$ in each frame and the prediction of all pair-wise relationships $\{r_{ijk}^{t}\}_{ijk}$ between detected objects. 
    \item The goal of \textbf{SGA} is to build scene graphs $\topsequence{\mathcal{G}}{t}{T+1}{T+H}$ for future frames $V_{T+1}^{T+H} = \topsequence{I}{t}{T+1}{T+H}$ of the video based on the observed segment of the video $V_{1}^{T}$, here $H$ represents the anticipation horizon. Thus, it entails anticipating objects and their pair-wise relationships in future scenes. We note that anticipating the emergence of new objects in future frames is a significantly harder problem. So, we presuppose the continuity of observed objects in future frames (i,e $ \botsequence{o^t}{i}{1}{N(t)} = \botsequence{o^T}{i}{1}{N(T)}, \forall t > T$) and predict the evolution of relationships $\Bigl\{\topsequence{r_{ijk}}{t}{T+1}{T+H}\Bigr\}_{ijk}$ in future scenes.
\end{enumerate}

\noindent \textbf{Graph Building Strategies.} To build scene graphs for future frames, we employ two strategies that are widely established within the VidSGG literature: 
\begin{itemize}
    \item[--] \textbf{With Constraint Strategy}: This approach enforces a unique interaction constraint between any pair of objects in a scene. Specifically, for a pair of objects $\left(o_{i}^{t}, o_{j}^{t} \right)$, there exists a single relationship predicate $p^t_{ij}$ that describes the interaction between the objects, and we incorporate relationship triplets $\{r_{ij}^t\}_{ij}$ into the scene graph $\mathcal{G}^t$.
    \item[--] \textbf{No Constraint Strategy}: In contrast, this approach embraces a more complex and detailed graph structure by permitting multiple edges (where each edge describes a relationship predicate) between any pair of interacting objects $\left(o_{i}^{t}, o_{j}^{t} \right)$. Specifically, we incorporate all predicted relationship triplets $\{r_{ijk}^{t}\}_{ijk}$ into the scene graph $\mathcal{G}^t$.
\end{itemize}

\section{Technical Approach}

\begin{figure}[!t]
    \centering
    \includegraphics[width=0.92\textwidth]{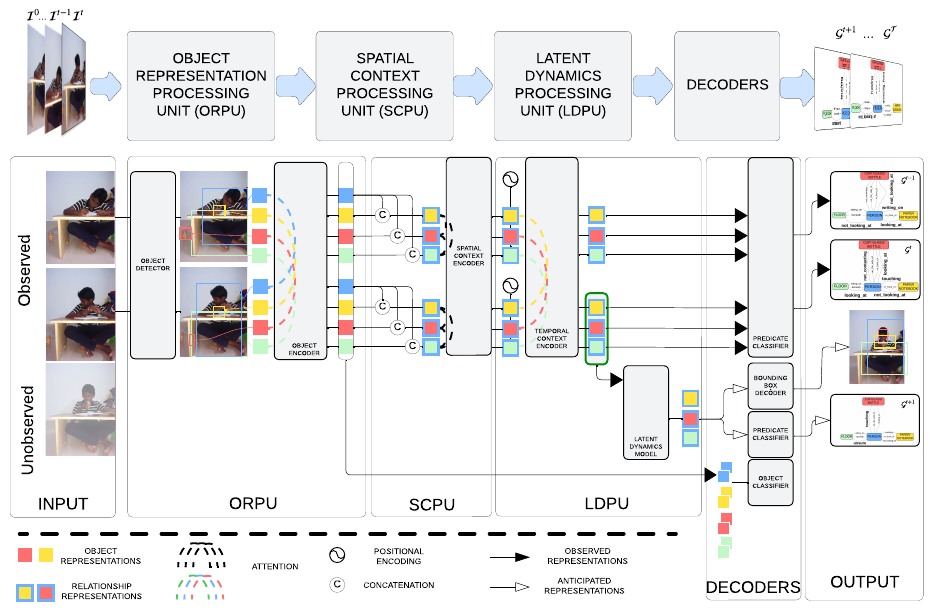}
    \caption{\textbf{Overview of \method.} The forward pass of SceneSayer begins with ORPU, where initial object proposals are generated for each frame. These proposals are then fed to a temporal encoder to ensure that the object representations remain consistent over time. Next, in SCPU, we construct initial relationship representations by concatenating the representations of interacting objects. These representations are further refined using a spatial encoder, embedding the scene's spatial context into these relationship representations. Then, the representations undergo further enhancement in LDPU, where another temporal encoder fine-tunes them, imbuing the data with comprehensive spatio-temporal scene knowledge. These refined relationship representations from the final observed frame are then input to a Latent Dynamics Model (LDM), where a generative model, either a NeuralODE or a NeuralSDE, generates relationship representations of interacting objects in future frames by solving the corresponding differential equations. Finally, these future representations are decoded into relationship predicates to construct anticipated scene graphs.  }
    \label{fig:technical_approach}
\end{figure}

We propose SceneSayer for Scene Graph Anticipation. As illustrated in Fig \ref{fig:technical_approach}, SceneSayer incorporates an \orpu (ORPU) that captures representations of the objects detected in a frame, a \scpu (SCPU)\footnote{Both ORPU and SCPU can be adapted from any existing VidSGG models.} that builds spatial context-aware relationship representations and a \ldpu (LDPU), which is designed to understand spatio-temporal dynamics and predict the evolution of relationships between interacting objects. Distinctively, SceneSayer employs a \textit{continuous-time framework} to model the latent dynamics of the evolution of relationships. In the following sections, we provide a detailed description of these units and the methodologies employed for training and testing SceneSayer.

\subsection{\ORPU}

To achieve temporally consistent object representations across observed frames, our approach first extracts visual features using a pre-trained object detector and then further enhances these extracted features by passing them through an encoder. Specifically, We utilize a pre-trained Faster R-CNN\cite{ren2016faster} for extracting visual features ($\botsequence{\mathbf{v}^t}{i}{1}{N(t)}$), bounding boxes ($\botsequence{\mathbf{b}^t}{i}{1}{N(t)}$), and object category distributions ($\botsequence{\hat{\mathbf{c}}^t}{i}{1}{N(t)}$), for object proposals $\botsequence{o^t}{i}{1}{N(t)}$ in the observed frames. We then compute the matrix $\mathbf{V}_{i}$ by stacking visual features $\topsequence{\mathbf{v}_i}{t}{1}{T}$. We employ a transformer encoder \cite{vaswani_et_al_2017} to aggregate temporal information across extracted visual features from all frames. The input for our encoder consists of sequences $\mathbf{V}_{i}$, which serve simultaneously as queries $\mathbb{Q}$, keys $\mathbb{K}$ and values $\mathbb{V}$. Formally, the transformation process at the $n$-th encoder layer is given by




\begin{equation}
    \mathbf{V}_{i}^{(n)} = \operatorname{ObjectEncoder}\left(\mathbb{Q} = \mathbb{K} = \mathbb{V} =  \mathbf{V}_{i}^{(n-1)} \right) 
\end{equation}



\subsection{\SCPU}

To learn spatial context-aware relationship representations. We first construct relationship representations between interacting objects in a scene and further process them by passing through an encoder. Specifically, following \cite{cong_et_al_sttran_2021}, we construct relationship representation $\mathbf{z}_{ij}^t$ by concatenating visual and semantic features of the objects as follows: 
\begin{equation}
    \mathbf{z}_{ij}^{t} = \operatorname{Concat} (\mathbf{W}_{1} \mathbf{v}^{t}_{i}, \mathbf{W}_{2} \mathbf{v}^{t}_{j}, \mathbf{W}_{3}\mathbf{U}_{ij}^{t}, \mathbf{S}_{i}^{t}, \mathbf{S}_{j}^{t})
\end{equation}

Here, $\mathbf{W}_{1}, \mathbf{W}_{2}, \mathbf{W}_{3}$ represent the learnable weights of linear layers. $\mathbf{U}_{ij}^{t}$ represent processed feature maps of the union box computed by RoIAlign \cite{ren2016faster}. $\mathbf{S}_{i}^{t}$ and $\mathbf{S}_{j}^{t}$ correspond to the semantic embedding vectors of the object categories. Subsequently, we employ a transformer encoder to integrate spatial contextual information. Here, for each observed frame $I^{t}$, we construct $\mathbf{Z}^{t}$ by stacking all relationship features $\{\mathbf{z}_{ij}^{t}\}_{ij}$ corresponding to objects observed in a frame. We then feed it as input for a transformer encoder which operates on queries $\mathbb{Q}$, keys $\mathbb{K}$, and values $\mathbb{V}$ that are derived from the same source, the preceding layer's output or the initial relationship features. Formally, the transformation process at the $n$-th encoder layer is given by:
\begin{equation}
    {[\mathbf{Z}^{t}}]^{(n)} = \operatorname{SpatialEncoder}\left(\mathbb{Q} = \mathbb{K} = \mathbb{V} = [{\mathbf{Z}^{t}}]^{(n-1)} \right) 
\end{equation}
 

\subsection{\LDPU}

To understand spatio-temporal dynamics of the evolution of relationships between interacting objects, departing from traditional approaches that architectural variants of transformers \cite{girdhar2021anticipative, thickstun2023anticipatory}, we take a continuous time approach and learn a governing differential equation in latent space of relationship representations. Our LDPU contains two components: (1) Temporal Context Encoder and (2) Latent Dynamics Model. 


\noindent \textbf{Temporal Context Encoder.} To learn spatio-temporal context-aware relationship representations, we further process the output of SCPU by passing it through a transformer encoder that fine-tunes the representation. Specifically, we compute the matrix $\mathbf{Z}_{ij}$ by stacking the relationship representations $\topsequence{\mathbf{z}_{ij}}{t}{1}{T}$. We feed $\mathbf{Z}_{ij}$ through an encoder that aggregates temporal information across all observed relationship representations refer Eq.~\ref{eq:TemporalEncoder}. Here, the input $\mathbf{Z}_{ij}$, serves simultaneously as queries $\mathbb{Q}$, keys $\mathbb{K}$, and values $\mathbb{V}$. 


\begin{equation}
    \label{eq:TemporalEncoder}
    \mathbf{Z}_{ij}^{(n)} = \operatorname{TemporalEncoder}\left(\mathbf{Q} = \mathbf{K} = \mathbf{V} = \mathbf{Z}_{ij}^{(n)} \right)
\end{equation}



\noindent \textbf{Latent Dynamics Models.} We begin by abstracting the complexities introduced by external factors that add uncertainty to the evolution of relationships between interacting objects and construct a model based on the premise that the core dynamics driving the changes in these relationships are governed by a non-linear deterministic process. Specifically, we leverage the expressive nature of NeuralODEs \cite{Chen_2018} in learning time-invariant vector fields from data that approximate the underlying non-linear deterministic process and propose \textbf{SceneSayerODE}. We cast the problem of anticipating relationship representations of interacting objects as an instance of the ODE-IVP, where the initial condition is given by the relationship representation of the last observed interaction of the object pair \objpair. The evolution of this relationship representation over time is then mathematically described as:

\begin{equation}
    \mathbf{z}_{ij}^{T+H} = \mathbf{z}_{ij}^{T} +  \int_{T}^{T+H} \mathbf{f}_{\theta}(\mathbf{z}_{ij}^t) \, \mathbf{d}t 
    \label{eq:SceneSayerODE}
\end{equation}


In video data that is captured from a single viewpoint, we lose information due to issues such as blurry imagery and occlusions, leading to uncertain interpretations of the scenes. Therefore, it is crucial to integrate uncertainty into the modelling frameworks to represent these stochastic dynamics accurately. We assume the presence of a non-linear stochastic differential equation that governs this evolution and propose \textbf{SceneSayerSDE} which uses NeuralSDEs \cite{kidger2021efficient} to learn it from data.  Specifically, we formulate the problem of anticipating future interactions as an SDE-IVP. Here, the initial conditions are set using the representations of the last observed representation object pair, thus facilitating a data-driven learning process that accounts for the inherent uncertainties. The evolution of relationship representation over time is then described as:

\begin{equation}
    \mathbf{z}_{ij}^{T+H} = \mathbf{z}_{ij}^{T} + \int_{T}^{T+H} \mu_{\theta}(\mathbf{z}_{ij}^t) \mathbf{d}t +  \int_{T}^{T+H} \sigma_{\phi}(\mathbf{z}_{ij}^t) \mathbf{d} \mathbf{W}(t)
\end{equation}



\subsection{Decoders}

\begin{itemize}
    \item[--] \textbf{Predicate Classification Head.} We employ two predicate classification heads (two-layer multi-layer perceptron) in our approach; we use one head for classifying relationship representations between objects of the observed scenes and the other head for classifying anticipated relationship representations of the future scenes.
    \item[--] \textbf{Bounding Box Regression Head.} We employed a regression head (two-layer multi-layer perceptron) that takes the anticipated relationship representations as input and outputs bounding box information corresponding to the interacting objects\footnote{We observed that decoding bounding boxes of only actors also produced compelling results.}. 
    \item[--] \textbf{Object Classification Head.} We decode the output of the object encoder using a two-layer multi-layer perception to generate object category distributions.
\end{itemize}




\subsection{Loss Function.} 

Our training objective is constructed as a combination of loss functions implemented on two types of representations (corresponding to objects and relationships): (a) those derived from observed frames, and (b) those anticipated based on these observed frames.

\noindent \textbf{(a) Observed Representations.} We apply the following loss functions over representations derived from observed frames: (\RNum{1}) \textit{Object Classification Loss}  on the object representations and (\RNum{2}) \textit{Predicate Classification Loss} on the relationship representations.

\begin{equation*} 
    \underbrace{
        \mathcal{L}_{i} = \sum_{t=1}^{\bar{T}} \mathcal{L}_{i}^{t},
        \quad
        \mathcal{L}_{i}^{t} = -\sum_{n=1}^{|\mathcal{C}|} y_{i,n}^{t} \log(\hat{\mathbf{c}}_{i,n}^{t})
    }_\text{Object Classification Loss (\RNum{1})}; 
    \quad
    \underbrace{
        \mathcal{L}_{\text{gen}} = \sum_{t=1}^{\bar{T}} \mathcal{L}_{\text{gen}}^t, 
        \quad
        \mathcal{L}_{\text{gen}}^t = \sum_{ij} \mathcal{L}_{p^t_{ij}}
    }_\text{Predicate Classification Loss (\RNum{2})}
\end{equation*}

(\RNum{1}) \textbf{Object Classification Loss ($\mathcal{L}_{i}$).} We evaluate object classification performance by applying cross-entropy loss to the outputs from the ORPU as detailed above where $y_{i}$ represents the target for object $i$, and $\hat{\mathbf{c}}_{i}^{t}$ is the distribution over object categories.

(\RNum{2}) \textbf{Predicate Classification Loss ($\mathcal{L}_{\text{gen}}$).} focuses on classifying the relationship representations between pairs of objects \objpair across all frames (\(t \in [1, \bar{T}]\)) as detailed above. Here, $\mathcal{L}_{p^t_{ij}}$ represents multi-label margin loss and is computed as follows:

\begin{equation} \label{eq:predicate_classification_loss}
    \mathcal{L}_{p^t_{ij}}=\sum_{u \in \mathcal{P}^{+}} \sum_{v \in \mathcal{P}^{-}} \max (0,1-\hat{\mathbf{p}}^t_{ij}[v]+\hat{\mathbf{p}}^t_{ij}[u])
\end{equation}

\noindent \textbf{(b) Anticipated Representations.} We apply the following loss functions over anticipated relationship representations for each observation window: (\RNum{3}) \textit{Predicate Classification Loss}, (\RNum{4}) \textit{Bounding Box Regression Loss}, and (\RNum{5}) \textit{Reconstruction Loss}

(\RNum{3}) \textbf{Predicate Classification Loss ($\mathcal{L}_{\text{ant}}^{(1:T)}$).} focuses on classifying the anticipated relationship representations between observed pairs of objects \objpair in each window. We utilize the frames observed in each window ($t \in [1, T]$) to anticipate the relationship representations in subsequent future frames ($t \in [T+1, \min(T+H, \bar{T})]$) where $T, \bar{T}$ and $H$ denote the number of frames observed in the window, the total number of frames in the video and the length of the anticipation horizon respectively.

\begin{equation}
    \mathcal{L}_{\text{ant}}^{(1:T)} = \sum_{t=T+1}^{\min(T+H, \bar{T})} \mathcal{L}_{\text{ant}}^t, 
    \quad
    \mathcal{L}_{\text{ant}}^t = \sum_{ij} \mathcal{L}_{p^t_{ij}}
\end{equation}

(\RNum{4}) \textbf{Bounding Box Regression Loss ($\mathcal{L}_{\text{boxes}}^{(1:T)}$).} We input anticipated relationship representations into a dedicated linear\footnote{In AG, dedicated refers to the allocation of decoders: one for subject and one for objects.} layer to estimate the bounding boxes for objects. Inspired by the YOLO models, we compute the difference between the predicted and the ground truth bounding boxes using the Smoothed L1 loss as follows:

\begin{equation} \label{eq:bounding_box_regression_loss}
    \mathcal{L}_{\text{boxes}}^{(1:T)} = \sum_{t = T+1}^{\min(T+H, \bar{T})} \mathcal{L}_{\text{boxes}}^{t},
    \quad \mathcal{L}_{\text{boxes}}^{t} = \sum_{k \in \text{boxes}} \text{L}_{\text{smooth}}(b_k^t - \hat{b}_k^t)
\end{equation}

(\RNum{5}) \textbf{Reconstruction Loss ($\mathcal{L}_{\text{recon}}^{(1:T)}$).} We apply a Smoothed L1 loss to ensure that the anticipated representations mirror the output from the temporal encoder in LDPU.


\begin{equation}
    \mathcal{L}_{\text{recon}}^{(1:T)} = \sum_{t = T+1}^{\min(T+H, \bar{T})} \mathcal{L}_{\text{recon}}^{t},
    \quad
    \mathcal{L}_{\text{recon}}^t = \frac{1}{N(t) \times N(t)} \sum_{ij}^ {(N(t) \times N(t))}\text{L}_{\text{smooth}}(\mathbf{z}_{ij}^t - \hat{\mathbf{z}}_{ij}^t)
\end{equation}

Thus, the total objective for training the proposed method can be written as:

\begin{equation}
    \mathcal{L} = 
    \underbrace{
        \sum_{t=1}^{\bar{T}} \left(\lambda_{1} \mathcal{L}_{\text{gen}}^{t} + 
        \lambda_{2} \sum_{i} \mathcal{L}_{i}^{t} \right)
    }_\text{Loss Over Observed Representations} 
    + 
    \underbrace{
        \sum_{T=3}^{\bar{T}-1} \left( 
            \lambda_{3} \mathcal{L}_{\text{ant}}^{(1:T)} +
            \lambda_{4} \mathcal{L}_{\text{boxes}}^{(1:T)} + 
            \lambda_{5} \mathcal{L}_{\text{recon}}^{(1:T)}
        \right)
    }_\text{Loss Over Anticipated Representations}
\end{equation}

\section{Experiments}

\textbf{Dataset.} We apply the proposed method to anticipate future interactions on the Action Genome \cite{Ji_2019}. We pre-process the data and filter out videos with less than 3 annotated frames. Thus, we obtained 11.4K videos in total; we adhered to the train and test split provided by the dataset. The dataset encompasses 35 object classes and 25 relationship classes. These 25 relationship classes are grouped into three categories, namely: (1) \textbf{Attention Relations} comprise relationship classes which primarily describe attention of the subject towards the object, (2) \textbf{Spatial Relations} comprise relationship classes that describe the spatial relationship between two objects, and (3) \textbf{Contacting Relations} comprises relationship classes that indicate different ways the object is contacted. 




\textbf{Remark.} This dataset primarily includes videos featuring a single actor engaging with objects\footnote{Adopting an actor-centric approach aligns with the literature on VidSGG. While SceneSayer can anticipate relationship representations between any two objects in a scene, we specifically tailor it to address the dynamics between the interacting actor (subject) and the objects.} in various real-world environments. In the context of the action genome, a subject-object pair can demonstrate multiple spatial and contacting relationships.


\textbf{Evaluation Metric.} We evaluate our models using the standard \textit{Recall@K} and \textit{meanRecall@K} metrics, where $K$ is set to values within the set \{10, 20, 50\}. Recall@K metric assists in assessing the ability of our model to anticipate the relationships between observed objects in future frames. The long-tailed distribution of relationships in the training set \cite{nag_et_al_tempura_2023} can generate biased scene graphs. While the performance on more common relationships can dominate the Recall@K metrics, the mean recall metric introduced in \cite{chen2019knowledgeembedded} is a more balanced metric that scores the model's generalisation to all predicate classes. In the following sections, we will assess the performance of models by modifying the initial fraction of the video, denoted as \(\mathcal{F}\) provided as input. We set \(\mathcal{F}\) to 0.3, 0.5, 0.7, and 0.9 to facilitate a comprehensive understanding of the model's proficiency in short-term and long-term relationship anticipation. 



\textbf{Settings.} We define three settings to evaluate models for the SGA task, each varying by the amount of scene information provided as input: 
\begin{itemize}
    \item[--] \textbf{Action Genome Scenes (AGS):} In AGS, the model's input is limited to raw frames of the video. Thus, with minimal context and without any additional information we intend to challenge the model's ability to interpret and anticipate the scenes.
    \item[--] \textbf{Partially Grounded Action Genome Scenes (PGAGS):} In this intermediate setting, along with raw frames of the video we additionally input the model with precise bounding box information of active interacting objects observed in the scene.
    \item[--] \textbf{Grounded Action Genome Scenes (GAGS):} In GAGS, we provide the most comprehensive level of scene information. Specifically, the model takes precise bounding box information and the categories of the active objects observed in the interaction as input. In video data, observed frames are fraught with challenges such as object occlusions and blurry imagery, which complicate object detection and interpretation of interaction. Thus, in a setting where we provide complete information regarding a scene as input and reduce the noise induced by these challenges, we aim to evaluate the model's ability to understand spatio-temporal dynamics.
\end{itemize}


\begin{figure}[!ht]
    \centering
    \includegraphics[width=0.8\textwidth]{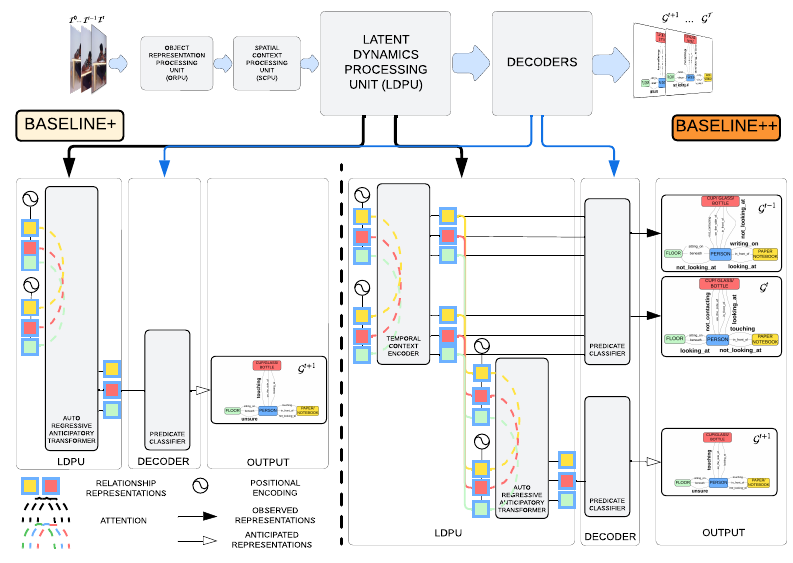}
    \caption{\textbf{Overview of Baselines.}  In our proposed \textit{Variant 1} (shown to the left), we input relationship representations to an anticipatory transformer to generate relationship representations for future frames auto-regressively. A predicate classification network (MLP) is then employed to decode these anticipated relationship representations. Meanwhile, in \textit{Variant 2} (shown to the right), we enhance relationship representations by passing them through a temporal encoder. These representations are fed to an auto-regressive anticipatory transformer to anticipate future relationship representations. Here, we employ two predicate classification heads (MLPs): one decodes the observed relationship representations, and the other decodes the anticipated relationship representations. In both variants proposed, the auto-regressive anticipatory transformer acts as the generative model, predicting the evolution of relationships between the interacting objects.}
    \label{fig:baselines}
\end{figure}

\subsection{Baselines}
We select two methods from VidSGG literature, STTran \cite{cong_et_al_sttran_2021} and DSGDetr \cite{Feng_2021}, as our strong baselines\footnote{Although VidSGG literature witnessed many approaches recently, most use the representation processing pipeline proposed by the two selected transformer-based VidSGG methods.}for adaptations. In our adaptation of selected methods to anticipate the relationships, we retain the \orpu (ORPU) and \scpu (SCPU) as proposed originally and introduce two variants of \ldpu (LDPU)\footnote{In SceneSayer, we developed a novel ORPU but adapted the SCPU from \cite{Feng_2021} and trained all units end-to-end from scratch. We noticed a drop in performance when we used pre-trained ORPU and SCPU (kept frozen during further training) units from \cite{Feng_2021,cong_et_al_sttran_2021} and trained the LDPU.} (see Fig. \ref{fig:baselines}). The two proposed variants differ in (1) the architecture of the model and (2) the loss functions employed to train them. 

In terms of architectural differences, in our \textit{baseline\textbf{+}} variants, we employ an anticipatory transformer built using the vanilla transformer architecture \cite{vaswani_et_al_2017} to generate future relationship representations by processing the relationship representations in the observed temporal context. Meanwhile, in our \textit{baseline\textbf{++}} variants, we 
introduce an additional temporal encoder to process the representations before we feed them through the anticipatory transformer. This component further refines the representations by  enhancing the model's understanding of spatio-temporal dynamics. In the context of loss functions, the approach followed by our \textit{baseline\textbf{+}} variants focuses solely on decoding the anticipated relationship representations. However, in contrast, our \textit{baseline\textbf{++}} variants adopt a more comprehensive strategy. These not only decode the anticipated representations but also simultaneously decode the observed representations. This dual-decoding approach allows for a more nuanced understanding of relationships.


\subsection{Results}

\textbf{Action Genome Scenes.} We present the results in Table. \ref{tab:anticipation_results_com_recalls_ags} where we observe that both STTran++ and DSGDetr++ consistently outperform their basic counterparts. Furthermore, it can be clearly noticed that the SceneSayerSDE model consistently outperforms the SceneSayerODE model, which, in turn, performs better than the baseline variants. Although SceneSayerODE/SDE models fall short on the R@50 metric in the No Constraint graph generation strategy we note that the R@K metric holds importance for lower values of K and, in R@10 and R@20 metrics, they consistently outperform baselines. We note that SceneSayerSDE is up to $\sim 70\%$ better on the R@10 metric than the best baseline variant. Additionally, the SceneSayerODE/SDE models exhibit reduced prediction bias, as evidenced by their performance across the mean recall metrics.

\begin{table}[!t]
    \centering
    \captionsetup{font=small}
    \caption{Results for \textbf{SGA of AGS}, when trained using anticipatory horizon of 3 future frames.}
    \label{tab:anticipation_results_com_recalls_ags}
    \setlength{\tabcolsep}{5pt} 
    \renewcommand{\arraystretch}{1.2} 
    \resizebox{0.70\textwidth}{!}{
    \begin{tabular}{ll|cccccc|cccccc}
    \hline
         & & \multicolumn{6}{c|}{\textbf{Recall (R)}} & \multicolumn{6}{c}{\textbf{Mean Recall (mR)}} \\ 
        \cmidrule(lr){3-8} \cmidrule(lr){9-14} 
         \multicolumn{2}{c|}{\textbf{\textbf{SGA of AGS}}} & \multicolumn{3}{c}{\textbf{With Constraint}} & \multicolumn{3}{c|}{\textbf{No Constraint}} & \multicolumn{3}{c}{\textbf{With Constraint}} & \multicolumn{3}{c}{\textbf{No Constraint}}\\ 
        \cmidrule(lr){1-2}\cmidrule(lr){3-5} \cmidrule(lr){6-8}\cmidrule(lr){9-11} \cmidrule(lr){12-14} 
         $\mathcal{F}$ & \textbf{Method} & \textbf{10} & \textbf{20} & \textbf{50} & \textbf{10} & \textbf{20} & \textbf{50} & \textbf{10} & \textbf{20} & \textbf{50}  & \textbf{10} & \textbf{20} & \textbf{50}   \\ \hline
        \multirow{4}{*}{0.3} & STTran+ \cite{cong_et_al_sttran_2021} & 12.5 & 19.6 & 20.7 & 13.9 & 21.6 & 40.8 & 3.4 & 5.3 & 5.7 & 3.5 & 7.3 & 20.3  \\ 
        & DSGDetr+ \cite{Feng_2021} & 12.8 & 19.6 & 20.4 & 14.3 & 21.8 & 41.3 & 3.5 & 5.3 & 5.6 & 3.6 & 7.6 & 21.2  \\ 
        & STTran++ \cite{cong_et_al_sttran_2021} & 18.5 & 27.9 & 29.5 & 15.4 & 27.2 & 48.6 & 5.9 & 10.4 & 11.3 & 6.2 & 14.1 & 31.2  \\ 
        & DSGDetr++ \cite{Feng_2021} & 19.5 & 28.3 & 29.4 & 16.8 & 29.0 & \cellcolor{highlightColor}  \textbf{48.9} & 6.0 & 10.3 & 11.0 & 8.4 & 16.7 & 32.3  \\ 
        & \textbf{SceneSayerODE (Ours)} & 23.1 & 29.2 & 31.4 & 23.3 & 32.5 & 45.1 & 10.6 & 13.8 & 15.0 & 13.3 & 20.1 & 33.0  \\ 
        & \textbf{SceneSayerSDE (Ours)} & \cellcolor{highlightColor}  \textbf{25.0} & \cellcolor{highlightColor}  \textbf{31.7} & \cellcolor{highlightColor}  \textbf{34.3} & \cellcolor{highlightColor}  \textbf{25.9} & \cellcolor{highlightColor}  \textbf{35.0} & 47.4 & \cellcolor{highlightColor}  \textbf{11.4} & \cellcolor{highlightColor}  \textbf{15.3} & \cellcolor{highlightColor}  \textbf{16.9} & \cellcolor{highlightColor}  \textbf{15.6} & \cellcolor{highlightColor}  \textbf{23.1} & \cellcolor{highlightColor}  \textbf{37.1}  \\ \hline
        \multirow{4}{*}{0.5} & STTran+ \cite{cong_et_al_sttran_2021} & 13.1 & 21.1 & 22.2 & 14.9 & 22.6 & 42.9 & 3.6 & 5.8 & 6.2 & 3.7 & 7.6 & 21.4  \\ 
        & DSGDetr+ \cite{Feng_2021} & 13.6 & 20.9 & 21.9 & 15.2 & 23.1 & 43.3 & 3.8 & 5.8 & 6.1 & 3.9 & 8.0 & 22.2  \\ 
        & STTran++ \cite{cong_et_al_sttran_2021} & 19.7 & 30.2 & 31.8 & 16.6 & 29.1 & 51.5 & 6.3 & 11.3 & 12.3 & 6.6 & 14.7 & 33.4  \\ 
        & DSGDetr++ \cite{Feng_2021} & 20.7 & 30.3 & 31.6 & 17.4 & 30.5 & \cellcolor{highlightColor}  \textbf{51.9} & 6.4 & 11.0 & 11.7 & 8.4 & 17.0 & 33.9  \\ 
        & \textbf{SceneSayerODE (Ours)} & 25.9 & 32.6 & 34.8 & 26.4 & 36.6 & 49.8 & 11.6 & 15.2 & 16.4 & 14.3 & 21.4 & 36.0  \\ 
        & \textbf{SceneSayerSDE (Ours)} & \cellcolor{highlightColor}  \textbf{27.3} & \cellcolor{highlightColor}  \textbf{34.8} & \cellcolor{highlightColor}  \textbf{37.0} & \cellcolor{highlightColor}  \textbf{28.4} & \cellcolor{highlightColor}  \textbf{38.6} & 51.4 & \cellcolor{highlightColor}  \textbf{12.4} & \cellcolor{highlightColor}  \textbf{16.6} & \cellcolor{highlightColor}  \textbf{18.0} & \cellcolor{highlightColor}  \textbf{16.3} & \cellcolor{highlightColor}  \textbf{25.1} & \cellcolor{highlightColor}  \textbf{39.9}  \\ \hline
        \multirow{4}{*}{0.7} & STTran+ \cite{cong_et_al_sttran_2021} & 14.9 & 23.4 & 24.7 & 16.6 & 25.1 & 47.2 & 4.1 & 6.5 & 7.0 & 4.2 & 8.5 & 24.0  \\ 
        & DSGDetr+ \cite{Feng_2021} & 15.5 & 23.4 & 24.3 & 16.8 & 25.3 & 47.4 & 4.3 & 6.5 & 6.9 & 4.3 & 8.8 & 24.7  \\ 
        & STTran++ \cite{cong_et_al_sttran_2021} & 22.1 & 33.6 & 35.2 & 19.0 & 32.8 & \cellcolor{highlightColor}  \textbf{56.8} & 7.0 & 12.6 & 13.6 & 7.7 & 17.1 & 36.8  \\ 
        & DSGDetr++ \cite{Feng_2021} & 22.9 & 33.6 & 34.9 & 19.8 & 34.1 & 56.7 & 7.1 & 12.6 & 13.3 & 9.5 & 19.2 & 37.2  \\ 
        & \textbf{SceneSayerODE (Ours)} & 30.3 & 36.6 & 38.9 & 32.1 & 42.8 & 55.6 & 12.8 & 16.4 & 17.8 & 16.5 & 24.4 & 39.6  \\ 
        & \textbf{SceneSayerSDE (Ours)} & \cellcolor{highlightColor}  \textbf{31.4} & \cellcolor{highlightColor}  \textbf{38.0} & \cellcolor{highlightColor}  \textbf{40.5} & \cellcolor{highlightColor}  \textbf{33.3} & \cellcolor{highlightColor}  \textbf{44.0} & 56.4 & \cellcolor{highlightColor}  \textbf{13.8} & \cellcolor{highlightColor}  \textbf{17.7} & \cellcolor{highlightColor}  \textbf{19.3} & \cellcolor{highlightColor}  \textbf{18.1} & \cellcolor{highlightColor}  \textbf{27.3} & \cellcolor{highlightColor}  \textbf{44.4}  \\ \hline
        \multirow{4}{*}{0.9} & STTran+ \cite{cong_et_al_sttran_2021} & 15.2 & 24.1 & 25.4 & 17.5 & 26.8 & 49.6 & 4.5 & 7.0 & 7.5 & 4.6 & 9.2 & 24.3  \\ 
        & DSGDetr+ \cite{Feng_2021} & 16.2 & 24.8 & 25.9 & 17.9 & 27.7 & 51.4 & 4.8 & 7.2 & 7.6 & 4.7 & 9.7 & 25.9  \\ 
        & STTran++ \cite{cong_et_al_sttran_2021} & 23.6 & 35.5 & 37.4 & 20.2 & 35.0 & 60.2 & 7.4 & 13.4 & 14.6 & 8.9 & 18.4 & 38.8  \\ 
        & DSGDetr++ \cite{Feng_2021} & 24.4 & 36.1 & 37.6 & 22.2 & 37.1 & 61.0 & 7.4 & 13.8 & 14.8 & 11.4 & 21.0 & 39.5  \\ 
        & \textbf{SceneSayerODE (Ours)} & 33.9 & 40.4 & 42.6 & 36.6 & 48.3 & 61.3 & 14.0 & 18.1 & 19.3 & 17.8 & 27.4 & 43.4  \\ 
        & \textbf{SceneSayerSDE (Ours)} & \cellcolor{highlightColor}  \textbf{34.8} & \cellcolor{highlightColor}  \textbf{41.9} & \cellcolor{highlightColor}  \textbf{44.1} & \cellcolor{highlightColor}  \textbf{37.3} & \cellcolor{highlightColor}  \textbf{48.6} & \cellcolor{highlightColor}  \textbf{61.6} & \cellcolor{highlightColor}  \textbf{15.1} & \cellcolor{highlightColor}  \textbf{19.4} & \cellcolor{highlightColor}  \textbf{21.0} & \cellcolor{highlightColor}  \textbf{20.8} & \cellcolor{highlightColor}  \textbf{30.9} & \cellcolor{highlightColor}  \textbf{46.8}  \\ \hline
    \end{tabular}
    }
\end{table}

\begin{table}[!ht]
    \centering
    \captionsetup{font=small}
    \caption{Results for \textbf{SGA of PGAGS}, when trained using anticipatory horizon of 3 future frames.}
    \label{tab:anticipation_results_com_recalls_pgags}
    \setlength{\tabcolsep}{5pt} 
    \renewcommand{\arraystretch}{1.2} 
    \resizebox{0.70\textwidth}{!}{
    \begin{tabular}{ll|cccccc|cccccc}
    \hline
         & & \multicolumn{6}{c|}{\textbf{Recall (R)}} & \multicolumn{6}{c}{\textbf{Mean Recall (mR)}} \\ 
        \cmidrule(lr){3-8} \cmidrule(lr){9-14} 
         \multicolumn{2}{c|}{\textbf{\textbf{SGA of PGAGS}}} & \multicolumn{3}{c}{\textbf{With Constraint}} & \multicolumn{3}{c|}{\textbf{No Constraint}} & \multicolumn{3}{c}{\textbf{With Constraint}} & \multicolumn{3}{c}{\textbf{No Constraint}}\\ 
        \cmidrule(lr){1-2}\cmidrule(lr){3-5} \cmidrule(lr){6-8}\cmidrule(lr){9-11} \cmidrule(lr){12-14} 
         $\mathcal{F}$ & \textbf{Method} & \textbf{10} & \textbf{20} & \textbf{50} & \textbf{10} & \textbf{20} & \textbf{50} & \textbf{10} & \textbf{20} & \textbf{50}  & \textbf{10} & \textbf{20} & \textbf{50}   \\ \hline
        \multirow{4}{*}{0.3} & STTran+ \cite{cong_et_al_sttran_2021} & 22.3 & 22.9 & 22.9 & 25.5 & 38.3 & 45.6 & 8.6 & 9.1 & 9.1 & 13.1 & 24.8 & 42.3  \\ 
        & DSGDetr+ \cite{Feng_2021} & 13.6 & 14.1 & 14.1 & 22.2 & 33.9 & 46.0 & 4.8 & 5.0 & 5.0 & 7.9 & 16.9 & 40.9  \\ 
        & STTran++ \cite{cong_et_al_sttran_2021} & 22.1 & 22.8 & 22.8 & 28.1 & 39.0 & 45.2 & 9.2 & 9.8 & 9.8 & 17.7 & 30.6 & 42.0  \\ 
        & DSGDetr++ \cite{Feng_2021} & 18.2 & 18.8 & 18.8 & 27.7 & 39.2 & \cellcolor{highlightColor}  \textbf{47.3} & 8.9 & 9.4 & 9.4 & 15.3 & 26.6 & 44.0  \\ 
        & \textbf{SceneSayerODE (Ours)} & 27.0 & 27.9 & 27.9 & 33.0 & 40.9 & 46.5 & 12.9 & 13.4 & 13.4 & 19.4 & 27.9 & 46.9  \\ 
        & \textbf{SceneSayerSDE (Ours)} & \cellcolor{highlightColor}  \textbf{28.8} & \cellcolor{highlightColor}  \textbf{29.9} & \cellcolor{highlightColor}  \textbf{29.9} & \cellcolor{highlightColor}  \textbf{34.6} & \cellcolor{highlightColor}  \textbf{42.0} & 46.2 & \cellcolor{highlightColor}  \textbf{14.2} & \cellcolor{highlightColor}  \textbf{14.7} & \cellcolor{highlightColor}  \textbf{14.7} & \cellcolor{highlightColor}  \textbf{21.5} & \cellcolor{highlightColor}  \textbf{31.7} & \cellcolor{highlightColor}  \textbf{48.2}  \\ \hline
        \multirow{4}{*}{0.5} & STTran+ \cite{cong_et_al_sttran_2021} & 24.2 & 24.9 & 24.9 & 27.5 & 41.6 & 49.9 & 9.3 & 9.9 & 9.9 & 13.8 & 26.7 & 43.1  \\ 
        & DSGDetr+ \cite{Feng_2021} & 15.8 & 16.2 & 16.2 & 24.7 & 38.2 & 52.4 & 5.5 & 5.7 & 5.7 & 8.7 & 18.5 & 41.5  \\ 
        & STTran++ \cite{cong_et_al_sttran_2021} & 24.5 & 25.2 & 25.2 & 30.6 & 43.2 & 50.2 & 10.1 & 10.7 & 10.7 & 18.4 & 29.5 & 43.1  \\ 
        & DSGDetr++ \cite{Feng_2021} & 20.7 & 21.4 & 21.4 & 30.4 & 44.0 & \cellcolor{highlightColor}  \textbf{52.7} & 10.2 & 10.8 & 10.8 & 16.5 & 30.8 & 45.1  \\ 
        & \textbf{SceneSayerODE (Ours)} & 30.5 & 31.5 & 31.5 & 36.8 & 45.9 & 51.8 & 14.9 & 15.4 & 15.5 & 21.6 & 30.8 & 48.0  \\ 
        & \textbf{SceneSayerSDE (Ours)} & \cellcolor{highlightColor}  \textbf{32.2} & \cellcolor{highlightColor}  \textbf{33.3} & \cellcolor{highlightColor}  \textbf{33.3} & \cellcolor{highlightColor}  \textbf{38.4} & \cellcolor{highlightColor}  \textbf{46.9} & 51.8 & \cellcolor{highlightColor}  \textbf{15.8} & \cellcolor{highlightColor}  \textbf{16.6} & \cellcolor{highlightColor}  \textbf{16.6} & \cellcolor{highlightColor}  \textbf{23.5} & \cellcolor{highlightColor}  \textbf{35.0} & \cellcolor{highlightColor}  \textbf{49.6}  \\ \hline
        \multirow{4}{*}{0.7} & STTran+ \cite{cong_et_al_sttran_2021} & 28.9 & 29.4 & 29.4 & 33.6 & 49.9 & 58.7 & 10.7 & 11.3 & 11.3 & 16.3 & 32.2 & 49.9  \\ 
        & DSGDetr+ \cite{Feng_2021} & 18.8 & 19.1 & 19.1 & 29.9 & 44.7 & 60.2 & 6.5 & 6.8 & 6.8 & 10.4 & 22.4 & 47.7  \\ 
        & STTran++ \cite{cong_et_al_sttran_2021} & 29.1 & 29.7 & 29.7 & 36.8 & 51.6 & 58.7 & 11.5 & 12.1 & 12.1 & 21.2 & 34.6 & 49.0  \\ 
        & DSGDetr++ \cite{Feng_2021} & 24.6 & 25.2 & 25.2 & 36.7 & 51.8 & \cellcolor{highlightColor}  \textbf{60.6} & 12.0 & 12.6 & 12.6 & 19.7 & 36.4 & 50.6  \\ 
        & \textbf{SceneSayerODE (Ours)} & 36.5 & 37.3 & 37.3 & 44.6 & 54.4 & 60.3 & 16.9 & 17.3 & 17.3 & 25.2 & 36.2 & 53.1  \\ 
        & \textbf{SceneSayerSDE (Ours)} & \cellcolor{highlightColor}  \textbf{37.6} & \cellcolor{highlightColor}  \textbf{38.5} & \cellcolor{highlightColor}  \textbf{38.5} & \cellcolor{highlightColor}  \textbf{45.6} & \cellcolor{highlightColor}  \textbf{54.6} & 59.3 & \cellcolor{highlightColor}  \textbf{18.4} & \cellcolor{highlightColor}  \textbf{19.1} & \cellcolor{highlightColor}  \textbf{19.1} & \cellcolor{highlightColor}  \textbf{28.3} & \cellcolor{highlightColor}  \textbf{40.9} & \cellcolor{highlightColor}  \textbf{54.9}  \\ \hline
        \multirow{4}{*}{0.9} & STTran+ \cite{cong_et_al_sttran_2021} & 30.9 & 31.3 & 31.3 & 39.9 & 56.6 & 63.8 & 11.5 & 11.9 & 11.9 & 20.3 & 37.7 & 54.3  \\ 
        & DSGDetr+ \cite{Feng_2021} & 21.3 & 21.6 & 21.6 & 38.4 & 54.9 & \cellcolor{highlightColor}  \textbf{68.7} & 7.5 & 7.7 & 7.7 & 13.7 & 29.0 & 55.4  \\ 
        & STTran++ \cite{cong_et_al_sttran_2021} & 31.1 & 31.6 & 31.6 & 43.5 & 57.6 & 63.9 & 12.4 & 12.8 & 12.8 & 25.3 & 39.6 & 54.0  \\ 
        & DSGDetr++ \cite{Feng_2021} & 27.6 & 28.1 & 28.1 & 45.8 & 61.5 & 68.5 & 13.2 & 13.7 & 13.7 & 25.8 & 42.9 & 58.3  \\ 
        & \textbf{SceneSayerODE (Ours)} & 41.6 & 42.2 & 42.2 & 52.7 & 61.8 & 66.5 & 19.0 & 19.4 & 19.4 & 29.4 & 42.2 & 59.2  \\ 
        & \textbf{SceneSayerSDE (Ours)} & \cellcolor{highlightColor}  \textbf{42.5} & \cellcolor{highlightColor}  \textbf{43.1} & \cellcolor{highlightColor}  \textbf{43.1} & \cellcolor{highlightColor}  \textbf{53.8} & \cellcolor{highlightColor}  \textbf{62.4} & 66.2 & \cellcolor{highlightColor}  \textbf{20.6} & \cellcolor{highlightColor}  \textbf{21.1} & \cellcolor{highlightColor}  \textbf{21.1} & \cellcolor{highlightColor}  \textbf{32.9} & \cellcolor{highlightColor}  \textbf{46.0} & \cellcolor{highlightColor}  \textbf{59.8}  \\ \hline
    \end{tabular}
    }
\end{table}

\textbf{Partially Grounded Action Genome Scenes.} We present the results in Table. \ref{tab:anticipation_results_com_recalls_pgags}. We observed that the STTran adaptation demonstrates superior performance compared to the DSGDetr adaptation across both model variants. Notably, the SceneSayer models significantly outperform all proposed baseline models. In particular, the SceneSayerSDE model achieves up to  $\sim 30\%$ improvement on the R@10 metric compared to the best-performing baseline model and both SceneSayerODE/SDE models outperform all proposed baseline models in mean recall metrics.


\textbf{Grounded Action Genome Scenes.} We present the results in Table. \ref{tab:anticipation_results_com_recalls_gags} and we observe that in our proposed baseline models, the adaptations of STTran perform better than the adaptations of DSGDetr and SceneSayer variants outperformed all baselines. 

\begin{table}[!ht]
    \centering
    \captionsetup{font=small}
    \caption{Results for \textbf{SGA of GAGS}, when trained using anticipatory horizon of 3 future frames.}
    \label{tab:anticipation_results_com_recalls_gags}
    \setlength{\tabcolsep}{5pt} 
    \renewcommand{\arraystretch}{1.2} 
    \resizebox{0.70\textwidth}{!}{
    \begin{tabular}{ll|cccccc|cccccc}
    \hline
         & & \multicolumn{6}{c|}{\textbf{Recall (R)}} & \multicolumn{6}{c}{\textbf{Mean Recall (mR)}} \\ 
        \cmidrule(lr){3-8} \cmidrule(lr){9-14} 
         \multicolumn{2}{c|}{\textbf{\textbf{SGA of GAGS}}} & \multicolumn{3}{c}{\textbf{With Constraint}} & \multicolumn{3}{c|}{\textbf{No Constraint}} & \multicolumn{3}{c}{\textbf{With Constraint}} & \multicolumn{3}{c}{\textbf{No Constraint}}\\ 
        \cmidrule(lr){1-2}\cmidrule(lr){3-5} \cmidrule(lr){6-8}\cmidrule(lr){9-11} \cmidrule(lr){12-14} 
         $\mathcal{F}$ & \textbf{Method} & \textbf{10} & \textbf{20} & \textbf{50} & \textbf{10} & \textbf{20} & \textbf{50} & \textbf{10} & \textbf{20} & \textbf{50}  & \textbf{10} & \textbf{20} & \textbf{50}   \\ \hline
        \multirow{4}{*}{0.3} & STTran+ \cite{cong_et_al_sttran_2021} & 30.8 & 32.8 & 32.8 & 30.6 & 47.3 & 62.8 & 7.1 & 7.8 & 7.8 & 9.5 & 19.6 & 45.7  \\ 
        & DSGDetr+ \cite{Feng_2021} & 27.0 & 28.9 & 28.9 & 30.5 & 45.1 & 62.8 & 6.7 & 7.4 & 7.4 & 9.5 & 19.4 & 45.2  \\ 
        & STTran++ \cite{cong_et_al_sttran_2021} & 30.7 & 33.1 & 33.1 & 35.9 & 51.7 & 64.1 & 11.8 & 13.3 & 13.3 & 16.5 & 29.3 & 50.2  \\ 
        & DSGDetr++ \cite{Feng_2021} & 25.7 & 28.2 & 28.2 & 36.1 & 50.7 & 64.0 & 11.1 & 12.8 & 12.8 & 19.7 & 32.0 & 51.1  \\ 
        & \textbf{SceneSayerODE (Ours)} & 34.9 & 37.3 & 37.3 & 40.5 & 54.1 & 63.9 & 15.1 & 16.6 & 16.6 & 19.6 & 31.6 & 55.8  \\ 
        & \textbf{SceneSayerSDE (Ours)} & \cellcolor{highlightColor}  \textbf{39.7} & \cellcolor{highlightColor}  \textbf{42.2} & \cellcolor{highlightColor}  \textbf{42.3} & \cellcolor{highlightColor}  \textbf{46.9} & \cellcolor{highlightColor}  \textbf{59.1} & \cellcolor{highlightColor}  \textbf{65.2} & \cellcolor{highlightColor}  \textbf{18.4} & \cellcolor{highlightColor}  \textbf{20.5} & \cellcolor{highlightColor}  \textbf{20.5} & \cellcolor{highlightColor}  \textbf{24.6} & \cellcolor{highlightColor}  \textbf{37.8} & \cellcolor{highlightColor}  \textbf{59.0}  \\ \hline
        \multirow{4}{*}{0.5} & STTran+ \cite{cong_et_al_sttran_2021} & 35.0 & 37.1 & 37.1 & 34.4 & 53.4 & 70.8 & 8.0 & 8.7 & 8.8 & 10.5 & 21.5 & 48.8  \\ 
        & DSGDetr+ \cite{Feng_2021} & 31.2 & 33.3 & 33.3 & 34.3 & 51.0 & 70.8 & 7.8 & 8.6 & 8.6 & 10.5 & 21.4 & 48.4  \\ 
        & STTran++ \cite{cong_et_al_sttran_2021} & 35.6 & 38.1 & 38.1 & 40.3 & 58.4 & 72.2 & 13.4 & 15.2 & 15.2 & 17.8 & 32.5 & 53.7  \\ 
        & DSGDetr++ \cite{Feng_2021} & 29.3 & 31.9 & 32.0 & 40.3 & 56.9 & 72.0 & 12.2 & 13.8 & 13.9 & 20.6 & 34.3 & 54.0  \\ 
        & \textbf{SceneSayerODE (Ours)} & 40.7 & 43.4 & 43.4 & 47.0 & 62.2 & 72.4 & 17.4 & 19.2 & 19.3 & 22.8 & 35.2 & 60.2  \\ 
        & \textbf{SceneSayerSDE (Ours)} & \cellcolor{highlightColor}  \textbf{45.0} & \cellcolor{highlightColor}  \textbf{47.7} & \cellcolor{highlightColor}  \textbf{47.7} & \cellcolor{highlightColor}  \textbf{52.5} & \cellcolor{highlightColor}  \textbf{66.4} & \cellcolor{highlightColor}  \textbf{73.5} & \cellcolor{highlightColor}  \textbf{20.7} & \cellcolor{highlightColor}  \textbf{23.0} & \cellcolor{highlightColor}  \textbf{23.1} & \cellcolor{highlightColor}  \textbf{26.6} & \cellcolor{highlightColor}  \textbf{40.8} & \cellcolor{highlightColor}  \textbf{63.8}  \\ \hline
        \multirow{4}{*}{0.7} & STTran+ \cite{cong_et_al_sttran_2021} & 40.0 & 41.8 & 41.8 & 41.0 & 62.5 & 80.4 & 9.1 & 9.8 & 9.8 & 12.6 & 26.3 & 57.5  \\ 
        & DSGDetr+ \cite{Feng_2021} & 35.5 & 37.3 & 37.3 & 41.0 & 59.8 & 80.7 & 8.9 & 9.6 & 9.6 & 12.6 & 26.2 & 57.4  \\ 
        & STTran++ \cite{cong_et_al_sttran_2021} & 41.3 & 43.6 & 43.6 & 48.2 & 68.8 & 82.0 & 16.3 & 18.2 & 18.2 & 22.3 & 39.5 & 63.1  \\ 
        & DSGDetr++ \cite{Feng_2021} & 33.9 & 36.3 & 36.3 & 48.0 & 66.7 & 81.9 & 14.2 & 15.9 & 15.9 & 24.5 & 41.1 & 63.4  \\ 
        & \textbf{SceneSayerODE (Ours)} & 49.1 & 51.6 & 51.6 & 58.0 & 74.0 & 82.8 & 21.0 & 22.9 & 22.9 & 27.3 & 43.2 & 70.5  \\ 
        & \textbf{SceneSayerSDE (Ours)} & \cellcolor{highlightColor}  \textbf{52.0} & \cellcolor{highlightColor}  \textbf{54.5} & \cellcolor{highlightColor}  \textbf{54.5} & \cellcolor{highlightColor}  \textbf{61.8} & \cellcolor{highlightColor}  \textbf{76.7} & \cellcolor{highlightColor}  \textbf{83.4} & \cellcolor{highlightColor}  \textbf{24.1} & \cellcolor{highlightColor}  \textbf{26.5} & \cellcolor{highlightColor}  \textbf{26.5} & \cellcolor{highlightColor}  \textbf{31.9} & \cellcolor{highlightColor}  \textbf{48.0} & \cellcolor{highlightColor}  \textbf{74.2}  \\ \hline
        \multirow{4}{*}{0.9} & STTran+ \cite{cong_et_al_sttran_2021} & 44.7 & 45.9 & 45.9 & 50.9 & 74.8 & 90.9 & 10.3 & 10.8 & 10.8 & 16.3 & 33.7 & 71.4  \\ 
        & DSGDetr+ \cite{Feng_2021} & 38.8 & 40.0 & 40.0 & 51.0 & 71.7 & 91.2 & 10.2 & 10.7 & 10.7 & 16.3 & 33.7 & 71.6  \\ 
        & STTran++ \cite{cong_et_al_sttran_2021} & 46.0 & 47.7 & 47.7 & 60.2 & 81.5 & 92.3 & 19.6 & 21.4 & 21.4 & 29.6 & 49.1 & 76.4  \\ 
        & DSGDetr++ \cite{Feng_2021} & 38.1 & 39.8 & 39.8 & 58.8 & 78.8 & 92.2 & 16.3 & 17.7 & 17.7 & 30.7 & 50.3 & 77.2  \\ 
        & \textbf{SceneSayerODE (Ours)} & 58.1 & 59.8 & 59.8 & 72.6 & 86.7 & 93.2 & 25.0 & 26.4 & 26.4 & 35.0 & 51.7 & 80.2  \\ 
        & \textbf{SceneSayerSDE (Ours)} & \cellcolor{highlightColor}  \textbf{60.3} & \cellcolor{highlightColor}  \textbf{61.9} & \cellcolor{highlightColor}  \textbf{61.9} & \cellcolor{highlightColor}  \textbf{74.8} & \cellcolor{highlightColor}  \textbf{88.0} & \cellcolor{highlightColor}  \textbf{93.5} & \cellcolor{highlightColor}  \textbf{28.5} & \cellcolor{highlightColor}  \textbf{29.8} & \cellcolor{highlightColor}  \textbf{29.8} & \cellcolor{highlightColor}  \textbf{40.0} & \cellcolor{highlightColor}  \textbf{57.7} & \cellcolor{highlightColor}  \textbf{87.2}  \\ \hline
    \end{tabular}
    }
\end{table}

\noindent Specifically, the proposed SceneSayerSDE model demonstrated remarkable efficiency, marking approximately a 26\%  improvement in the R@10 metric compared to the best-performing baseline. Furthermore, both SceneSayerSDE/ODE variants demonstrated superior performance consistently over the baseline variants on the mean recall metrics.

\textbf{Qualitative Results.} In Fig. \ref{fig:qualitative_results}, we provide qualitative SGA results on AG dataset under different settings. These results are the output of baseline variants and SceneSayer on 70\% observation. The comparative analysis demonstrates SceneSayer's superior performance. Specifically, under the AGS setting, SceneSayerODE outshines all with the fewest incorrect anticipations, closely followed by SceneSayerSDE. In the PGAGS and GAGS, SceneSayerSDE surpasses the performance of all other models.

\begin{figure}[!t]
    \begin{center}
        \includegraphics[width=\textwidth]{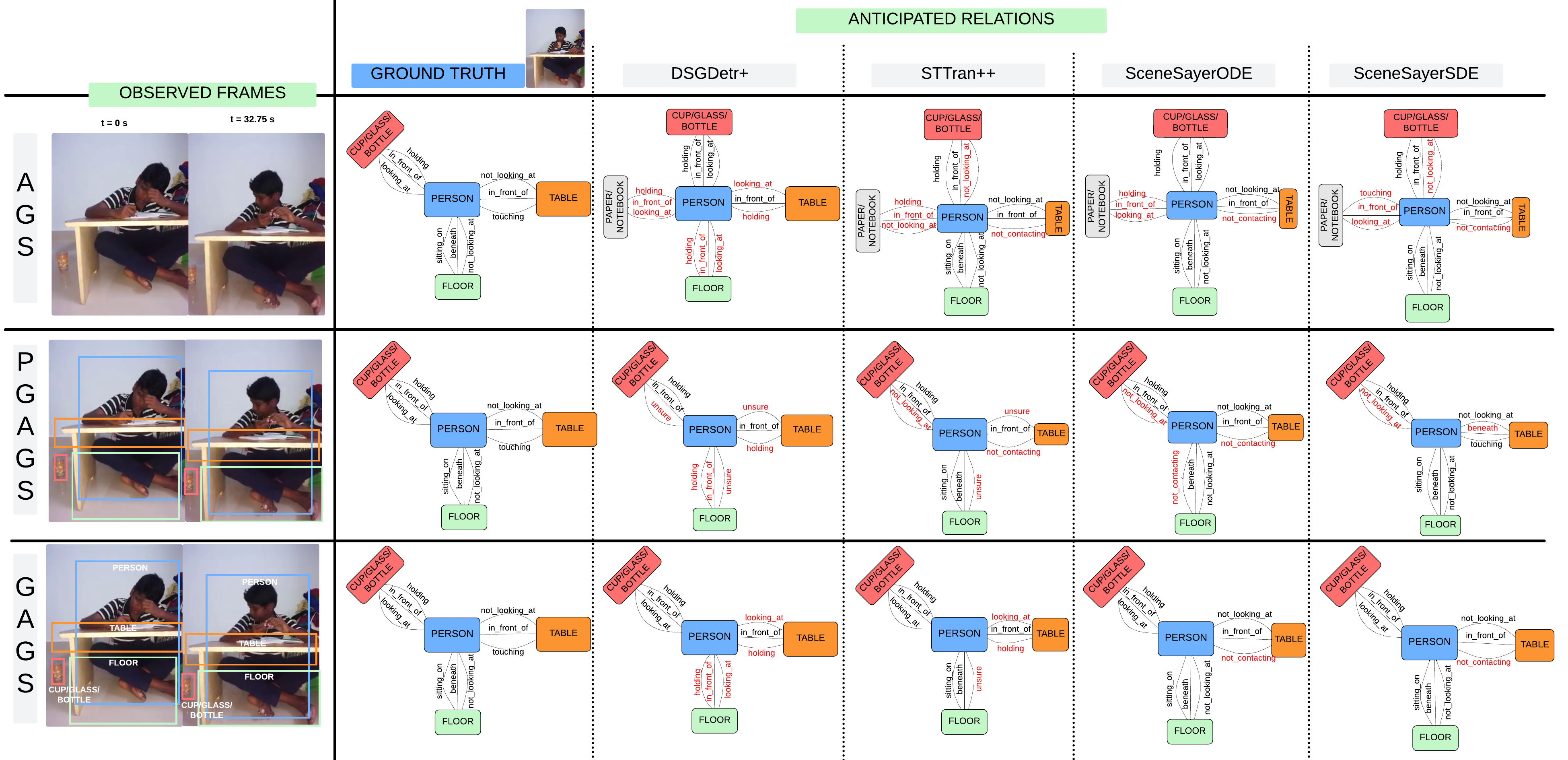}
    \end{center}
    \captionsetup{font=small}
    \caption{\textbf{Qualitative Results} To the left, we show a sampled subset of the frames observed by the models. The second column provides a ground truth scene graph corresponding to a future frame. In the subsequent columns, we contrast the performance of baseline variants with the proposed SceneSayer models. In each graph above, correct anticipations of relationships are denoted with text in black and incorrect anticipation of the relationships are highlighted with text in red.}
    \label{fig:qualitative_results}
\end{figure}

\section{Discussion, Limitations \& Future Work}

\underline{\textbf{Discussion}}: We introduced a novel task, Scene Graph Anticipation (SGA) and proposed approaches that model the latent dynamics of the evolution of relationship representations between interacting objects.
\underline{\textbf{Limitations}}: Our evaluation focused on assessing the model's ability to anticipate symbolic scene graphs in future frames and excluded the prediction performance of object localization in future frames. \underline{\textbf{Future Work}}: Our work opens up several avenues for future work. First, an exciting direction is to include localization in the framework. Second, SGA as a tool can be used to develop methods for Surveillance Systems, Robotics, etc. Third, leveraging commonsense knowledge derived from Large Language Models or Vision Language Models to enhance the anticipation of symbolic scene graphs in future frames offers another promising direction.

\section*{Acknowledgements}
This work was supported by IBM AI Horizons Network (AIHN) grant and IBM SUR Award. Parag Singla was supported by Shanthi and K Ananth Krishnan Young Faculty Chair Professorship in AI. 
Rohith Peddi and Vibhav Gogate were supported in part by the DARPA
Perceptually-Enabled Task Guidance (PTG) Program under contract number HR00112220005, by the DARPA Assured Neuro Symbolic Learning and Reasoning (ANSR) Program under contract number HR001122S0039, by the National Science Foundation grant IIS-1652835 and by the AFOSR award
FA9550-23-1-0239. We would like to thank IIT Delhi HPC facility\footnote{https://supercomputing.iitd.ac.in/}
for computational resources. We would also like to thank Karan Aggarwal for useful discussions during the initial phase of the project. We would like to express our gratitude to all the reviewers for their insightful comments and queries, which have substantially enhanced the quality of our manuscript.

\newpage

\appendix



\newpage

\section{Overview}

In this paper, we introduce Scene Graph Anticipation (SGA), a novel task that anticipates relationships between interacting objects in future frames and constructs scene graphs for future frames based on anticipated relationships.

\paragraph{Adaptation of Existing Methodologies.} We select and adapt two state-of-the-art VidSGG methods \cite{cong_et_al_sttran_2021, Feng_2021} for the task of SGA (see Sec. \ref{sec:technical_approach}). We elaborate on the modifications and enhancements made to these chosen methods to learn the spatio-temporal dynamics of interacting objects, thereby predicting the relationships in future frames.

\paragraph{Novel Approaches.} We present SceneSayer, a continuous-time framework designed to learn the latent dynamics governing the evolution of relationships between interacting objects (see Sec. \ref{sec:technical_approach}). We also discuss the tailored loss functions and the hyperparameters employed to train the proposed adaptations and novel approach in Sec. \ref{sec:technical_approach}.

\paragraph{Ablation - Components.} We analyse the impact of various components within our proposed models. Here, in Sec. \ref{sec:loss_function}, we examine the significance of each component of the proposed loss functions and, in Sec. \ref{sec:solvers}, we present results on the application of different solvers to solve the learned differential equations.

\paragraph{Ablation - Train/Evaluation Horizons.}

Sections \ref{sec:context_fraction}, \ref{sec:future_frame} present results and an in-depth discussion on the models' performance across varying training and evaluation horizons.

\paragraph{Conclusion and Paper Contributions}

This paper makes the following contributions:
\begin{enumerate}
    \item Introduce the Scene Graph Anticipation (SGA) task, aimed at learning the evolution of spatio-temporal dynamics of relationships.
    \item Adapt and enhance two state-of-the-art VidSGG methods for the SGA task.
    \item Propose SceneSayer, a novel continuous-time framework that learns the differential equation governing the evolution of relationships between interacting objects.
\end{enumerate}

\newpage

\section{Technical Approach}
\label{sec:technical_approach}

\subsection{Baseline - Variant 1}

\begin{figure}[!h]
    \centering
    \includegraphics[width=0.835\textwidth]{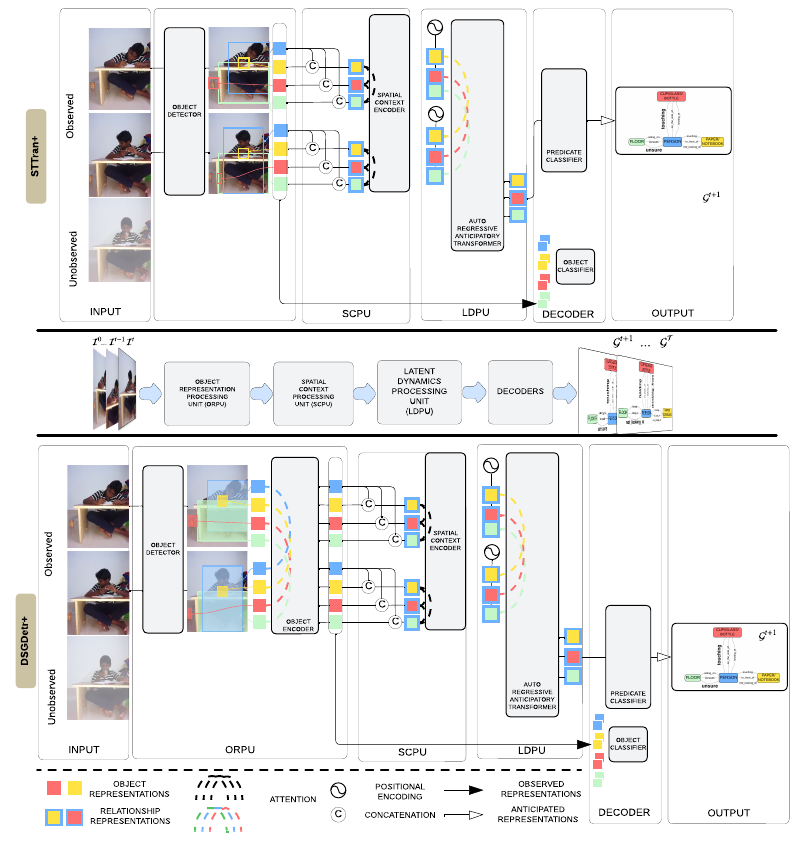}
    \caption{The \textbf{STTran+} and \textbf{DSGDetr+} models differ primarily in their ORPU; thereafter, they follow a unified pipeline. \textbf{Differences in ORPU:} (1) In \textbf{STTran+}, the ORPU begins with extracting features using a pre-trained object detector. These features are then processed in the SCPU to construct relationship representations. (2) Conversely, \textbf{DSGDetr+}'s ORPU starts with the initial bounding box and class label detection. Object tracking is established using the Hungarian Matching algorithm, and these tracks are passed through a transformer encoder to construct temporally consistent object representations. \textbf{Unified Pipeline Post-ORPU:} The relationship representations constructed from the concatenation of object representations from each model's unique ORPU are passed through a transformer encoder to aggregate information from the spatial context. This is followed by processing in the LDPU, where we employ an auto-regressive transformer to predict relationship representations of interacting objects in future frames. Finally, these representations are decoded for predicate classification. }
    \label{fig:baseline_variant_1_combined}
\end{figure}

In Fig. \ref{fig:baseline_variant_1_combined}, we present a variant of the proposed architecture to adapt the selected methods for the task of SGA. We re-use ORPUs and SCPUs from the selected methods and perform changes in the subsequent stages of the pipeline.

\subsubsection{Loss Function.} To train the models proposed in Fig. \ref{fig:baseline_variant_1_combined}. We employ the following loss functions: (\RNum{1}) Object Classification Loss, (\RNum{2}) Predicate Classification Loss, and (\RNum{3}) Reconstruction Loss. The total objective for training can be written as:



\begin{equation}
    \overbrace{
        \mathcal{L}_{i} = \sum_{t=1}^{\bar{T}} \mathcal{L}_{i}^{t},
        \quad
        \mathcal{L}_{i}^{t} = -\sum_{n=1}^{|\mathcal{C}|} y_{i,n}^{t} \log(\hat{\mathbf{c}}_{i,n}^{t})
    }^\text{Object Classification Loss (\RNum{1})}
\end{equation}

\begin{equation}
     \underbrace{
        \mathcal{L}_{\text{ant}}^{(1:T)} = \sum_{t=T+1}^{\min(T+H, \bar{T})} \mathcal{L}_{\text{ant}}^t, 
        \quad
        \mathcal{L}_{\text{ant}}^t = \sum_{ij} \mathcal{L}_{p^t_{ij}}
    }_\text{Predicate Classification Loss (\RNum{2})}
\end{equation}

\begin{equation}
    \underbrace{
        \mathcal{L}_{\text{recon}}^{(1:T)} = \sum_{t = T+1}^{\min(T+H, \bar{T})} \mathcal{L}_{\text{recon}}^{t},
        \quad
        \mathcal{L}_{\text{recon}}^t = \frac{1}{N(t) \times N(t)} \sum_{ij}^ {(N(t) \times N(t))}\text{L}_{\text{smooth}}(\mathbf{z}_{ij}^t - \hat{\mathbf{z}}_{ij}^t)
    }_\text{Reconstruction Loss (\RNum{3})}
\end{equation}

Thus the total objective can be written as follows:

\begin{equation}
    \mathcal{L} = 
    \underbrace{
        \sum_{t=1}^{\bar{T}} \left(\lambda_{1} \sum_{i} \mathcal{L}_{i}^{t} \right)
    }_\text{Loss Over Observed Representations} + 
    \underbrace{
        \sum_{T=3}^{\bar{T}-1} \left( 
            \lambda_{2} \mathcal{L}_{\text{ant}}^{(1:T)} +
            \lambda_{3} \mathcal{L}_{\text{recon}}^{(1:T)}
        \right)
    }_\text{Loss Over Anticipated Representations}
\end{equation}

\subsubsection{Implementation Details.} Although the algorithm \ref{alg:train_baseline_variant_1_combined} presents a sequential version of training. We leverage the masking of frames to train the model in parallel.  We chose hyperparameters $\lambda_1, \lambda_2, \lambda_3 \text{ as } \{1, 2, 2\} \text{, respectively}$

\begin{algorithm}
\SetCustomAlgoRuledWidth{\textwidth}
\DontPrintSemicolon
\footnotesize{
\KwIn{Train Horizon: $H$, Observed frames: $\topsequence{I}{t}{1}{T}$, ORPU: $\theta_{orpu}$, Spatial Encoder: $\theta_{sp}$ and Anticipatory Transformer: $\theta_{at}$}}
\textcolor{blue}{*** Randomly initialize $\theta_{sp},  \theta_{orpu},\theta_{at}$ ***} \;
\SetKwBlock{Begin}{function}{end function}{
    \For{epochs $n$ in $1, \cdots, N$}
    {
        \For{videos $v$ in $1, \cdots, V$}
        {
             \For{$T$ in $3, \cdots, N_{F}-H$}
            {   
                \textcolor{blue}{*** Fetch features, bounding boxes, classes ***} \;
                $\{ \mathbf{v}^{t}, \mathbf{b}^{t}, \mathbf{c}^{t} \}_{t=1}^{T} $ = $\operatorname{ORPU}(\topsequence{I}{t}{1}{T})$ \;
                
                \textcolor{blue}{*** Construct relationship representations ***} \;
                $\{ \mathbf{z}_{ij}^{t} = \operatorname{Concat} (\mathbf{W}_{1} \mathbf{v}^{t}_{i}, \mathbf{W}_{2} \mathbf{v}^{t}_{j}, \mathbf{W}_{3}\mathbf{U}_{ij}^{t}, \mathbf{S}_{i}^{t}, \mathbf{S}_{j}^{t}) \}_{t=1}^{T}$ \;

                \textcolor{blue}{*** Stack all relationships in a frame ***} \;
                $\mathbf{Z}^{t} = \operatorname{Stack}(\{\mathbf{z}_{ij}^{t}\}_{ij})$  \;

                \textcolor{blue}{*** Pass through the spatial encoder ***} \;                   
                $\{ \mathbf{Z}^{t} = \operatorname{SpatialEncoder}(\mathbf{Z}^{t}) \}_{t=1}^{T}$ \;

                \textcolor{blue}{*** Stack relationships from observed frames ***} \;
                $\{ \mathbf{Z}_{ij}^{(1:T)} = \operatorname{Stack}( \{\mathbf{z}_{ij}^{t}\}_{t=1}^{T}) \}_{ij}$ \;

                \For{future $h$ in $1, \cdots, H$}
                {   
                    \textcolor{blue}{*** Forward pass through Anticipatory Transformer ***} \;
                     $\{ \mathbf{z}_{ij}^{T+h} = \operatorname{AnticipatoryTransformer}(\mathbf{Z}_{ij}^{(1:T+h-1)}) \}_{ij}$ \;
                    \textcolor{blue}{*** Append generated future relationships to context ***} \;
                     $\{  \mathbf{Z}_{ij}^{(1:T+h)} = \operatorname{Concat}( \mathbf{Z}_{ij}^{(1:T+h-1)}, \mathbf{z}_{ij}^{T+h}) \}_{ij}$ \;

                     Pass $\mathbf{z}_{ij}^{T+h}$ through decoder \;
                }
            
            }
        }
    }

}
\Return{An assignment to $\theta_{sp},  \theta_{orpu}, \theta_{at}$}
\caption{Training algorithm for STTran+ and DSGDetr+}
\label{alg:train_baseline_variant_1_combined}
\end{algorithm}

\newpage

\subsection{Baseline - Variant 2}

\begin{figure} [!h]
    \centering
    \includegraphics[width=0.83\textwidth]{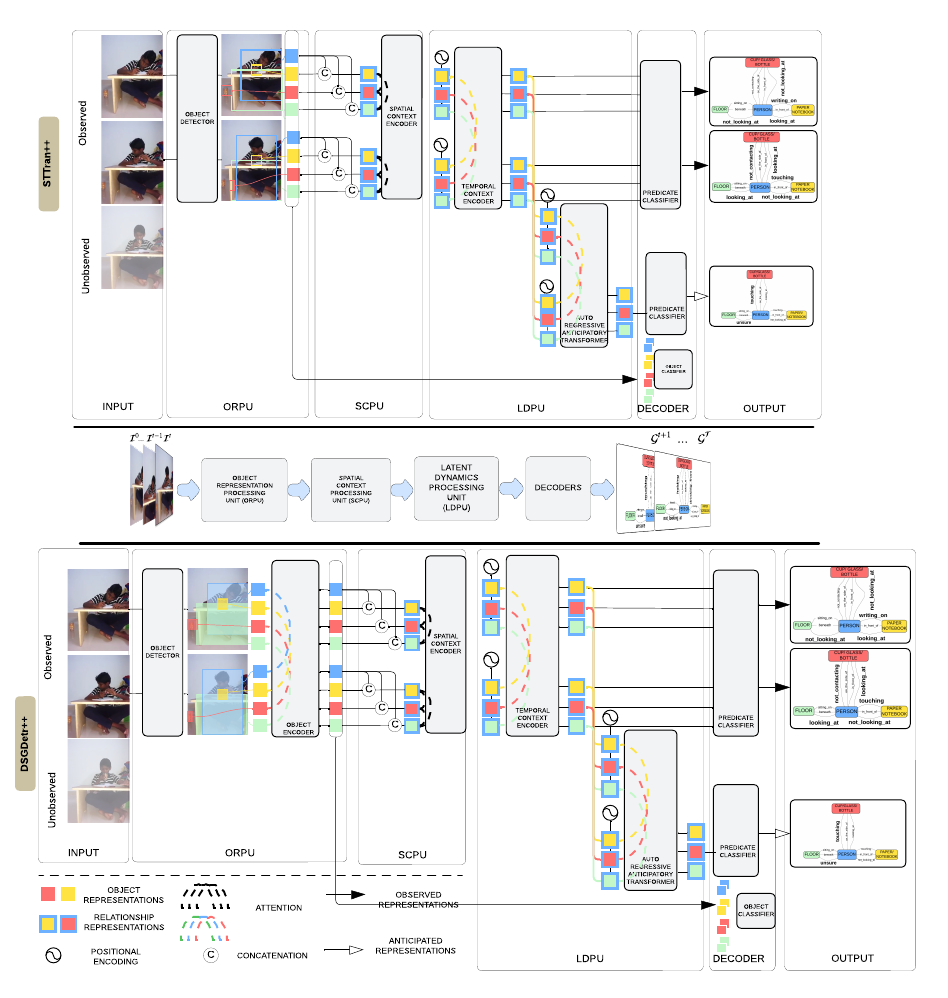}
    \caption{The \textbf{STTran++} and \textbf{DSGDetr++} models differ primarily in their ORPU; thereafter, they follow a unified pipeline. \textbf{Differences in ORPU:} (1) In \textbf{STTran++}, the ORPU begins with extracting features using a pre-trained object detector. These features are then processed in the SCPU to construct relationship representations. (2) Conversely, \textbf{DSGDetr++}'s ORPU starts with the initial bounding box and class label detection. Object tracking is established using the Hungarian Matching algorithm, and these tracks are passed through an encoder to construct temporally consistent object representations. \textbf{Unified Pipeline Post-ORPU:} The relationship representations constructed from the concatenation of object representations from each model's unique ORPU are passed through an encoder to aggregate information from the spatial context. \textcolor{teal}{These spatially aware representations are passed through an encoder to aggregate the spatio-temporal context}. This is followed by processing in the LDPU, where we employ an auto-regressive transformer to predict relationship representations of interacting objects in future frames. \textcolor{teal}{Finally, we employ a dual decoding strategy to decode both observed and predicted relationship representations}.}
    \label{fig:baseline_variant_2_combined}
\end{figure}

In Fig. \ref{fig:baseline_variant_2_combined}, we present a variant of the proposed architecture to adapt the selected methods for the task of SGA. We re-use ORPUs and SCPUs from the selected methods and perform changes in the subsequent stages of the pipeline.

\begin{algorithm}[!h]
\SetCustomAlgoRuledWidth{\textwidth}
\DontPrintSemicolon
\footnotesize{
\KwIn{Train Horizon: $H$, Observed frames: $\topsequence{I}{t}{1}{T}$, ORPU: $\theta_{orpu}$, Spatial Encoder: $\theta_{sp}$, Temporal Encoder: $\theta_{tp}$, and Anticipatory Transformer: $\theta_{at}$}}
\textcolor{blue}{*** Randomly initialize $ \theta_{orpu}, \theta_{sp}, \theta_{tp}, \theta_{at}$ ***} \;
\SetKwBlock{Begin}{function}{end function}{
    \For{epochs $n$ in $1, \cdots, N$}
    {
        \For{videos $v$ in $1, \cdots, V$}
        {
             \For{$T$ in $3, \cdots, N_{F}-H$}
            {   
                \textcolor{blue}{*** Fetch features, bounding boxes, classes ***} \;
                $\{ \mathbf{v}^{t}, \mathbf{b}^{t}, \mathbf{c}^{t} \}_{t=1}^{T} $ = $\operatorname{ORPU}(\topsequence{I}{t}{1}{T})$ \;
                
                \textcolor{blue}{*** Construct relationship representations ***} \;
                $\{ \mathbf{z}_{ij}^{t} = \operatorname{Concat} (\mathbf{W}_{1} \mathbf{v}^{t}_{i}, \mathbf{W}_{2} \mathbf{v}^{t}_{j}, \mathbf{W}_{3}\mathbf{U}_{ij}^{t}, \mathbf{S}_{i}^{t}, \mathbf{S}_{j}^{t}) \}_{t=1}^{T}$ \;

                \textcolor{blue}{*** Stack all relationships in a frame ***} \;
                $\mathbf{Z}^{t} = \operatorname{Stack}(\{\mathbf{z}_{ij}^{t}\}_{ij})$  \;

                \textcolor{blue}{*** Pass through the spatial encoder ***} \;                   
                $\{ \mathbf{Z}^{t} = \operatorname{SpatialEncoder}(\mathbf{Z}^{t}) \}_{t=1}^{T}$ \;

                \textcolor{blue}{*** Stack relationships from observed frames ***} \;
                $\{ \mathbf{Z}_{ij}^{(1:T)} = \operatorname{Stack}( \{\mathbf{z}_{ij}^{t}\}_{t=1}^{T}) \}_{ij}$ \;

                \textcolor{blue}{*** Pass through temporal encoder ***} \;
                $\{ \mathbf{Z}_{ij}^{(1:T)} =  \operatorname{TemporalEncoder}(\mathbf{Z}_{ij}^{(1:T)}) \}_{ij}$ \;

                \textcolor{blue}{*** Predicate Classification Decoder ***} \;
                Pass $\mathbf{Z}_{ij}^{(1:T)}$ through decoder \;

                \For{future $h$ in $1, \cdots, H$}
                {   
                    \textcolor{blue}{*** Forward pass through Anticipatory Transformer ***} \;
                     $\{ \mathbf{z}_{ij}^{T+h} = \operatorname{AnticipatoryTransformer}(\mathbf{Z}_{ij}^{(1:T+h-1)}) \}_{ij}$ \;
                    \textcolor{blue}{*** Append generated future relationships to context ***} \;
                     $\{  \mathbf{Z}_{ij}^{(1:T+h)} = \operatorname{Concat}( \mathbf{Z}_{ij}^{(1:T+h-1)}, \mathbf{z}_{ij}^{T+h}) \}_{ij}$ \;

                    \textcolor{blue}{*** Predicate Classification Decoder ***} \;
                     Pass $\mathbf{z}_{ij}^{T+h}$ through decoder \;
                }
            
            }
        }
    }

}
\Return{An assignment to $\theta_{sp}, \theta_{orpu}, \theta_{tp}, \theta_{at}$}
\caption{Training algorithm for STTran++ and DSGDetr++}
\label{alg:baseline_variant_2_combined}
\end{algorithm}

\subsubsection{Loss Function.} The total objective for training can be written as:


\begin{equation}
    \overbrace{
        \mathcal{L}_{i} = \sum_{t=1}^{\bar{T}} \mathcal{L}_{i}^{t},
        \quad
        \mathcal{L}_{i}^{t} = -\sum_{n=1}^{|\mathcal{C}|} y_{i,n}^{t} \log(\hat{\mathbf{c}}_{i,n}^{t})
    }^\text{Object Classification Loss (\RNum{1})}
\end{equation}

\begin{equation}
    \overbrace{
        \mathcal{L}_{\text{gen}} = \sum_{t=1}^{\bar{T}} \mathcal{L}_{\text{gen}}^t, 
        \quad
        \mathcal{L}_{\text{gen}}^t = \sum_{ij} \mathcal{L}_{p^t_{ij}}
    }^\text{Predicate Classification Loss (\RNum{2})}
\end{equation}

\begin{equation}
     \underbrace{
        \mathcal{L}_{\text{ant}}^{(1:T)} = \sum_{t=T+1}^{\min(T+H, \bar{T})} \mathcal{L}_{\text{ant}}^t, 
        \quad
        \mathcal{L}_{\text{ant}}^t = \sum_{ij} \mathcal{L}_{p^t_{ij}}
    }_\text{Predicate Classification Loss (\RNum{3})}
\end{equation}

\begin{equation}
    \underbrace{
        \mathcal{L}_{\text{recon}}^{(1:T)} = \sum_{t = T+1}^{\min(T+H, \bar{T})} \mathcal{L}_{\text{recon}}^{t},
        \quad
        \mathcal{L}_{\text{recon}}^t = \frac{1}{N(t) \times N(t)} \sum_{ij}^ {(N(t) \times N(t))}\text{L}_{\text{smooth}}(\mathbf{z}_{ij}^t - \hat{\mathbf{z}}_{ij}^t)
    }_\text{Reconstruction Loss (\RNum{4})}
\end{equation}

Thus, the total objective for training the proposed method can be written as:

\begin{equation}
    \mathcal{L} = 
    \underbrace{
        \sum_{t=1}^{\bar{T}} \left(\lambda_{1} \mathcal{L}_{\text{gen}}^{t} + 
        \lambda_{2} \sum_{i} \mathcal{L}_{i}^{t} \right)
    }_\text{Loss Over Observed Representations} 
    + 
    \underbrace{
        \sum_{T=3}^{\bar{T}-1} \left( 
            \lambda_{3} \mathcal{L}_{\text{ant}}^{(1:T)} +
            \lambda_{4} \mathcal{L}_{\text{recon}}^{(1:T)}
        \right)
    }_\text{Loss Over Anticipated Representations}
\end{equation}

\subsubsection{Implementation Details.} Similar to variant - 1 (Fig. \ref{fig:baseline_variant_1_combined}) we train the variant -2 model in parallel.  We chose hyperparameters $\lambda_1, \lambda_2, \lambda_3, \lambda_4 \text{ as } \{1, 1, 2, 2\} \text{, respectively}$

\newpage

\subsection{SceneSayerODE/SDE}

\begin{algorithm}[!h]
\SetCustomAlgoRuledWidth{\textwidth}
\DontPrintSemicolon
\footnotesize{
\KwIn{Train Horizon: $H$, Observed frames: $\topsequence{I}{t}{1}{T}$, Pre-trained object detector: $\operatorname{ObjectDetector}$, Object Encoder: $\theta_{ob}$, Spatial Encoder: $\theta_{sp}$, Temporal Encoder: $\theta_{tp}$, and ODE Function: $\theta_{ode}$}}
\textcolor{blue}{*** Randomly initialize $ \theta_{ob}, \theta_{sp}, \theta_{tp}, \theta_{ode}$ ***} \;
\SetKwBlock{Begin}{function}{end function}{
    \For{epochs $n$ in $1, \cdots, N$}
    {
        \For{videos $v$ in $1, \cdots, V$}
        {
             \For{$T$ in $3, \cdots, N_{F}-H$}
            {   
                \textcolor{blue}{*** Fetch features, bounding boxes, classes ***} \;
                $\{ \mathbf{v}^{t}, \mathbf{b}^{t}, \mathbf{c}^{t} \}_{t=1}^{T} $ = $\operatorname{ObjectDetector}(\topsequence{I}{t}{1}{T})$ \;

                \textcolor{blue}{*** Stack object features from observed frames ***} \;
                $\{ \mathbf{V}_{i}^{(1:T)} = \operatorname{Stack}( \{\mathbf{v}_{i}^{t}\}_{t=1}^{T}) \}_{i}$ \;

                \textcolor{blue}{*** Pass through Object Encoder ***} \;
                $\{  \mathbf{V}_{i}^{(1:T)} =  \operatorname{ObjectEncoder}( \mathbf{V}_{i}^{(1:T)}) \}_{i}$ \;

                \textcolor{blue}{*** Construct relationship representations ***} \;
                $\{ \mathbf{z}_{ij}^{t} = \operatorname{Concat} (\mathbf{W}_{1} \mathbf{v}^{t}_{i}, \mathbf{W}_{2} \mathbf{v}^{t}_{j}, \mathbf{W}_{3}\mathbf{U}_{ij}^{t}, \mathbf{S}_{i}^{t}, \mathbf{S}_{j}^{t}) \}_{t=1}^{T}$ \;

                \textcolor{blue}{*** Stack all relationships in a frame ***} \;
                $\mathbf{Z}^{t} = \operatorname{Stack}(\{\mathbf{z}_{ij}^{t}\}_{ij})$  \;

                \textcolor{blue}{*** Pass through the spatial encoder ***} \;                   
                $\{ \mathbf{Z}^{t} = \operatorname{SpatialEncoder}(\mathbf{Z}^{t}) \}_{t=1}^{T}$ \;

                \textcolor{blue}{*** Stack relationships from observed frames ***} \;
                $\{ \mathbf{Z}_{ij}^{(1:T)} = \operatorname{Stack}( \{\mathbf{z}_{ij}^{t}\}_{t=1}^{T}) \}_{ij}$ \;

                \textcolor{blue}{*** Pass through temporal encoder ***} \;
                $\{ \mathbf{Z}_{ij}^{(1:T)} =  \operatorname{TemporalEncoder}(\mathbf{Z}_{ij}^{(1:T)}) \}_{ij}$ \;

                \textcolor{blue}{*** Predicate Classification Decoder ***} \;
                Pass $\mathbf{Z}_{ij}^{(1:T)}$ through decoder \;

                \textcolor{blue}{*** Solve ODE to generate future relationship representations ***} \;
                $\{ \mathbf{z}_{ij} \}_{t=T+1}^{T+H} = \operatorname{ODE\_Solve}(\operatorname{f_{\theta_{ode}}, \mathbf{z}_{ij}^{T}, H})$

                \textcolor{blue}{*** Predicate Classification Decoder ***} \;
                Pass $\{ \mathbf{z}_{ij} \}_{t=T+1}^{T+H}$ through decoder.     

                \textcolor{blue}{*** Bounding box regression head ***} \;
                Pass $\{ \mathbf{z}_{ij} \}_{t=T+1}^{T+H}$ through bounding box decoder. 
            }
        }
    }

}
\Return{An assignment to $\theta_{sp}, \theta_{ob}, \theta_{tp}, \theta_{ode}$}
\caption{Algorithm for training SceneSayerODE}
\label{alg:scene_sayer_ode}
\end{algorithm}

\newpage

\begin{algorithm}[!h]
\SetCustomAlgoRuledWidth{\textwidth}
\DontPrintSemicolon
\footnotesize{
\KwIn{Train Horizon: $H$, Observed frames: $\topsequence{I}{t}{1}{T}$, Pre-trained object detector: $\operatorname{ObjectDetector}$, Object Encoder: $\theta_{ob}$, Spatial Encoder: $\theta_{sp}$, Temporal Encoder: $\theta_{tp}$, and SDE Function: $\theta_{sde}, \phi_{sde}$}}
\textcolor{blue}{*** Randomly initialize $ \theta_{ob}, \theta_{sp}, \theta_{tp}, \theta_{sde}, \phi_{sde}$ ***} \;
\SetKwBlock{Begin}{function}{end function}{
    \For{epochs $n$ in $1, \cdots, N$}
    {
        \For{videos $v$ in $1, \cdots, V$}
        {
             \For{$T$ in $3, \cdots, N_{F}-H$}
            {   
               \textcolor{blue}{*** Fetch features, bounding boxes, classes ***} \;
                $\{ \mathbf{v}^{t}, \mathbf{b}^{t}, \mathbf{c}^{t} \}_{t=1}^{T} $ = $\operatorname{ObjectDetector}(\topsequence{I}{t}{1}{T})$ \;

                \textcolor{blue}{*** Stack object features from observed frames ***} \;
                $\{ \mathbf{V}_{i}^{(1:T)} = \operatorname{Stack}( \{\mathbf{v}_{i}^{t}\}_{t=1}^{T}) \}_{i}$ \;

                \textcolor{blue}{*** Pass through Object Encoder ***} \;
                $\{  \mathbf{V}_{i}^{(1:T)} =  \operatorname{ObjectEncoder}( \mathbf{V}_{i}^{(1:T)}) \}_{i}$ \;
                
                \textcolor{blue}{*** Construct relationship representations ***} \;
                $\{ \mathbf{z}_{ij}^{t} = \operatorname{Concat} (\mathbf{W}_{1} \mathbf{v}^{t}_{i}, \mathbf{W}_{2} \mathbf{v}^{t}_{j}, \mathbf{W}_{3}\mathbf{U}_{ij}^{t}, \mathbf{S}_{i}^{t}, \mathbf{S}_{j}^{t}) \}_{t=1}^{T}$ \;

                \textcolor{blue}{*** Stack all relationships in a frame ***} \;
                $\mathbf{Z}^{t} = \operatorname{Stack}(\{\mathbf{z}_{ij}^{t}\}_{ij})$  \;

                \textcolor{blue}{*** Pass through the spatial encoder ***} \;                   
                $\{ \mathbf{Z}^{t} = \operatorname{SpatialEncoder}(\mathbf{Z}^{t}) \}_{t=1}^{T}$ \;

                \textcolor{blue}{*** Stack relationships from observed frames ***} \;
                $\{ \mathbf{Z}_{ij}^{(1:T)} = \operatorname{Stack}( \{\mathbf{z}_{ij}^{t}\}_{t=1}^{T}) \}_{ij}$ \;

                \textcolor{blue}{*** Pass through temporal encoder ***} \;
                $\{ \mathbf{Z}_{ij}^{(1:T)} =  \operatorname{TemporalEncoder}(\mathbf{Z}_{ij}^{(1:T)}) \}_{ij}$ \;

                \textcolor{blue}{*** Predicate Classification Decoder ***} \;
                Pass $\mathbf{Z}_{ij}^{(1:T)}$ through decoder \;

                \textcolor{blue}{*** Solve SDE to generate future relationship representations ***} \;
                $\{ \mathbf{z}_{ij} \}_{t=T+1}^{T+H} = \operatorname{SDE\_Solve}(\operatorname{\mu_{\theta_{sde}}, \sigma_{\phi_{sde}}, \mathbf{z}_{ij}^{T}, H})$

                \textcolor{blue}{*** Predicate Classification Decoder ***} \;
                Pass $\{ \mathbf{z}_{ij} \}_{t=T+1}^{T+H}$ through decoder.     

                \textcolor{blue}{*** Bounding box regression head ***} \;
                Pass $\{ \mathbf{z}_{ij} \}_{t=T+1}^{T+H}$ through bounding box decoder. 
            }
        }
    }

}
\Return{An assignment to $\theta_{sp}, \theta_{ob}, \theta_{tp}, \theta_{sde}, \phi_{sde}$}
\caption{Algorithm for training SceneSayerSDE}
\label{alg:scene_sayer_sde}
\end{algorithm}

\newpage

\subsubsection{Loss Function.} The total objective can be written as follows:


\begin{equation}
    \mathcal{L} = 
    \underbrace{
        \sum_{t=1}^{\bar{T}} \left(\lambda_{1} \mathcal{L}_{\text{gen}}^{t} + 
        \lambda_{2} \sum_{i} \mathcal{L}_{i}^{t} \right)
    }_\text{Loss Over Observed Representations} 
    + 
    \underbrace{
        \sum_{T=3}^{\bar{T}-1} \left( 
            \lambda_{3} \mathcal{L}_{\text{ant}}^{(1:T)} +
            \lambda_{4} \mathcal{L}_{\text{boxes}}^{(1:T)} + 
            \lambda_{5} \mathcal{L}_{\text{recon}}^{(1:T)}
        \right)
    }_\text{Loss Over Anticipated Representations}
\end{equation}

\subsubsection{Implementation Details.} Similar to variant - 1 (Fig. \ref{fig:baseline_variant_1_combined}) we train the SceneSayerODE model in parallel.  We chose hyperparameters $\lambda_1, \lambda_2, \lambda_3, \lambda_4, \lambda_5 \text{ as } \{1, 1, 2, 2, 2\} \text{, respectively}$

\newpage
\section{Ablation Overview}

\subsection{Dataset} We categorize objects that violate continuity in a video segment into three groups\footnote{(a) Disappearing Obj: Present in the final observed frame but absent in at least 5\% of subsequent frames; (b) Appearing Objects: Not visible in observed frames but emerge in future frames; (c) 
 Re-appearing Objects: Absent in the last observed frame but seen in earlier frames and re-emerge in future frames.}, presenting their statistics for different context lengths in Figures \ref{fig:TrainDataset} and \ref{fig:TestDataset}. \textcolor{OliveGreen}{\textbf{Even with 30\% context of the video, we notice that 60\% of the objects obey continuity, and the number increases as the context expands.}}

\begin{minipage}[t]{0.45\linewidth}

    \begin{center}
        \includegraphics[width=\linewidth]{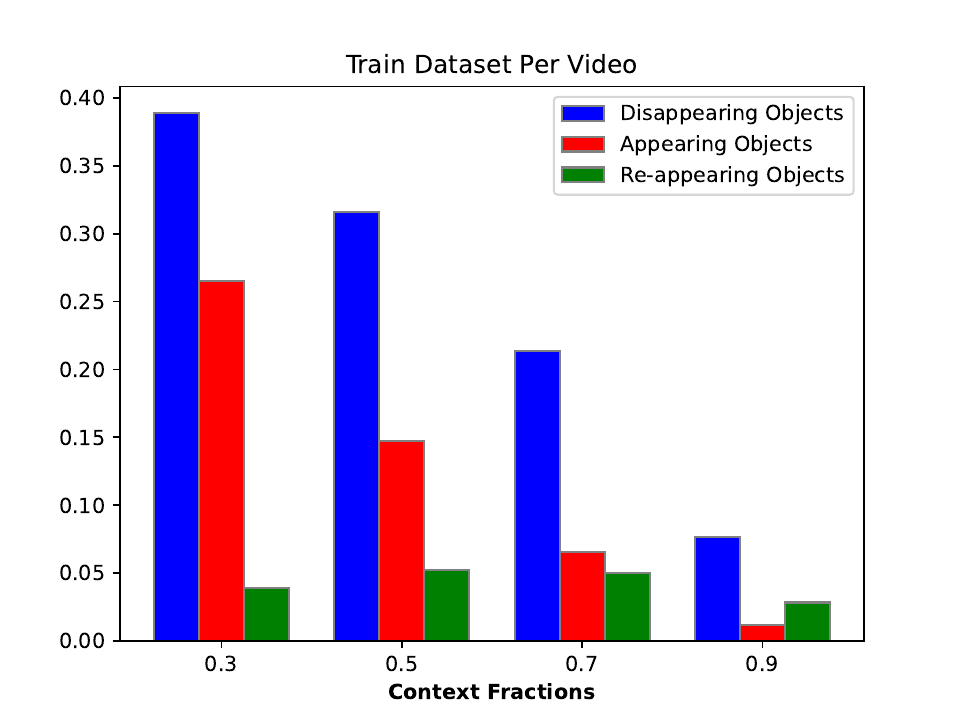}
    \end{center}    
    \label{fig:TrainDataset}
\end{minipage}
\hfill 
\begin{minipage}[t]{0.45\linewidth}

    \begin{center}
        \includegraphics[width=\linewidth]{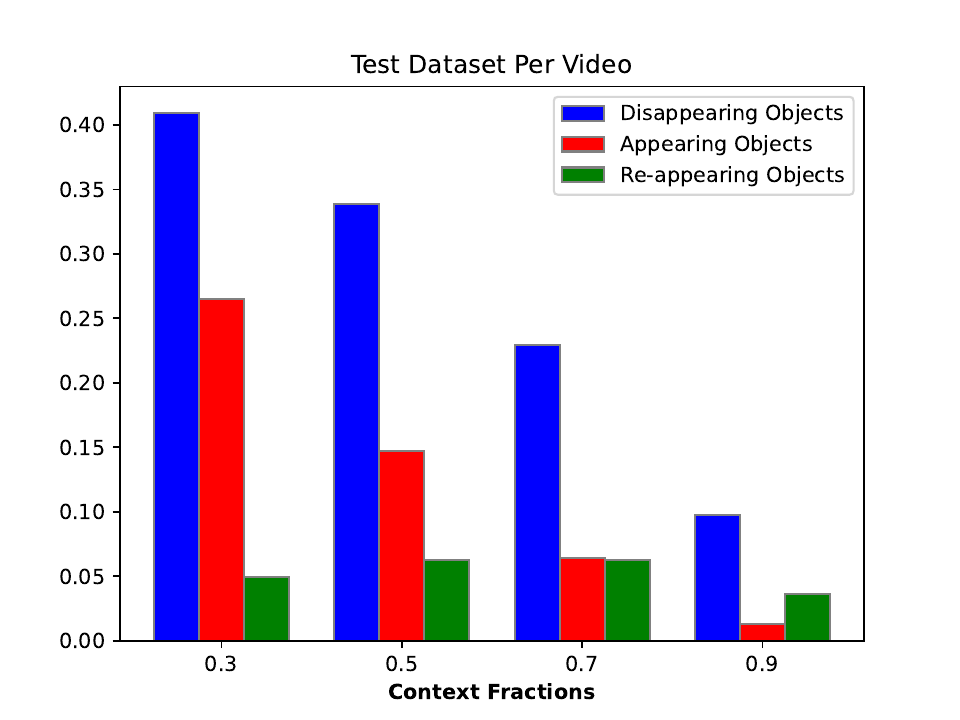}
    \end{center}    
    \label{fig:TestDataset}
\end{minipage}

\begin{table}[!t]
    \centering
    \caption{GAGS evaluation on (1) Entire test dataset \textbf{A} (numbers reported in paper), (2) Pre-processed test dataset \textbf{B}.}
    \renewcommand{\arraystretch}{1} 
    \resizebox{0.7\linewidth}{!}{
    \begin{tabular}{c|l|ccc|ccc}
    \hline
    \multicolumn{1}{l|}{} & \multicolumn{1}{l|}{\textbf{No Constraint}} & \multicolumn{3}{c|}{A (Original Dataset)} & \multicolumn{3}{c}{B (Pre-processed Dataset)}  \\ \toprule
    Context              & Method               & \textbf{R@10}         & \textbf{R@20}        & \textbf{R@50}        & \textbf{R@10}          & \textbf{R@20}          & \textbf{R@50}  \\ \midrule
    \multirow{6}{*}{0.3} & STTran+              & 30.6         & 47.3        & 62.8        & 46.3          & 71.3          & 92.5 \\
                         & DSGDetr+             & 30.5         & 45.1        & 62.8        & 46.4          & 68.0            & 92.7 \\
                         & STTran++             & 35.9         & 51.7        & 64.1        & 53.9          & 77.1          & 94.3 \\
                         & DSGDetr++            & 36.1         & 50.7        & 64.0           & 54.2          & 75.7          & 94.2 \\
                         & SceneSayerODE        & 40.5         & 54.1        & 63.9        & 60.0            & 80.6          & 96.7 \\
                         & SceneSayerSDE        & \textbf{46.9} & \textbf{59.1} & \textbf{65.2} & \textbf{70.9} & \textbf{89.4} & \textbf{99.0} \\
    \hline
    \end{tabular}
    }
    
    \label{tab:ComparisonOfOriginalPreProcessing}
\end{table}


We trained and evaluated all our models on the original dataset without any pre-processing\footnote{Apart from filtering videos with fewer than 3 annotated frames.} and presented the results in the paper (part \textbf{A}, Tab.\ref{tab:ComparisonOfOriginalPreProcessing}). We note that any violation of the continuity assumption results in decreased performance metrics. Following the suggestion from \textbf{R3}, we evaluated our models on a \textcolor{OliveGreen}{pre-processed test dataset} where the objects violating continuity assumptions are ignored, and presented results in part \textbf{B}, Tab.\ref{tab:ComparisonOfOriginalPreProcessing}. We observe that the trend persists with the pre-processed dataset. We acknowledge that handling object appearance/disappearance is an important direction for future work, and we are looking at Switching Differential Equations. \textcolor{RedViolet}{We have to solve several major technical challenges to make switching differential equations practical (e.g., optimization is NP-hard), requiring substantial work.}

\subsection{Settings} The introduction of three settings aims to comprehensively evaluate the proposed baselines and novel approaches for the Scene Graph Analysis (SGA) task, emphasizing the importance of information about the scene provided to the model. 

Firstly, we introduce \textbf{Action Genome Scenes (AGS)}, where the model receives only the video frames. The primary challenge here involves object detection and classification, often hindered by issues such as blurry imagery and object occlusions. Typically, Faster R-CNN\cite{ren2016faster} is used for object detection, but its performance can be compromised under these conditions. We suggest that employing state-of-the-art object detectors like those based on Grounding Dino \cite{liu2023grounding}, Segment Anything \cite{kirillov2023segany}, and variants that are built on these models \cite{you2023ferret, yang2023track} could potentially mitigate these issues.

Secondly, we propose the \textbf{Partially Grounded Action Genome Scenes (PGAGS)} setting to address the limitations of noisy object detectors employed. Here, models are provided with bounding box information for active interacting objects within the scene. This approach aims to bypass object detection challenges, allowing for a more accurate evaluation of the models' capabilities in processing latent dynamics without the interference of detection errors.

Lastly, we propose the \textbf{Grounded Action Genome Scenes (GAGS)} setting, which builds upon PGAGS by also supplying object labels. This addition helps overcome the difficulties models face with object classification due to occlusions and transitions in object geometry in video frames. By providing both bounding boxes and labels, GAGS aims to eliminate noise from relationship representations caused by object detection and classification inaccuracies, thereby offering a clear measure of the model's performance in predicting relationship dynamics.

\subsection{Loss Functions}

To train SceneSayer models, we employ an ensemble of loss functions that include (a) Object Classification Loss, (2) Predicate Classification Loss, (3) Bounding Box Regression Loss and (4) Reconstruction Loss. To systematically investigate the contribution of each loss component, we train and evaluate SceneSayerSDE models, excluding each component of the proposed losses under the proposed settings.

\subsection{Solvers}

The proposed methods, SceneSayerODE and SceneSayerSDE, leverage numerical solutions to address the challenge of solving non-linear ordinary differential equations (ODEs) and stochastic differential equations (SDEs), respectively. This approach is taken due to the complexity and impracticality of finding analytical solutions for such equations. For ODEs, the methods include both single-step and multi-step numerical techniques to enhance solution accuracy and efficiency. In the context of SDEs, as they depend on the specific interpretation of white noise, we apply the Euler method under the Ito Interpretation and a reversible Heun method for the Stratonovich interpretation.

\subsection{Train/Evaluation Horizons}

\subsubsection{Train Horizon.} We trained models to anticipate representations of frames using anticipation horizon $H_{t} \in \{1, 3, 5\}$. Specifically, we anticipate representations for a given horizon, decode the representation, compute loss and back-propagate it. In subsequent sections, we present the results of models trained using different anticipation horizons.

\subsubsection{Evaluation Horizon.} We evaluate trained models under two scenarios:

\begin{enumerate}
    \item \textbf{Context Fraction}: The input observation period, $T$, is varied and set to a percentage of the total number of frames, $\mathcal{N}_{F}$, with values being 30\%, 50\%, 70\%, and 90\% respectively. We consider the anticipation horizon as the total remaining frames. We present results and analysis in Sec. \ref{sec:context_fraction}.
    \item \textbf{Future Frames}: Here, the evaluation focuses on specific future time frames, defined by the anticipation horizon $H_{e}$, with values ranging from 1 to 5. we evaluate the trained models under an input observation $T \in \{3, \mathcal{N}_{F}-H_{e}\}$ and present the mean of obtained results. We present results and analysis in  Sec. \ref{sec:future_frame}.
\end{enumerate}

\newpage

\section{Ablation Results - Loss Functions}
\label{sec:loss_function}

\subsubsection{Analysis.} In Tables \ref{tab:anticipation_results_sga_ags_context_3_ab_loss}, \ref{tab:anticipation_results_sga_ags_future_frame_3_ab_loss}, we observe that (a) The proposed loss functions aid in learning spatio-temporal relationship dynamics of tail classes. (b) We note a significant drop in the performance of the trained models when we ignore reconstruction loss (c) These losses also aid in enhancing the model's ability to anticipate important relationships as witnessed by metrics \textit{R@10, R@20}.

\subsection{Action Genome Scenes}

\begin{table}[!h]
    \centering
    \captionsetup{font=small}
    \caption{Results for \textbf{SGA of AGS}, when trained using anticipatory horizon of 3 future frames.}
    \label{tab:anticipation_results_sga_ags_context_3_ab_loss}
    \setlength{\tabcolsep}{5pt} 
    \renewcommand{\arraystretch}{1.2} 
    \resizebox{\textwidth}{!}{
    \begin{tabular}{ll|cccccc|cccccc}
    \hline
         & & \multicolumn{6}{c|}{\textbf{Recall (R)}} & \multicolumn{6}{c}{\textbf{Mean Recall (mR)}} \\ 
        \cmidrule(lr){3-8} \cmidrule(lr){9-14} 
         \multicolumn{2}{c|}{\textbf{\textbf{SGA of AGS}}} & \multicolumn{3}{c}{\textbf{With Constraint}} & \multicolumn{3}{c|}{\textbf{No Constraint}} & \multicolumn{3}{c}{\textbf{With Constraint}} & \multicolumn{3}{c}{\textbf{No Constraint}}\\ 
        \cmidrule(lr){1-2}\cmidrule(lr){3-5} \cmidrule(lr){6-8}\cmidrule(lr){9-11} \cmidrule(lr){12-14} 
         $\mathcal{F}$ & \textbf{Method} & \textbf{10} & \textbf{20} & \textbf{50} & \textbf{10} & \textbf{20} & \textbf{50} & \textbf{10} & \textbf{20} & \textbf{50}  & \textbf{10} & \textbf{20} & \textbf{50}   \\ \hline
        \multirow{4}{*}{0.3} & \textbf{SceneSayerSDE (w/o BB)} & 16.0 & 27.4 & 32.8 & 22.0 & 32.9 & 46.8 & 5.4 & 11.7 & 15.6 & 14.1 & 20.4 & 30.8  \\ 
        & \textbf{SceneSayerSDE (w/o Recon)} & 14.8 & 24.3 & 28.5 & 20.5 & 31.1 & 44.6 & 4.2 & 8.0 & 10.1 & 11.3 & 16.3 & 24.8  \\ 
        & \textbf{SceneSayerSDE (w/o GenLoss)} & 15.7 & 24.7 & 28.6 & 21.6 & 32.8 & 46.1 & 4.9 & 8.8 & 10.7 & 13.6 & 19.8 & 29.0  \\ 
        & \textbf{SceneSayerSDE (Ours)} & \cellcolor{highlightColor}  \textbf{25.0} & \cellcolor{highlightColor}  \textbf{31.7} & \cellcolor{highlightColor}  \textbf{34.3} & \cellcolor{highlightColor}  \textbf{25.9} & \cellcolor{highlightColor}  \textbf{35.0} & \cellcolor{highlightColor}  \textbf{47.4} & \cellcolor{highlightColor}  \textbf{11.4} & \cellcolor{highlightColor}  \textbf{15.3} & \cellcolor{highlightColor}  \textbf{16.9} & \cellcolor{highlightColor}  \textbf{15.6} & \cellcolor{highlightColor}  \textbf{23.1} & \cellcolor{highlightColor}  \textbf{37.1}  \\ \hline
        \multirow{4}{*}{0.5} & \textbf{SceneSayerSDE (w/o BB)} & 17.5 & 29.6 & 35.5 & 23.2 & 34.7 & 50.2 & 5.8 & 12.4 & 16.8 & 14.6 & 21.2 & 32.9  \\ 
        & \textbf{SceneSayerSDE (w/o Recon)} & 15.9 & 26.1 & 30.8 & 21.8 & 33.2 & 48.5 & 4.5 & 8.6 & 11.0 & 11.9 & 17.4 & 26.7  \\ 
        & \textbf{SceneSayerSDE (w/o GenLoss)} & 17.0 & 26.7 & 30.8 & 23.3 & 35.0 & 50.0 & 5.2 & 9.3 & 11.6 & 14.7 & 21.0 & 31.3  \\ 
        & \textbf{SceneSayerSDE (Ours)} & \cellcolor{highlightColor}  \textbf{27.3} & \cellcolor{highlightColor}  \textbf{34.8} & \cellcolor{highlightColor}  \textbf{37.0} & \cellcolor{highlightColor}  \textbf{28.4} & \cellcolor{highlightColor}  \textbf{38.6} & \cellcolor{highlightColor}  \textbf{51.4} & \cellcolor{highlightColor}  \textbf{12.4} & \cellcolor{highlightColor}  \textbf{16.6} & \cellcolor{highlightColor}  \textbf{18.0} & \cellcolor{highlightColor}  \textbf{16.3} & \cellcolor{highlightColor}  \textbf{25.1} & \cellcolor{highlightColor}  \textbf{39.9}  \\ \hline
        \multirow{4}{*}{0.7} & \textbf{SceneSayerSDE (w/o BB)} & 19.1 & 32.4 & 38.7 & 27.5 & 39.9 & 54.8 & 6.3 & 13.2 & 17.7 & 16.3 & 23.6 & 34.3  \\ 
        & \textbf{SceneSayerSDE (w/o Recon)} & 17.3 & 28.9 & 33.8 & 26.0 & 38.4 & 53.4 & 5.0 & 9.8 & 12.3 & 13.9 & 19.8 & 29.9  \\ 
        & \textbf{SceneSayerSDE (w/o GenLoss)} & 18.5 & 28.8 & 33.0 & 27.4 & 39.9 & 54.2 & 5.7 & 10.1 & 12.3 & 16.5 & 23.2 & 33.1  \\ 
        & \textbf{SceneSayerSDE (Ours)} & \cellcolor{highlightColor}  \textbf{31.4} & \cellcolor{highlightColor}  \textbf{38.0} & \cellcolor{highlightColor}  \textbf{40.5} & \cellcolor{highlightColor}  \textbf{33.3} & \cellcolor{highlightColor}  \textbf{44.0} & \cellcolor{highlightColor}  \textbf{56.4} & \cellcolor{highlightColor}  \textbf{13.8} & \cellcolor{highlightColor}  \textbf{17.7} & \cellcolor{highlightColor}  \textbf{19.3} & \cellcolor{highlightColor}  \textbf{18.1} & \cellcolor{highlightColor}  \textbf{27.3} & \cellcolor{highlightColor}  \textbf{44.4}  \\ \hline
        \multirow{4}{*}{0.9} & \textbf{SceneSayerSDE (w/o BB)} & 20.6 & 35.0 & 41.7 & 31.3 & 44.2 & 59.0 & 6.7 & 13.9 & 19.0 & 18.6 & 25.7 & 37.2  \\ 
        & \textbf{SceneSayerSDE (w/o Recon)} & 19.0 & 31.1 & 36.1 & 28.5 & 42.6 & 57.8 & 5.9 & 11.1 & 13.8 & 15.4 & 22.1 & 32.1  \\ 
        & \textbf{SceneSayerSDE (w/o GenLoss)} & 20.2 & 31.2 & 35.5 & 31.1 & 44.5 & 58.4 & 6.2 & 10.9 & 13.2 & 18.0 & 25.3 & 35.0  \\ 
        & \textbf{SceneSayerSDE (Ours)} & \cellcolor{highlightColor}  \textbf{34.8} & \cellcolor{highlightColor}  \textbf{41.9} & \cellcolor{highlightColor}  \textbf{44.1} & \cellcolor{highlightColor}  \textbf{37.3} & \cellcolor{highlightColor}  \textbf{48.6} & \cellcolor{highlightColor}  \textbf{61.6} & \cellcolor{highlightColor}  \textbf{15.1} & \cellcolor{highlightColor}  \textbf{19.4} & \cellcolor{highlightColor}  \textbf{21.0} & \cellcolor{highlightColor}  \textbf{20.8} & \cellcolor{highlightColor}  \textbf{30.9} & \cellcolor{highlightColor}  \textbf{46.8}  \\ \hline
    \end{tabular}
    }
\end{table}

\begin{table}[!h]
    \centering
    \captionsetup{font=small}
    \caption{Results for \textbf{SGA of AGS}, when trained using anticipatory horizon of 3 future frames.}
    \label{tab:anticipation_results_sga_ags_future_frame_3_ab_loss}
    \setlength{\tabcolsep}{5pt} 
    \renewcommand{\arraystretch}{1.2} 
    \resizebox{\textwidth}{!}{
    \begin{tabular}{ll|cccccc|cccccc}
    \hline
         & & \multicolumn{6}{c|}{\textbf{Recall (R)}} & \multicolumn{6}{c}{\textbf{Mean Recall (mR)}} \\ 
        \cmidrule(lr){3-8} \cmidrule(lr){9-14} 
         \multicolumn{2}{c|}{\textbf{\textbf{SGA of AGS}}} & \multicolumn{3}{c}{\textbf{With Constraint}} & \multicolumn{3}{c|}{\textbf{No Constraint}} & \multicolumn{3}{c}{\textbf{With Constraint}} & \multicolumn{3}{c}{\textbf{No Constraint}}\\ 
        \cmidrule(lr){1-2}\cmidrule(lr){3-5} \cmidrule(lr){6-8}\cmidrule(lr){9-11} \cmidrule(lr){12-14} 
         $\mathcal{F}$ & \textbf{Method} & \textbf{10} & \textbf{20} & \textbf{50} & \textbf{10} & \textbf{20} & \textbf{50} & \textbf{10} & \textbf{20} & \textbf{50}  & \textbf{10} & \textbf{20} & \textbf{50}   \\ \hline
        \multirow{4}{*}{1} & \textbf{SceneSayerSDE (w/o BB)} & 20.8 & 35.1 & 41.5 & 30.4 & 43.3 & 58.0 & 6.7 & 13.9 & 18.4 & 17.4 & 24.9 & 36.0  \\ 
        & \textbf{SceneSayerSDE (w/o Recon)} & 19.6 & 31.3 & 36.0 & 27.9 & 41.8 & 57.4 & 6.2 & 11.2 & 13.9 & 15.1 & 21.8 & 31.6  \\ 
        & \textbf{SceneSayerSDE (w/o GenLoss)} & 19.7 & 30.6 & 34.6 & 29.6 & 43.0 & 57.2 & 6.0 & 10.5 & 12.7 & 16.9 & 24.0 & 33.7  \\ 
        & \textbf{SceneSayerSDE (Ours)} & \cellcolor{highlightColor} \textbf{34.1} & \cellcolor{highlightColor} \textbf{41.1} & \cellcolor{highlightColor} \textbf{43.4} & \cellcolor{highlightColor} \textbf{36.4} & \cellcolor{highlightColor} \textbf{47.2} & \cellcolor{highlightColor} \textbf{59.9} & \cellcolor{highlightColor} \textbf{15.1} & \cellcolor{highlightColor} \textbf{19.3} & \cellcolor{highlightColor} \textbf{20.8} & \cellcolor{highlightColor} \textbf{20.2} & \cellcolor{highlightColor} \textbf{29.9} & \cellcolor{highlightColor} \textbf{45.9}  \\ \hline
        \multirow{4}{*}{2} & \textbf{SceneSayerSDE (w/o BB)} & 20.6 & 34.7 & 41.1 & 29.6 & 42.3 & 57.5 & 6.6 & 13.8 & 18.3 & 17.2 & 24.6 & 35.7  \\ 
        & \textbf{SceneSayerSDE (w/o Recon)} & 19.2 & 30.9 & 35.7 & 27.2 & 40.8 & 56.7 & 6.1 & 11.1 & 13.8 & 14.9 & 21.5 & 31.2  \\ 
        & \textbf{SceneSayerSDE (w/o GenLoss)} & 19.6 & 30.3 & 34.4 & 28.8 & 42.0 & 56.7 & 6.0 & 10.4 & 12.7 & 16.6 & 23.8 & 33.5  \\ 
        & \textbf{SceneSayerSDE (Ours)} & \cellcolor{highlightColor} \textbf{33.5} & \cellcolor{highlightColor} \textbf{40.6} & \cellcolor{highlightColor} \textbf{43.0} & \cellcolor{highlightColor} \textbf{35.3} & \cellcolor{highlightColor} \textbf{46.4} & \cellcolor{highlightColor} \textbf{59.3} & \cellcolor{highlightColor} \textbf{15.2} & \cellcolor{highlightColor} \textbf{19.2} & \cellcolor{highlightColor} \textbf{20.6} & \cellcolor{highlightColor} \textbf{19.9} & \cellcolor{highlightColor} \textbf{29.6} & \cellcolor{highlightColor} \textbf{45.5}  \\ \hline
        \multirow{4}{*}{3} & \textbf{SceneSayerSDE (w/o BB)} & 20.4 & 34.4 & 40.8 & 29.1 & 41.7 & 57.0 & 6.5 & 13.7 & 18.2 & 17.0 & 24.4 & 35.5  \\ 
        & \textbf{SceneSayerSDE (w/o Recon)} & 18.9 & 30.6 & 35.4 & 26.8 & 40.1 & 56.2 & 5.9 & 10.9 & 13.6 & 14.8 & 21.3 & 31.0  \\ 
        & \textbf{SceneSayerSDE (w/o GenLoss)} & 19.4 & 30.1 & 34.2 & 28.3 & 41.4 & 56.2 & 5.9 & 10.4 & 12.6 & 16.5 & 23.6 & 33.3  \\ 
        & \textbf{SceneSayerSDE (Ours)} & \cellcolor{highlightColor} \textbf{33.0} & \cellcolor{highlightColor} \textbf{40.2} & \cellcolor{highlightColor} \textbf{42.6} & \cellcolor{highlightColor} \textbf{34.6} & \cellcolor{highlightColor} \textbf{45.7} & \cellcolor{highlightColor} \textbf{58.7} & \cellcolor{highlightColor} \textbf{15.1} & \cellcolor{highlightColor} \textbf{19.0} & \cellcolor{highlightColor} \textbf{20.4} & \cellcolor{highlightColor} \textbf{19.6} & \cellcolor{highlightColor} \textbf{29.2} & \cellcolor{highlightColor} \textbf{45.2}  \\ \hline
        \multirow{4}{*}{4} & \textbf{SceneSayerSDE (w/o BB)} & 20.1 & 34.0 & 40.4 & 28.6 & 41.2 & 56.5 & 6.5 & 13.6 & 18.1 & 16.9 & 24.2 & 35.3  \\ 
        & \textbf{SceneSayerSDE (w/o Recon)} & 18.6 & 30.2 & 35.1 & 26.5 & 39.6 & 55.6 & 5.8 & 10.7 & 13.4 & 14.7 & 21.1 & 30.7  \\ 
        & \textbf{SceneSayerSDE (w/o GenLoss)} & 19.2 & 29.9 & 34.0 & 27.9 & 40.9 & 55.7 & 5.9 & 10.3 & 12.5 & 16.4 & 23.4 & 33.1  \\ 
        & \textbf{SceneSayerSDE (Ours)} & \cellcolor{highlightColor} \textbf{32.9} & \cellcolor{highlightColor} \textbf{40.2} & \cellcolor{highlightColor} \textbf{42.6} & \cellcolor{highlightColor} \textbf{34.2} & \cellcolor{highlightColor} \textbf{45.4} & \cellcolor{highlightColor} \textbf{58.6} & \cellcolor{highlightColor} \textbf{15.2} & \cellcolor{highlightColor} \textbf{19.1} & \cellcolor{highlightColor} \textbf{20.4} & \cellcolor{highlightColor} \textbf{19.5} & \cellcolor{highlightColor} \textbf{29.1} & \cellcolor{highlightColor} \textbf{45.2}  \\ \hline
        \multirow{4}{*}{5} & \textbf{SceneSayerSDE (w/o BB)} & 19.9 & 33.7 & 40.0 & 28.3 & 40.7 & 56.0 & 6.4 & 13.5 & 18.0 & 16.7 & 24.0 & 35.1  \\ 
        & \textbf{SceneSayerSDE (w/o Recon)} & 18.4 & 29.9 & 34.9 & 26.1 & 39.1 & 55.1 & 5.7 & 10.5 & 13.2 & 14.5 & 20.9 & 30.5  \\ 
        & \textbf{SceneSayerSDE (w/o GenLoss)} & 19.0 & 29.6 & 33.7 & 27.5 & 40.4 & 55.3 & 5.8 & 10.2 & 12.4 & 16.2 & 23.2 & 33.0  \\ 
        & \textbf{SceneSayerSDE (Ours)} & \cellcolor{highlightColor} \textbf{32.9} & \cellcolor{highlightColor} \textbf{40.2} & \cellcolor{highlightColor} \textbf{42.6} & \cellcolor{highlightColor} \textbf{34.2} & \cellcolor{highlightColor} \textbf{45.4} & \cellcolor{highlightColor} \textbf{58.6} & \cellcolor{highlightColor} \textbf{15.2} & \cellcolor{highlightColor} \textbf{19.1} & \cellcolor{highlightColor} \textbf{20.4} & \cellcolor{highlightColor} \textbf{19.5} & \cellcolor{highlightColor} \textbf{29.1} & \cellcolor{highlightColor} \textbf{45.2}  \\ \hline
    \end{tabular}
    }
\end{table}

\newpage








\newpage

\section{Ablation Results - Solvers}
\label{sec:solvers}

\subsection{ODE Multi-step vs Single-step}

In Chi et al. (2023) \cite{chi_et_al_adams_2023}, multi-step Neural Ordinary Differential Equation (ODE) solvers are emphasized for their superior performance over single-step solvers, particularly in robust representation extrapolation when dealing with noisy camera motion. This is notably applicable in the context of videos from the Action Genome Dataset. Consequently, the experiments involving \textbf{SceneSayerODE} employ the Explicit Adams solver to numerically solve the learned non-linear ordinary differential equation. Formally, we define it as follows: 

\begin{equation}
    \mathbf{Z}_{t+1} = \mathbf{Z}_{t} + h \sum_{j=0}^{k} b_{j} \mathbf{F}_{t-j}, \quad \mathbf{F}_{i} = \operatorname{f}_{\theta_{ode}}(\mathbf{Z}_{i}, t)
\end{equation}

\subsubsection{Implementation Details.} Here, $\operatorname{f}_{\theta_{ode}}(\mathbf{Z}_{i}, t)$ is the ODE function learned from the data. We set $h=\frac{1}{25}$ and chose a $4^{th}-$order Adams-Bashforth method setting $k=4$.

\subsection{SDE Ito vs Stratonovich}

Given a continuous semimartingale stochastic process $\mathbf{Y}_t$ adapted to the natural filtration of the Brownian motion $\mathbf{W}_t$ in [0, T], the Stratanovich integral is:

\begin{align}
    \int_{0}^{T} \mathbf{Y}_t \circ \mathbf{dW}_t = \lim_{|\Pi|\to 0}\sum_{k=1}^{N} \frac{(\mathbf{Y}_{t_k} + \mathbf{Y}_{t_{k-1}})}{2} (\mathbf{W}_{t_k} - \mathbf{W}_{t_{k-1}}) \\
    \nonumber
    \text{where } \Pi = \{0 = t_0 <  t_1 < ... < t_N = T\}, |\Pi| = \max_{k} (t_k - t_{k - 1})
\end{align}

The Itô integral employs the left endpoint method, contrasting with the Stratanovich integral's symmetric approach. This symmetry confers a computational advantage to Stratanovich solvers, particularly evident when constructing the corresponding backward integral. Consequently, using Itô solvers, particularly those based on the Euler-Maruyama method, is more computationally intensive than employing Stratanovich solvers, such as those based on the Reversible-Heun method \cite{kidger2021efficient}. This distinction is crucial in the context of solving Neural Stochastic Differential Equations (SDEs).

We present comparative results between the Itô and Stratanovich solvers Table \ref{tab:ablation_results_recall}. In these experiments, the step size was set at $\frac{1}{25}$ seconds for solvers under both methods. The results underscore the Stratanovich SDE solution's superior predictive accuracy and its enhanced ability to generalize across tail relationship classes. This suggests that the Stratanovich interpretation of noise is more apt for modelling evolution in latent space for the SGA task.

\begin{table}[!ht]
    \centering
    \captionsetup{font=small}
    \caption{Ablation recall results using different solvers in the \text{SGA of PGAGS} setting.}
    \label{tab:ablation_results_recall}
    \setlength{\tabcolsep}{5pt} 
    \renewcommand{\arraystretch}{1.3} 
    \resizebox{\textwidth}{!}{
    \begin{tabular}{ll|cccccc|cccccc}
    \hline
        \multicolumn{2}{c}{\textbf{SGA of PGAGS}} & \multicolumn{3}{c}{\textbf{With Constraint}} & \multicolumn{3}{c}{\textbf{No Constraint}} & \multicolumn{3}{c}{\textbf{With Constraint}} & \multicolumn{3}{c}{\textbf{No Constraint}} \\ 
        \cmidrule(lr){1-2}\cmidrule(lr){3-5} \cmidrule(lr){6-8} \cmidrule(lr){9-11} \cmidrule(lr){12-14}
         $\mathcal{F}$ & \textbf{Method} & \textbf{R@10} & \textbf{R@20} & \textbf{R@50} & \textbf{R@10} & \textbf{R@20} & \textbf{R@50} & \textbf{mR@10} & \textbf{mR@20} & \textbf{mR@50} & \textbf{mR@10} & \textbf{mR@20} & \textbf{mR@50}  \\ \hline
\multirow{2}{*}{0.3}      & Euler (Ito)      & 26.31          & 27.29          & 27.29          & 31.75          & 40.08          & 46.09    & 13.21          & 13.76          & 13.77          & 20.24          & 33.13          & 47.04        \\
                          & Stratanovich     & \textbf{28.95} & \textbf{29.94} & \textbf{29.94} & \textbf{35.28} & \textbf{42.75} & \textbf{46.83} & \textbf{16.03} & \textbf{16.80} & \textbf{16.80} & \textbf{23.63} & \textbf{33.61} & \textbf{48.81} \\ \hline
\multirow{2}{*}{0.5}      & Euler (Ito)      & 30.81          & 31.88          & 31.89          & 36.28          & 45.89          & 52.13    & 15.75          & 16.50          & 16.50          & 21.92          & 33.21          & 49.30         \\
                          & Stratanovich     & \textbf{32.24} & \textbf{33.27} & \textbf{33.27} & \textbf{38.95} & \textbf{47.47} & \textbf{52.28}  & \textbf{18.03} & \textbf{18.90} & \textbf{18.90} & \textbf{25.63} & \textbf{35.71} & \textbf{50.06}\\ \hline
\multirow{2}{*}{0.7}      & Euler (Ito)      & 36.67          & 37.49          & 37.49          & 44.58          & 54.16          & 59.77      & 18.69          & 19.19          & 19.19          & 27.08          & 38.64          & 54.81      \\
                          & Stratanovich     & \textbf{38.51} & \textbf{39.35} & \textbf{39.35} & \textbf{46.42} & \textbf{55.78} & \textbf{60.34} & \textbf{20.96} & \textbf{21.64} & \textbf{21.64} & \textbf{30.00} & \textbf{42.30} & \textbf{57.11} \\ \hline
\multirow{2}{*}{0.9}      & Euler (Ito)      & 42.92          & 43.66          & 43.66          & 53.72          & 62.47          & 66.57    & 21.66          & 22.12          & 22.12          & 32.62          & 45.47          & 60.11        \\
                          & Stratanovich     & \textbf{43.64} & \textbf{44.25} & \textbf{44.25} & \textbf{54.25} & \textbf{62.73} & \textbf{66.72}  & \textbf{23.92} & \textbf{24.42} & \textbf{24.42} & \textbf{35.10} & \textbf{47.37} & \textbf{60.94} \\ \bottomrule
    \end{tabular}
    }
\end{table}

\newpage

\section{Additional Qualitative Results}
\label{sec:qualitative_results}

\begin{figure}[!htbp]
    \begin{center}
        \includegraphics[width=\textwidth]{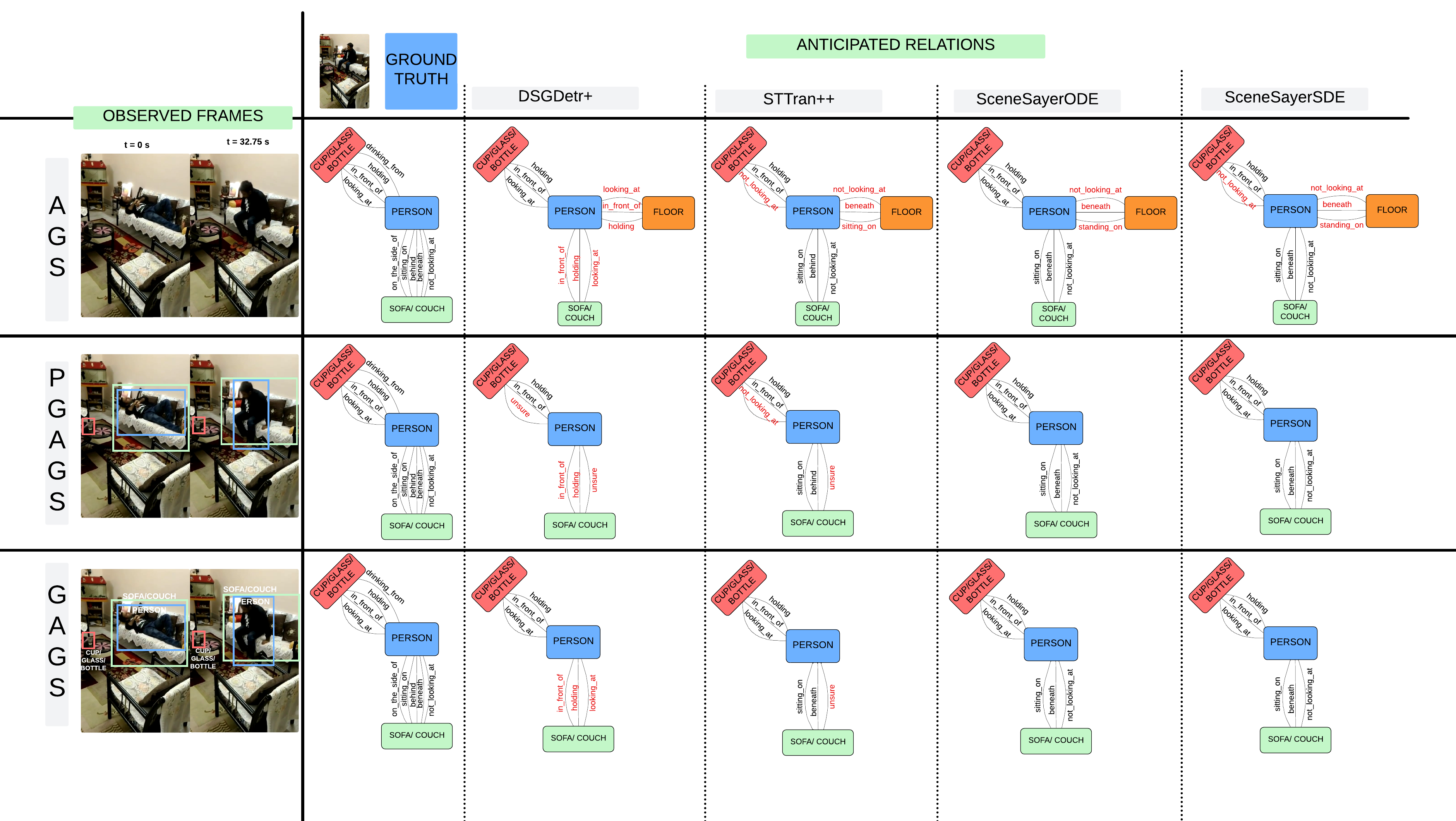}
    \end{center}
    \captionsetup{font=small}
    \caption{\textbf{Qualitative Results} To the left, we show a sampled subset of the frames observed by the models. The second column provides a ground truth scene graph corresponding to a future frame. In the subsequent columns, we contrast the performance of baseline variants with the proposed SceneSayer models. In each graph above, correct anticipations of relationships are denoted with text in black and incorrect anticipation of the relationships are highlighted with text in red.}
    \label{fig:qualitative_results}
\end{figure}

\newpage

\section{Ablation Results - Context Fraction }  \label{sec:context_fraction}

We summarise our findings from ablation performed in Sec \ref{sec:context_fraction} and \ref{sec:future_frame}.

\subsubsection{Train Horizon vs Scene Information.} Our experiments indicate that as the level of information provided to the model increases, there is a direct improvement in performance for models that are trained with a longer anticipation horizon. This suggests that more contextual information allows these models to make more accurate predictions when they are designed to anticipate further into the future (see Tables \ref{tab:anticipation_results_sga_gags_context_1}, \ref{tab:anticipation_results_sga_gags_context_3}, \ref{tab:anticipation_results_sga_gags_context_5}). However, the situation changes when we specifically look at models trained with data from Action Genome Scenes (AGS). The models that are trained to predict just one future annotated frame outperform those trained to anticipate three or five future frames (see Tables \ref{tab:anticipation_results_sga_ags_context_1}, \ref{tab:anticipation_results_sga_ags_context_3}, \ref{tab:anticipation_results_sga_ags_context_5}, \ref{tab:anticipation_results_sga_pgags_context_1}, \ref{tab:anticipation_results_sga_pgags_context_3}, \ref{tab:anticipation_results_sga_pgags_context_5}). 

\remark{We attribute this observed change to the learnability of the problem. By learnability, we refer to the capacity of the model to learn and generalize from the training data effectively. In scenarios where the models are provided with limited information, setting a longer anticipation horizon can actually impair the models' learnability. This is because when there is less information available, training the models to predict further into the future may lead them to form inaccurate internal representations, thereby hindering their ability to effectively learn the underlying patterns in the data. Thus, it suggests an optimal balance between the amount of scene information and the anticipation horizon.}

\subsubsection{Evaluation Horizon vs SceneSayer variants.} Our analysis reveals that the \textit{SceneSayerODE} variant often performs on par with, or better than, the \textit{SceneSayerSDE} variant in scenarios requiring anticipating relationships over a shorter horizon. We additionally note that this trend is more evident in settings with limited information such as \textbf{AGS}, \textbf{PGAGS} (see Tables \ref{tab:anticipation_results_sga_ags_future_frame_1}, \ref{tab:anticipation_results_sga_pgags_future_frame_1}, \ref{tab:anticipation_results_sga_gags_future_frame_1}).

\remark{The superior performance of the \textit{SceneSayerODE} variant in short-term forecasting is ascribed to its ability to model deterministic processes effectively. Since relationships are less likely to undergo significant changes over shorter horizons, the deterministic approach of ODEs aligns well with such scenarios. On the other hand, the \textit{SceneSayerSDE} variant, modelling stochastic processes, becomes more applicable as the evaluation horizon extends. The inherent variation in long-term relationships justifies the need for a stochastic modelling approach.}

\newpage

\subsection{Action Genome Scenes}

\begin{table}[!h]
    \centering
    \captionsetup{font=small}
    \caption{Results for \textbf{SGA of AGS}, when trained using anticipatory horizon of 1 future frames.}
    \label{tab:anticipation_results_sga_ags_context_1}
    \setlength{\tabcolsep}{5pt} 
    \renewcommand{\arraystretch}{1.2} 
    \resizebox{\textwidth}{!}{
\begin{tabular}{ll|cccccc|cccccc}
    \hline
         & & \multicolumn{6}{c|}{\textbf{Recall (R)}} & \multicolumn{6}{c}{\textbf{Mean Recall (mR)}} \\ 
        \cmidrule(lr){3-8} \cmidrule(lr){9-14} 
         \multicolumn{2}{c|}{\textbf{\textbf{SGA of AGS}}} & \multicolumn{3}{c}{\textbf{With Constraint}} & \multicolumn{3}{c|}{\textbf{No Constraint}} & \multicolumn{3}{c}{\textbf{With Constraint}} & \multicolumn{3}{c}{\textbf{No Constraint}}\\ 
        \cmidrule(lr){1-2}\cmidrule(lr){3-5} \cmidrule(lr){6-8}\cmidrule(lr){9-11} \cmidrule(lr){12-14} 
         $\mathcal{F}$ & \textbf{Method} & \textbf{10} & \textbf{20} & \textbf{50} & \textbf{10} & \textbf{20} & \textbf{50} & \textbf{10} & \textbf{20} & \textbf{50}  & \textbf{10} & \textbf{20} & \textbf{50}   \\ \hline
        \multirow{4}{*}{0.3} & STTran+ \cite{cong_et_al_sttran_2021} & 13.2 & 21.2 & 22.4 & 15.2 & 24.2 & 44.3 & 4.1 & 6.9 & 7.5 & 4.8 & 10.2 & 25.8  \\ 
        & DSGDetr+ \cite{Feng_2021} & 12.9 & 19.7 & 20.6 & 14.4 & 22.0 & 41.1 & 3.5 & 5.4 & 5.7 & 3.6 & 7.6 & 20.9  \\ 
        & STTran++ \cite{cong_et_al_sttran_2021} & 11.2 & 20.1 & 21.6 & 15.9 & 25.4 & 45.6 & 5.1 & 9.1 & 9.9 & 6.2 & 12.0 & 26.9  \\ 
        & DSGDetr++ \cite{Feng_2021} & 10.0 & 18.2 & 19.7 & 15.2 & 24.1 & 44.8 & 4.7 & 8.3 & 9.0 & 5.4 & 10.6 & 25.8  \\ 
        & \textbf{SceneSayerODE (Ours)} & 23.5 & 29.6 & 31.7 & 24.1 & 33.6 & 45.8 & \cellcolor{highlightColor}  \textbf{12.1} & 15.4 & 16.5 & \cellcolor{highlightColor}  \textbf{15.4} & 22.4 & 34.8  \\
\hline
    \end{tabular}
    }
\end{table}

\newpage

\begin{table}[!h]
    \centering
    \captionsetup{font=small}
    \caption{Results for \textbf{SGA of AGS}, when trained using anticipatory horizon of 3 future frames.}
    \label{tab:anticipation_results_sga_ags_context_3}
    \setlength{\tabcolsep}{5pt} 
    \renewcommand{\arraystretch}{1.2} 
    \resizebox{\textwidth}{!}{
    \begin{tabular}{ll|cccccc|cccccc}
    \hline
         & & \multicolumn{6}{c|}{\textbf{Recall (R)}} & \multicolumn{6}{c}{\textbf{Mean Recall (mR)}} \\ 
        \cmidrule(lr){3-8} \cmidrule(lr){9-14} 
         \multicolumn{2}{c|}{\textbf{\textbf{SGA of AGS}}} & \multicolumn{3}{c}{\textbf{With Constraint}} & \multicolumn{3}{c|}{\textbf{No Constraint}} & \multicolumn{3}{c}{\textbf{With Constraint}} & \multicolumn{3}{c}{\textbf{No Constraint}}\\ 
        \cmidrule(lr){1-2}\cmidrule(lr){3-5} \cmidrule(lr){6-8}\cmidrule(lr){9-11} \cmidrule(lr){12-14} 
         $\mathcal{F}$ & \textbf{Method} & \textbf{10} & \textbf{20} & \textbf{50} & \textbf{10} & \textbf{20} & \textbf{50} & \textbf{10} & \textbf{20} & \textbf{50}  & \textbf{10} & \textbf{20} & \textbf{50}   \\ \hline
        \multirow{4}{*}{0.3} & STTran+ \cite{cong_et_al_sttran_2021} & 12.5 & 19.6 & 20.7 & 13.9 & 21.6 & 40.8 & 3.4 & 5.3 & 5.7 & 3.5 & 7.3 & 20.3  \\ 
        & DSGDetr+ \cite{Feng_2021} & 12.8 & 19.6 & 20.4 & 14.3 & 21.8 & 41.3 & 3.5 & 5.3 & 5.6 & 3.6 & 7.6 & 21.2  \\ 
        & STTran++ \cite{cong_et_al_sttran_2021} & 18.5 & 27.9 & 29.5 & 15.4 & 27.2 & 48.6 & 5.9 & 10.4 & 11.3 & 6.2 & 14.1 & 31.2  \\ 
        & DSGDetr++ \cite{Feng_2021} & 19.5 & 28.3 & 29.4 & 16.8 & 29.0 & \cellcolor{highlightColor}  \textbf{48.9} & 6.0 & 10.3 & 11.0 & 8.4 & 16.7 & 32.3  \\ 
        & \textbf{SceneSayerODE (Ours)} & 23.1 & 29.2 & 31.4 & 23.3 & 32.5 & 45.1 & 10.6 & 13.8 & 15.0 & 13.3 & 20.1 & 33.0  \\
\hline
    \end{tabular}
    }
\end{table}

\newpage

\begin{table}[!h]
    \centering
    \captionsetup{font=small}
    \caption{Results for \textbf{SGA of AGS}, when trained using anticipatory horizon of 5 future frames.}
    \label{tab:anticipation_results_sga_ags_context_5}
    \setlength{\tabcolsep}{5pt} 
    \renewcommand{\arraystretch}{1.2} 
    \resizebox{\textwidth}{!}{
    \begin{tabular}{ll|cccccc|cccccc}
    \hline
         & & \multicolumn{6}{c|}{\textbf{Recall (R)}} & \multicolumn{6}{c}{\textbf{Mean Recall (mR)}} \\ 
        \cmidrule(lr){3-8} \cmidrule(lr){9-14} 
         \multicolumn{2}{c|}{\textbf{\textbf{SGA of AGS}}} & \multicolumn{3}{c}{\textbf{With Constraint}} & \multicolumn{3}{c|}{\textbf{No Constraint}} & \multicolumn{3}{c}{\textbf{With Constraint}} & \multicolumn{3}{c}{\textbf{No Constraint}}\\ 
        \cmidrule(lr){1-2}\cmidrule(lr){3-5} \cmidrule(lr){6-8}\cmidrule(lr){9-11} \cmidrule(lr){12-14} 
         $\mathcal{F}$ & \textbf{Method} & \textbf{10} & \textbf{20} & \textbf{50} & \textbf{10} & \textbf{20} & \textbf{50} & \textbf{10} & \textbf{20} & \textbf{50}  & \textbf{10} & \textbf{20} & \textbf{50}   \\ \hline
        \multirow{4}{*}{0.3} & STTran+ \cite{cong_et_al_sttran_2021} & 16.5 & 23.6 & 24.7 & 13.9 & 21.6 & 41.9 & 3.8 & 5.7 & 6.1 & 3.5 & 7.3 & 20.2  \\ 
        & DSGDetr+ \cite{Feng_2021} & 16.7 & 23.7 & 24.6 & 14.2 & 21.9 & 42.1 & 3.8 & 5.7 & 6.1 & 3.6 & 7.5 & 21.0  \\ 
        & STTran++ \cite{cong_et_al_sttran_2021} & 14.9 & 24.7 & 26.3 & 16.6 & 28.3 & \cellcolor{highlightColor}  \textbf{48.5} & 5.9 & 10.7 & 11.6 & 11.5 & 18.6 & 32.7  \\ 
        & DSGDetr++ \cite{Feng_2021} & 16.4 & 25.5 & 26.7 & 16.7 & 28.2 & 46.5 & 6.0 & 10.5 & 11.3 & 9.0 & 17.4 & 32.5  \\ 
        & \textbf{SceneSayerODE (Ours)} & \cellcolor{highlightColor}  \textbf{25.3} & 31.3 & 33.2 & 26.3 & 35.5 & 47.9 & 11.0 & 14.0 & 15.1 & 13.8 & 20.6 & 34.2  \\
\hline
    \end{tabular}
    }
\end{table}

\newpage

\subsection{Partially Grounded Action Genome Scenes}

\begin{table}[!h]
    \centering
    \captionsetup{font=small}
    \caption{Results for \textbf{SGA of PGAGS}, when trained using anticipatory horizon of 1 future frames.}
    \label{tab:anticipation_results_sga_pgags_context_1}
    \setlength{\tabcolsep}{5pt} 
    \renewcommand{\arraystretch}{1.2} 
    \resizebox{\textwidth}{!}{
    \begin{tabular}{ll|cccccc|cccccc}
    \hline
         & & \multicolumn{6}{c|}{\textbf{Recall (R)}} & \multicolumn{6}{c}{\textbf{Mean Recall (mR)}} \\ 
        \cmidrule(lr){3-8} \cmidrule(lr){9-14} 
         \multicolumn{2}{c|}{\textbf{\textbf{SGA of PGAGS}}} & \multicolumn{3}{c}{\textbf{With Constraint}} & \multicolumn{3}{c|}{\textbf{No Constraint}} & \multicolumn{3}{c}{\textbf{With Constraint}} & \multicolumn{3}{c}{\textbf{No Constraint}}\\ 
        \cmidrule(lr){1-2}\cmidrule(lr){3-5} \cmidrule(lr){6-8}\cmidrule(lr){9-11} \cmidrule(lr){12-14} 
         $\mathcal{F}$ & \textbf{Method} & \textbf{10} & \textbf{20} & \textbf{50} & \textbf{10} & \textbf{20} & \textbf{50} & \textbf{10} & \textbf{20} & \textbf{50}  & \textbf{10} & \textbf{20} & \textbf{50}   \\ \hline
        \multirow{4}{*}{0.3} & STTran+ \cite{cong_et_al_sttran_2021} & 14.9 & 15.4 & 15.4 & 24.0 & 35.2 & 45.4 & 6.3 & 6.6 & 6.6 & 11.0 & 20.6 & 41.9  \\ 
        & DSGDetr+ \cite{Feng_2021} & 17.4 & 17.8 & 17.8 & 22.8 & 34.9 & 46.8 & 4.7 & 4.9 & 4.9 & 8.2 & 17.2 & 42.0  \\ 
        & STTran++ \cite{cong_et_al_sttran_2021} & 20.7 & 21.5 & 21.5 & 25.9 & 38.3 & 45.5 & 8.0 & 8.6 & 8.6 & 12.7 & 26.3 & 42.1  \\ 
        & DSGDetr++ \cite{Feng_2021} & 20.5 & 21.3 & 21.3 & 26.3 & 37.5 & \cellcolor{highlightColor}  \textbf{47.2} & 8.0 & 8.6 & 8.6 & 12.7 & 22.1 & 43.4  \\ 
        & \textbf{SceneSayerODE (Ours)} & 25.1 & 26.1 & 26.1 & 30.5 & 39.3 & 46.3 & 13.6 & 14.1 & 14.1 & 20.3 & 29.4 & 47.7  \\
\hline
    \end{tabular}
    }
\end{table}

\newpage

\begin{table}[!h]
    \centering
    \captionsetup{font=small}
    \caption{Results for \textbf{SGA of PGAGS}, when trained using anticipatory horizon of 3 future frames.}
    \label{tab:anticipation_results_sga_pgags_context_3}
    \setlength{\tabcolsep}{5pt} 
    \renewcommand{\arraystretch}{1.2} 
    \resizebox{\textwidth}{!}{
    \begin{tabular}{ll|cccccc|cccccc}
    \hline
         & & \multicolumn{6}{c|}{\textbf{Recall (R)}} & \multicolumn{6}{c}{\textbf{Mean Recall (mR)}} \\ 
        \cmidrule(lr){3-8} \cmidrule(lr){9-14} 
         \multicolumn{2}{c|}{\textbf{\textbf{SGA of PGAGS}}} & \multicolumn{3}{c}{\textbf{With Constraint}} & \multicolumn{3}{c|}{\textbf{No Constraint}} & \multicolumn{3}{c}{\textbf{With Constraint}} & \multicolumn{3}{c}{\textbf{No Constraint}}\\ 
        \cmidrule(lr){1-2}\cmidrule(lr){3-5} \cmidrule(lr){6-8}\cmidrule(lr){9-11} \cmidrule(lr){12-14} 
         $\mathcal{F}$ & \textbf{Method} & \textbf{10} & \textbf{20} & \textbf{50} & \textbf{10} & \textbf{20} & \textbf{50} & \textbf{10} & \textbf{20} & \textbf{50}  & \textbf{10} & \textbf{20} & \textbf{50}   \\ \hline
        \multirow{4}{*}{0.3} & STTran+ \cite{cong_et_al_sttran_2021} & 22.3 & 22.9 & 22.9 & 25.5 & 38.3 & 45.6 & 8.6 & 9.1 & 9.1 & 13.1 & 24.8 & 42.3  \\ 
        & DSGDetr+ \cite{Feng_2021} & 13.6 & 14.1 & 14.1 & 22.2 & 33.9 & 46.0 & 4.8 & 5.0 & 5.0 & 7.9 & 16.9 & 40.9  \\ 
        & STTran++ \cite{cong_et_al_sttran_2021} & 22.1 & 22.8 & 22.8 & 28.1 & 39.0 & 45.2 & 9.2 & 9.8 & 9.8 & 17.7 & 30.6 & 42.0  \\ 
        & DSGDetr++ \cite{Feng_2021} & 18.2 & 18.8 & 18.8 & 27.7 & 39.2 & \cellcolor{highlightColor}  \textbf{47.3} & 8.9 & 9.4 & 9.4 & 15.3 & 26.6 & 44.0  \\ 
        & \textbf{SceneSayerODE (Ours)} & 27.0 & 27.9 & 27.9 & 33.0 & 40.9 & 46.5 & 12.9 & 13.4 & 13.4 & 19.4 & 27.9 & 46.9  \\
\hline
    \end{tabular}
    }
\end{table}

\newpage

\begin{table}[!h]
    \centering
    \captionsetup{font=small}
    \caption{Results for \textbf{SGA of PGAGS}, when trained using anticipatory horizon of 5 future frames.}
    \label{tab:anticipation_results_sga_pgags_context_5}
    \setlength{\tabcolsep}{5pt} 
    \renewcommand{\arraystretch}{1.2} 
    \resizebox{\textwidth}{!}{
    \begin{tabular}{ll|cccccc|cccccc}
    \hline
         & & \multicolumn{6}{c|}{\textbf{Recall (R)}} & \multicolumn{6}{c}{\textbf{Mean Recall (mR)}} \\ 
        \cmidrule(lr){3-8} \cmidrule(lr){9-14} 
         \multicolumn{2}{c|}{\textbf{\textbf{SGA of PGAGS}}} & \multicolumn{3}{c}{\textbf{With Constraint}} & \multicolumn{3}{c|}{\textbf{No Constraint}} & \multicolumn{3}{c}{\textbf{With Constraint}} & \multicolumn{3}{c}{\textbf{No Constraint}}\\ 
        \cmidrule(lr){1-2}\cmidrule(lr){3-5} \cmidrule(lr){6-8}\cmidrule(lr){9-11} \cmidrule(lr){12-14} 
         $\mathcal{F}$ & \textbf{Method} & \textbf{10} & \textbf{20} & \textbf{50} & \textbf{10} & \textbf{20} & \textbf{50} & \textbf{10} & \textbf{20} & \textbf{50}  & \textbf{10} & \textbf{20} & \textbf{50}   \\ \hline
        \multirow{4}{*}{0.3} & STTran+ \cite{cong_et_al_sttran_2021} & 20.2 & 20.6 & 20.6 & 21.4 & 34.8 & 44.7 & 4.9 & 5.1 & 5.1 & 7.7 & 16.4 & 40.0  \\ 
        & DSGDetr+ \cite{Feng_2021} & 13.9 & 14.3 & 14.3 & 22.6 & 35.3 & 46.8 & 4.8 & 5.0 & 5.0 & 8.1 & 17.7 & 41.8  \\ 
        & STTran++ \cite{cong_et_al_sttran_2021} & 24.5 & 25.1 & 25.1 & 28.6 & 40.2 & 45.2 & 9.6 & 10.2 & 10.2 & 17.5 & 27.2 & 42.9  \\ 
        & DSGDetr++ \cite{Feng_2021} & 23.3 & 24.0 & 24.0 & 29.3 & 40.8 & \cellcolor{highlightColor}  \textbf{47.1} & 9.1 & 9.7 & 9.7 & 17.7 & 27.4 & 44.0  \\ 
        & \textbf{SceneSayerODE (Ours)} & 27.7 & 28.7 & 28.7 & 34.1 & 42.0 & 46.8 & 12.2 & 12.8 & 12.8 & 17.4 & 27.0 & 46.3  \\
\hline
    \end{tabular}
    }
\end{table}

\newpage

\subsection{Grounded Action Genome Scenes}

\begin{table}[!h]
    \centering
    \captionsetup{font=small}
    \caption{Results for \textbf{SGA of GAGS}, when trained using anticipatory horizon of 1 future frames.}
    \label{tab:anticipation_results_sga_gags_context_1}
    \setlength{\tabcolsep}{5pt} 
    \renewcommand{\arraystretch}{1.2} 
    \resizebox{\textwidth}{!}{
    \begin{tabular}{ll|cccccc|cccccc}
    \hline
         & & \multicolumn{6}{c|}{\textbf{Recall (R)}} & \multicolumn{6}{c}{\textbf{Mean Recall (mR)}} \\ 
        \cmidrule(lr){3-8} \cmidrule(lr){9-14} 
         \multicolumn{2}{c|}{\textbf{\textbf{SGA of GAGS}}} & \multicolumn{3}{c}{\textbf{With Constraint}} & \multicolumn{3}{c|}{\textbf{No Constraint}} & \multicolumn{3}{c}{\textbf{With Constraint}} & \multicolumn{3}{c}{\textbf{No Constraint}}\\ 
        \cmidrule(lr){1-2}\cmidrule(lr){3-5} \cmidrule(lr){6-8}\cmidrule(lr){9-11} \cmidrule(lr){12-14} 
         $\mathcal{F}$ & \textbf{Method} & \textbf{10} & \textbf{20} & \textbf{50} & \textbf{10} & \textbf{20} & \textbf{50} & \textbf{10} & \textbf{20} & \textbf{50}  & \textbf{10} & \textbf{20} & \textbf{50}   \\ \hline
        \multirow{4}{*}{0.3} & STTran+ \cite{cong_et_al_sttran_2021} & 23.0 & 25.1 & 25.2 & 31.7 & 47.3 & 63.5 & 8.4 & 9.3 & 9.3 & 12.5 & 23.6 & 48.6  \\ 
        & DSGDetr+ \cite{Feng_2021} & 30.8 & 32.8 & 32.8 & 30.5 & 47.3 & 62.8 & 7.1 & 7.8 & 7.8 & 9.5 & 19.6 & 46.0  \\ 
        & STTran++ \cite{cong_et_al_sttran_2021} & 24.7 & 26.9 & 26.9 & 34.1 & 48.9 & 63.7 & 9.4 & 10.5 & 10.5 & 14.0 & 25.2 & 48.6  \\ 
        & DSGDetr++ \cite{Feng_2021} & 24.3 & 26.5 & 26.5 & 33.2 & 49.2 & 64.0 & 9.2 & 10.4 & 10.5 & 13.9 & 25.6 & 49.2  \\ 
        & \textbf{SceneSayerODE (Ours)} & 37.1 & 39.5 & 39.5 & 43.1 & 56.2 & 64.4 & \cellcolor{highlightColor}  \textbf{17.1} & 18.7 & 18.7 & \cellcolor{highlightColor}  \textbf{23.4} & \cellcolor{highlightColor}  \textbf{36.5} & 57.1  \\
\hline
    \end{tabular}
    }
\end{table}

\newpage

\begin{table}[!h]
    \centering
    \captionsetup{font=small}
    \caption{Results for \textbf{SGA of GAGS}, when trained using anticipatory horizon of 3 future frames.}
    \label{tab:anticipation_results_sga_gags_context_3}
    \setlength{\tabcolsep}{5pt} 
    \renewcommand{\arraystretch}{1.2} 
    \resizebox{\textwidth}{!}{
    \begin{tabular}{ll|cccccc|cccccc}
    \hline
         & & \multicolumn{6}{c|}{\textbf{Recall (R)}} & \multicolumn{6}{c}{\textbf{Mean Recall (mR)}} \\ 
        \cmidrule(lr){3-8} \cmidrule(lr){9-14} 
         \multicolumn{2}{c|}{\textbf{\textbf{SGA of GAGS}}} & \multicolumn{3}{c}{\textbf{With Constraint}} & \multicolumn{3}{c|}{\textbf{No Constraint}} & \multicolumn{3}{c}{\textbf{With Constraint}} & \multicolumn{3}{c}{\textbf{No Constraint}}\\ 
        \cmidrule(lr){1-2}\cmidrule(lr){3-5} \cmidrule(lr){6-8}\cmidrule(lr){9-11} \cmidrule(lr){12-14} 
         $\mathcal{F}$ & \textbf{Method} & \textbf{10} & \textbf{20} & \textbf{50} & \textbf{10} & \textbf{20} & \textbf{50} & \textbf{10} & \textbf{20} & \textbf{50}  & \textbf{10} & \textbf{20} & \textbf{50}   \\ \hline
        \multirow{4}{*}{0.3} & STTran+ \cite{cong_et_al_sttran_2021} & 30.8 & 32.8 & 32.8 & 30.6 & 47.3 & 62.8 & 7.1 & 7.8 & 7.8 & 9.5 & 19.6 & 45.7  \\ 
        & DSGDetr+ \cite{Feng_2021} & 27.0 & 28.9 & 28.9 & 30.5 & 45.1 & 62.8 & 6.7 & 7.4 & 7.4 & 9.5 & 19.4 & 45.2  \\ 
        & STTran++ \cite{cong_et_al_sttran_2021} & 30.7 & 33.1 & 33.1 & 35.9 & 51.7 & 64.1 & 11.8 & 13.3 & 13.3 & 16.5 & 29.3 & 50.2  \\ 
        & DSGDetr++ \cite{Feng_2021} & 25.7 & 28.2 & 28.2 & 36.1 & 50.7 & 64.0 & 11.1 & 12.8 & 12.8 & 19.7 & 32.0 & 51.1  \\ 
        & \textbf{SceneSayerODE (Ours)} & 34.9 & 37.3 & 37.3 & 40.5 & 54.1 & 63.9 & 15.1 & 16.6 & 16.6 & 19.6 & 31.6 & 55.8  \\
\hline
    \end{tabular}
    }
\end{table}

\newpage

\begin{table}[!h]
    \centering
    \captionsetup{font=small}
    \caption{Results for \textbf{SGA of GAGS}, when trained using anticipatory horizon of 5 future frames.}
    \label{tab:anticipation_results_sga_gags_context_5}
    \setlength{\tabcolsep}{5pt} 
    \renewcommand{\arraystretch}{1.2} 
    \resizebox{\textwidth}{!}{
    \begin{tabular}{ll|cccccc|cccccc}
    \hline
         & & \multicolumn{6}{c|}{\textbf{Recall (R)}} & \multicolumn{6}{c}{\textbf{Mean Recall (mR)}} \\ 
        \cmidrule(lr){3-8} \cmidrule(lr){9-14} 
         \multicolumn{2}{c|}{\textbf{\textbf{SGA of GAGS}}} & \multicolumn{3}{c}{\textbf{With Constraint}} & \multicolumn{3}{c|}{\textbf{No Constraint}} & \multicolumn{3}{c}{\textbf{With Constraint}} & \multicolumn{3}{c}{\textbf{No Constraint}}\\ 
        \cmidrule(lr){1-2}\cmidrule(lr){3-5} \cmidrule(lr){6-8}\cmidrule(lr){9-11} \cmidrule(lr){12-14} 
         $\mathcal{F}$ & \textbf{Method} & \textbf{10} & \textbf{20} & \textbf{50} & \textbf{10} & \textbf{20} & \textbf{50} & \textbf{10} & \textbf{20} & \textbf{50}  & \textbf{10} & \textbf{20} & \textbf{50}   \\ \hline
        \multirow{4}{*}{0.3} & STTran+ \cite{cong_et_al_sttran_2021} & 30.8 & 32.8 & 32.8 & 30.5 & 46.2 & 62.8 & 7.1 & 7.8 & 7.8 & 9.5 & 19.7 & 45.6  \\ 
        & DSGDetr+ \cite{Feng_2021} & 27.0 & 28.9 & 28.9 & 30.5 & 46.6 & 62.6 & 6.7 & 7.4 & 7.4 & 9.5 & 19.4 & 46.1  \\ 
        & STTran++ \cite{cong_et_al_sttran_2021} & 30.0 & 32.2 & 32.3 & 34.5 & 49.8 & 63.8 & 10.4 & 11.6 & 11.6 & 15.0 & 27.6 & 50.3  \\ 
        & DSGDetr++ \cite{Feng_2021} & 31.2 & 33.7 & 33.8 & 36.0 & 51.2 & 63.7 & 11.6 & 13.3 & 13.3 & 18.8 & 32.1 & 50.5  \\ 
        & \textbf{SceneSayerODE (Ours)} & 36.7 & 40.0 & 40.0 & 44.5 & 57.4 & 64.7 & 13.8 & 16.5 & 16.5 & 20.3 & 32.0 & 55.1  \\
\hline
    \end{tabular}
    }
\end{table}

\newpage

\section{Ablation Results - Future Frame }\label{sec:future_frame}

\subsection{Action Genome Scenes}

\begin{table}[!h]
    \centering
    \captionsetup{font=small}
    \caption{Results for \textbf{SGA of AGS}, when trained using anticipatory horizon of 1 future frames.}
    \label{tab:anticipation_results_sga_ags_future_frame_1}
    \setlength{\tabcolsep}{5pt} 
    \renewcommand{\arraystretch}{1.2} 
    \resizebox{\textwidth}{!}{
    \begin{tabular}{ll|cccccc|cccccc}
    \hline
         & & \multicolumn{6}{c|}{\textbf{Recall (R)}} & \multicolumn{6}{c}{\textbf{Mean Recall (mR)}} \\ 
        \cmidrule(lr){3-8} \cmidrule(lr){9-14} 
         \multicolumn{2}{c|}{\textbf{\textbf{SGA of AGS}}} & \multicolumn{3}{c}{\textbf{With Constraint}} & \multicolumn{3}{c|}{\textbf{No Constraint}} & \multicolumn{3}{c}{\textbf{With Constraint}} & \multicolumn{3}{c}{\textbf{No Constraint}}\\ 
        \cmidrule(lr){1-2}\cmidrule(lr){3-5} \cmidrule(lr){6-8}\cmidrule(lr){9-11} \cmidrule(lr){12-14} 
         $\mathcal{F}$ & \textbf{Method} & \textbf{10} & \textbf{20} & \textbf{50} & \textbf{10} & \textbf{20} & \textbf{50} & \textbf{10} & \textbf{20} & \textbf{50}  & \textbf{10} & \textbf{20} & \textbf{50}   \\ \hline
        \multirow{4}{*}{1} & STTran+ \cite{cong_et_al_sttran_2021} & 18.1 & 29.3 & 30.9 & 22.5 & 34.5 & 56.9 & 7.2 & 12.7 & 13.7 & 9.8 & 19.7 & 36.6  \\ 
        & DSGDetr+ \cite{Feng_2021} & 16.4 & 24.8 & 25.9 & 18.9 & 28.1 & 51.6 & 4.8 & 7.2 & 7.7 & 5.0 & 9.8 & 25.8  \\ 
        & STTran++ \cite{cong_et_al_sttran_2021} & 17.6 & 30.2 & 32.2 & 24.9 & 35.8 & 57.6 & 7.8 & 15.6 & 17.0 & 13.3 & 21.9 & 38.3  \\ 
        & DSGDetr++ \cite{Feng_2021} & 16.1 & 27.6 & 29.3 & 24.2 & 35.7 & 57.8 & 7.9 & 15.4 & 16.8 & 11.9 & 20.8 & 37.2  \\ 
        & \textbf{SceneSayerODE (Ours)} & \cellcolor{highlightColor} \textbf{34.9} & \cellcolor{highlightColor} \textbf{41.6} & \cellcolor{highlightColor} \textbf{43.7} & \cellcolor{highlightColor} \textbf{37.1} & \cellcolor{highlightColor} \textbf{48.1} & \cellcolor{highlightColor} \textbf{60.2} & \cellcolor{highlightColor} \textbf{17.4} & \cellcolor{highlightColor} \textbf{21.3} & \cellcolor{highlightColor} \textbf{22.6} & \cellcolor{highlightColor} \textbf{21.8} & \cellcolor{highlightColor} \textbf{31.5} & \cellcolor{highlightColor} \textbf{46.6}  \\
\hline
    \end{tabular}
    }
\end{table}

\newpage

\begin{table}[!h]
    \centering
    \captionsetup{font=small}
    \caption{Results for \textbf{SGA of AGS}, when trained using anticipatory horizon of 3 future frames.}
    \label{tab:anticipation_results_sga_ags_future_frame_3}
    \setlength{\tabcolsep}{5pt} 
    \renewcommand{\arraystretch}{1.2} 
    \resizebox{\textwidth}{!}{
    \begin{tabular}{ll|cccccc|cccccc}
    \hline
         & & \multicolumn{6}{c|}{\textbf{Recall (R)}} & \multicolumn{6}{c}{\textbf{Mean Recall (mR)}} \\ 
        \cmidrule(lr){3-8} \cmidrule(lr){9-14} 
         \multicolumn{2}{c|}{\textbf{\textbf{SGA of AGS}}} & \multicolumn{3}{c}{\textbf{With Constraint}} & \multicolumn{3}{c|}{\textbf{No Constraint}} & \multicolumn{3}{c}{\textbf{With Constraint}} & \multicolumn{3}{c}{\textbf{No Constraint}}\\ 
        \cmidrule(lr){1-2}\cmidrule(lr){3-5} \cmidrule(lr){6-8}\cmidrule(lr){9-11} \cmidrule(lr){12-14} 
         $\mathcal{F}$ & \textbf{Method} & \textbf{10} & \textbf{20} & \textbf{50} & \textbf{10} & \textbf{20} & \textbf{50} & \textbf{10} & \textbf{20} & \textbf{50}  & \textbf{10} & \textbf{20} & \textbf{50}   \\ \hline
        \multirow{4}{*}{1} & STTran+ \cite{cong_et_al_sttran_2021} & 15.5 & 24.3 & 25.6 & 17.6 & 27.0 & 50.0 & 4.6 & 7.1 & 7.6 & 4.6 & 9.4 & 24.8  \\ 
        & DSGDetr+ \cite{Feng_2021} & 16.4 & 24.6 & 25.5 & 18.4 & 27.6 & 51.3 & 4.8 & 7.1 & 7.5 & 4.8 & 9.7 & 25.7  \\ 
        & STTran++ \cite{cong_et_al_sttran_2021} & 22.5 & 33.9 & 35.6 & 20.2 & 34.6 & 58.0 & 7.3 & 13.0 & 14.0 & 9.6 & 19.6 & 37.4  \\ 
        & DSGDetr++ \cite{Feng_2021} & 23.9 & 34.8 & 36.2 & 22.6 & 36.6 & 59.6 & 7.6 & 14.0 & 15.0 & 12.2 & 22.1 & 39.4  \\ 
        & \textbf{SceneSayerODE (Ours)} & \cellcolor{highlightColor} \textbf{34.1} & 40.9 & 43.0 & \cellcolor{highlightColor} \textbf{36.4} & \cellcolor{highlightColor} \textbf{47.4} & \cellcolor{highlightColor} \textbf{60.1} & 14.0 & 17.8 & 19.2 & 17.3 & 25.9 & 42.2  \\
\hline
    \end{tabular}
    }
\end{table}

\newpage

\begin{table}[!h]
    \centering
    \captionsetup{font=small}
    \caption{Results for \textbf{SGA of AGS}, when trained using anticipatory horizon of 5 future frames.}
    \label{tab:anticipation_results_sga_ags_future_frame_5}
    \setlength{\tabcolsep}{5pt} 
    \renewcommand{\arraystretch}{1.2} 
    \resizebox{\textwidth}{!}{
    \begin{tabular}{ll|cccccc|cccccc}
    \hline
         & & \multicolumn{6}{c|}{\textbf{Recall (R)}} & \multicolumn{6}{c}{\textbf{Mean Recall (mR)}} \\ 
        \cmidrule(lr){3-8} \cmidrule(lr){9-14} 
         \multicolumn{2}{c|}{\textbf{\textbf{SGA of AGS}}} & \multicolumn{3}{c}{\textbf{With Constraint}} & \multicolumn{3}{c|}{\textbf{No Constraint}} & \multicolumn{3}{c}{\textbf{With Constraint}} & \multicolumn{3}{c}{\textbf{No Constraint}}\\ 
        \cmidrule(lr){1-2}\cmidrule(lr){3-5} \cmidrule(lr){6-8}\cmidrule(lr){9-11} \cmidrule(lr){12-14} 
         $\mathcal{F}$ & \textbf{Method} & \textbf{10} & \textbf{20} & \textbf{50} & \textbf{10} & \textbf{20} & \textbf{50} & \textbf{10} & \textbf{20} & \textbf{50}  & \textbf{10} & \textbf{20} & \textbf{50}   \\ \hline
        \multirow{4}{*}{1} & STTran+ \cite{cong_et_al_sttran_2021} & 20.1 & 28.8 & 30.1 & 17.4 & 27.1 & 51.0 & 4.5 & 7.0 & 7.5 & 4.5 & 9.3 & 24.3  \\ 
        & DSGDetr+ \cite{Feng_2021} & 21.0 & 29.2 & 30.3 & 18.3 & 28.3 & 52.5 & 4.8 & 7.1 & 7.5 & 4.7 & 9.7 & 25.7  \\ 
        & STTran++ \cite{cong_et_al_sttran_2021} & 18.3 & 29.4 & 31.0 & 20.2 & 34.3 & 58.0 & 7.4 & 12.8 & 13.8 & 11.0 & 20.7 & 37.9  \\ 
        & DSGDetr++ \cite{Feng_2021} & 19.1 & 29.5 & 30.8 & 20.2 & 33.9 & 56.6 & 7.6 & 12.9 & 13.6 & 9.4 & 19.8 & 38.6  \\ 
        & \textbf{SceneSayerODE (Ours)} & \cellcolor{highlightColor} \textbf{34.2} & \cellcolor{highlightColor} \textbf{40.7} & 42.6 & \cellcolor{highlightColor} \textbf{36.4} & \cellcolor{highlightColor} \textbf{47.5} & \cellcolor{highlightColor} \textbf{60.2} & 13.3 & 16.8 & 18.0 & 16.2 & 24.3 & 40.7  \\
\hline
    \end{tabular}
    }
\end{table}

\newpage

\subsection{Partially Grounded Action Genome Scenes}

\begin{table}[!h]
    \centering
    \captionsetup{font=small}
    \caption{Results for \textbf{SGA of PGAGS}, when trained using anticipatory horizon of 1 future frames.}
    \label{tab:anticipation_results_sga_pgags_future_frame_1}
    \setlength{\tabcolsep}{5pt} 
    \renewcommand{\arraystretch}{1.2} 
    \resizebox{\textwidth}{!}{
    \begin{tabular}{ll|cccccc|cccccc}
    \hline
         & & \multicolumn{6}{c|}{\textbf{Recall (R)}} & \multicolumn{6}{c}{\textbf{Mean Recall (mR)}} \\ 
        \cmidrule(lr){3-8} \cmidrule(lr){9-14} 
         \multicolumn{2}{c|}{\textbf{\textbf{SGA of PGAGS}}} & \multicolumn{3}{c}{\textbf{With Constraint}} & \multicolumn{3}{c|}{\textbf{No Constraint}} & \multicolumn{3}{c}{\textbf{With Constraint}} & \multicolumn{3}{c}{\textbf{No Constraint}}\\ 
        \cmidrule(lr){1-2}\cmidrule(lr){3-5} \cmidrule(lr){6-8}\cmidrule(lr){9-11} \cmidrule(lr){12-14} 
         $\mathcal{F}$ & \textbf{Method} & \textbf{10} & \textbf{20} & \textbf{50} & \textbf{10} & \textbf{20} & \textbf{50} & \textbf{10} & \textbf{20} & \textbf{50}  & \textbf{10} & \textbf{20} & \textbf{50}   \\ \hline
        \multirow{4}{*}{1} & STTran+ \cite{cong_et_al_sttran_2021} & 24.7 & 25.1 & 25.1 & 43.5 & 56.2 & 62.6 & 12.3 & 12.7 & 12.7 & 25.0 & 39.5 & 52.4  \\ 
        & DSGDetr+ \cite{Feng_2021} & 25.0 & 25.3 & 25.3 & 40.0 & 55.3 & 66.5 & 7.3 & 7.5 & 7.5 & 14.6 & 29.7 & 53.4  \\ 
        & STTran++ \cite{cong_et_al_sttran_2021} & 33.3 & 33.7 & 33.7 & 46.3 & 58.9 & 63.6 & 15.9 & 16.6 & 16.6 & 27.4 & 42.1 & 53.0  \\ 
        & DSGDetr++ \cite{Feng_2021} & 34.9 & 35.4 & 35.4 & 48.2 & 60.7 & \cellcolor{highlightColor} \textbf{66.6} & 17.1 & 17.8 & 17.8 & 29.1 & 43.2 & 58.3  \\ 
        & \textbf{SceneSayerODE (Ours)} & \cellcolor{highlightColor} \textbf{43.5} & \cellcolor{highlightColor} \textbf{44.5} & \cellcolor{highlightColor} \textbf{44.5} & \cellcolor{highlightColor} \textbf{52.5} & \cellcolor{highlightColor} \textbf{61.6} & 66.0 & 21.8 & 22.4 & 22.4 & 31.1 & 44.0 & 60.3  \\
\hline
    \end{tabular}
    }
\end{table}

\newpage

\begin{table}[!h]
    \centering
    \captionsetup{font=small}
    \caption{Results for \textbf{SGA of PGAGS}, when trained using anticipatory horizon of 3 future frames.}
    \label{tab:anticipation_results_sga_pgags_future_frame_3}
    \setlength{\tabcolsep}{5pt} 
    \renewcommand{\arraystretch}{1.2} 
    \resizebox{\textwidth}{!}{
    \begin{tabular}{ll|cccccc|cccccc}
    \hline
         & & \multicolumn{6}{c|}{\textbf{Recall (R)}} & \multicolumn{6}{c}{\textbf{Mean Recall (mR)}} \\ 
        \cmidrule(lr){3-8} \cmidrule(lr){9-14} 
         \multicolumn{2}{c|}{\textbf{\textbf{SGA of PGAGS}}} & \multicolumn{3}{c}{\textbf{With Constraint}} & \multicolumn{3}{c|}{\textbf{No Constraint}} & \multicolumn{3}{c}{\textbf{With Constraint}} & \multicolumn{3}{c}{\textbf{No Constraint}}\\ 
        \cmidrule(lr){1-2}\cmidrule(lr){3-5} \cmidrule(lr){6-8}\cmidrule(lr){9-11} \cmidrule(lr){12-14} 
         $\mathcal{F}$ & \textbf{Method} & \textbf{10} & \textbf{20} & \textbf{50} & \textbf{10} & \textbf{20} & \textbf{50} & \textbf{10} & \textbf{20} & \textbf{50}  & \textbf{10} & \textbf{20} & \textbf{50}   \\ \hline
        \multirow{4}{*}{1} & STTran+ \cite{cong_et_al_sttran_2021} & 30.7 & 31.1 & 31.1 & 42.4 & 57.6 & 63.8 & 11.5 & 11.8 & 11.8 & 22.1 & 38.9 & 55.6  \\ 
        & DSGDetr+ \cite{Feng_2021} & 21.1 & 21.3 & 21.3 & 39.9 & 55.0 & 66.5 & 7.2 & 7.3 & 7.3 & 14.6 & 29.7 & 54.0  \\ 
        & STTran++ \cite{cong_et_al_sttran_2021} & 30.4 & 30.8 & 30.8 & 44.9 & 57.3 & 63.0 & 12.3 & 12.7 & 12.7 & 26.5 & 40.1 & 54.0  \\ 
        & DSGDetr++ \cite{Feng_2021} & 27.0 & 27.4 & 27.4 & 47.0 & \cellcolor{highlightColor} \textbf{60.9} & \cellcolor{highlightColor} \textbf{66.9} & 12.9 & 13.3 & 13.3 & 27.3 & 43.0 & 56.9  \\ 
        & \textbf{SceneSayerODE (Ours)} & 41.4 & 42.3 & 42.3 & 50.6 & 60.2 & 65.5 & 18.2 & 18.7 & 18.7 & 25.9 & 38.3 & 57.5  \\
\hline
    \end{tabular}
    }
\end{table}

\newpage

\begin{table}[!h]
    \centering
    \captionsetup{font=small}
    \caption{Results for \textbf{SGA of PGAGS}, when trained using anticipatory horizon of 5 future frames.}
    \label{tab:anticipation_results_sga_pgags_future_frame_5}
    \setlength{\tabcolsep}{5pt} 
    \renewcommand{\arraystretch}{1.2} 
    \resizebox{\textwidth}{!}{
    \begin{tabular}{ll|cccccc|cccccc}
    \hline
         & & \multicolumn{6}{c|}{\textbf{Recall (R)}} & \multicolumn{6}{c}{\textbf{Mean Recall (mR)}} \\ 
        \cmidrule(lr){3-8} \cmidrule(lr){9-14} 
         \multicolumn{2}{c|}{\textbf{\textbf{SGA of PGAGS}}} & \multicolumn{3}{c}{\textbf{With Constraint}} & \multicolumn{3}{c|}{\textbf{No Constraint}} & \multicolumn{3}{c}{\textbf{With Constraint}} & \multicolumn{3}{c}{\textbf{No Constraint}}\\ 
        \cmidrule(lr){1-2}\cmidrule(lr){3-5} \cmidrule(lr){6-8}\cmidrule(lr){9-11} \cmidrule(lr){12-14} 
         $\mathcal{F}$ & \textbf{Method} & \textbf{10} & \textbf{20} & \textbf{50} & \textbf{10} & \textbf{20} & \textbf{50} & \textbf{10} & \textbf{20} & \textbf{50}  & \textbf{10} & \textbf{20} & \textbf{50}   \\ \hline
        \multirow{4}{*}{1} & STTran+ \cite{cong_et_al_sttran_2021} & 28.3 & 28.5 & 28.5 & 37.7 & 54.1 & 63.1 & 7.0 & 7.1 & 7.1 & 13.9 & 29.2 & 54.2  \\ 
        & DSGDetr+ \cite{Feng_2021} & 21.4 & 21.6 & 21.6 & 40.0 & 56.3 & \cellcolor{highlightColor} \textbf{66.7} & 7.4 & 7.6 & 7.6 & 14.7 & 30.0 & 54.4  \\ 
        & STTran++ \cite{cong_et_al_sttran_2021} & 32.3 & 32.6 & 32.6 & 43.7 & 57.0 & 62.1 & 12.5 & 12.9 & 12.9 & 24.9 & 39.9 & 55.3  \\ 
        & DSGDetr++ \cite{Feng_2021} & 32.7 & 33.0 & 33.0 & 47.1 & 60.5 & 66.6 & 12.6 & 13.0 & 13.0 & 27.5 & 41.8 & 56.1  \\ 
        & \textbf{SceneSayerODE (Ours)} & 41.6 & 42.5 & 42.5 & 50.7 & 60.6 & 65.7 & 17.3 & 17.9 & 17.9 & 24.9 & 38.8 & 58.4  \\
\hline
    \end{tabular}
    }
\end{table}

\newpage

\subsection{Grounded Action Genome Scenes}

\begin{table}[!h]
    \centering
    \captionsetup{font=small}
    \caption{Results for \textbf{SGA of GAGS}, when trained using anticipatory horizon of 1 future frames.}
    \label{tab:anticipation_results_sga_gags_future_frame_1}
    \setlength{\tabcolsep}{5pt} 
    \renewcommand{\arraystretch}{1.2} 
    \resizebox{\textwidth}{!}{
    \begin{tabular}{ll|cccccc|cccccc}
    \hline
         & & \multicolumn{6}{c|}{\textbf{Recall (R)}} & \multicolumn{6}{c}{\textbf{Mean Recall (mR)}} \\ 
        \cmidrule(lr){3-8} \cmidrule(lr){9-14} 
         \multicolumn{2}{c|}{\textbf{\textbf{SGA of GAGS}}} & \multicolumn{3}{c}{\textbf{With Constraint}} & \multicolumn{3}{c|}{\textbf{No Constraint}} & \multicolumn{3}{c}{\textbf{With Constraint}} & \multicolumn{3}{c}{\textbf{No Constraint}}\\ 
        \cmidrule(lr){1-2}\cmidrule(lr){3-5} \cmidrule(lr){6-8}\cmidrule(lr){9-11} \cmidrule(lr){12-14} 
         $\mathcal{F}$ & \textbf{Method} & \textbf{10} & \textbf{20} & \textbf{50} & \textbf{10} & \textbf{20} & \textbf{50} & \textbf{10} & \textbf{20} & \textbf{50}  & \textbf{10} & \textbf{20} & \textbf{50}   \\ \hline
        \multirow{4}{*}{1} & STTran+ \cite{cong_et_al_sttran_2021} & 37.9 & 39.0 & 39.0 & 61.9 & 81.4 & \cellcolor{highlightColor} \textbf{93.2} & 16.4 & 17.3 & 17.4 & 30.6 & 49.8 & 76.5  \\ 
        & DSGDetr+ \cite{Feng_2021} & 44.9 & 45.8 & 45.8 & 56.1 & 78.6 & 92.5 & 10.5 & 10.9 & 10.9 & 18.6 & 36.6 & 74.0  \\ 
        & STTran++ \cite{cong_et_al_sttran_2021} & 41.1 & 42.4 & 42.4 & 65.2 & 83.4 & \cellcolor{highlightColor} \textbf{93.2} & 20.1 & 21.4 & 21.4 & 32.7 & 51.8 & 77.3  \\ 
        & DSGDetr++ \cite{Feng_2021} & 41.6 & 43.0 & 43.0 & 63.9 & 83.2 & \cellcolor{highlightColor} \textbf{93.2} & 17.6 & 18.7 & 18.7 & 32.6 & 51.9 & 77.5  \\ 
        & \textbf{SceneSayerODE (Ours)} & \cellcolor{highlightColor} \textbf{60.3} & \cellcolor{highlightColor} \textbf{62.9} & \cellcolor{highlightColor} \textbf{62.9} & \cellcolor{highlightColor} \textbf{71.7} & \cellcolor{highlightColor} \textbf{86.7} & 93.1 & \cellcolor{highlightColor} \textbf{27.1} & \cellcolor{highlightColor} \textbf{28.9} & \cellcolor{highlightColor} \textbf{28.9} & \cellcolor{highlightColor} \textbf{36.5} & \cellcolor{highlightColor} \textbf{53.6} & 79.7  \\
\hline
    \end{tabular}
    }
\end{table}

\newpage

\begin{table}[!h]
    \centering
    \captionsetup{font=small}
    \caption{Results for \textbf{SGA of GAGS}, when trained using anticipatory horizon of 3 future frames.}
    \label{tab:anticipation_results_sga_gags_future_frame_3}
    \setlength{\tabcolsep}{5pt} 
    \renewcommand{\arraystretch}{1.2} 
    \resizebox{\textwidth}{!}{
    \begin{tabular}{ll|cccccc|cccccc}
    \hline
         & & \multicolumn{6}{c|}{\textbf{Recall (R)}} & \multicolumn{6}{c}{\textbf{Mean Recall (mR)}} \\ 
        \cmidrule(lr){3-8} \cmidrule(lr){9-14} 
         \multicolumn{2}{c|}{\textbf{\textbf{SGA of GAGS}}} & \multicolumn{3}{c}{\textbf{With Constraint}} & \multicolumn{3}{c|}{\textbf{No Constraint}} & \multicolumn{3}{c}{\textbf{With Constraint}} & \multicolumn{3}{c}{\textbf{No Constraint}}\\ 
        \cmidrule(lr){1-2}\cmidrule(lr){3-5} \cmidrule(lr){6-8}\cmidrule(lr){9-11} \cmidrule(lr){12-14} 
         $\mathcal{F}$ & \textbf{Method} & \textbf{10} & \textbf{20} & \textbf{50} & \textbf{10} & \textbf{20} & \textbf{50} & \textbf{10} & \textbf{20} & \textbf{50}  & \textbf{10} & \textbf{20} & \textbf{50}   \\ \hline
        \multirow{4}{*}{1} & STTran+ \cite{cong_et_al_sttran_2021} & 44.9 & 45.8 & 45.8 & 56.2 & 78.7 & 92.4 & 10.5 & 10.9 & 10.9 & 18.6 & 36.7 & 74.0  \\ 
        & DSGDetr+ \cite{Feng_2021} & 38.7 & 39.7 & 39.7 & 56.2 & 76.2 & 92.7 & 10.5 & 10.9 & 10.9 & 18.6 & 36.6 & 74.1  \\ 
        & STTran++ \cite{cong_et_al_sttran_2021} & 45.9 & 47.2 & 47.2 & 65.2 & 84.5 & \cellcolor{highlightColor} \textbf{93.4} & 20.0 & 21.3 & 21.4 & 32.4 & 51.5 & 77.5  \\ 
        & DSGDetr++ \cite{Feng_2021} & 38.7 & 40.0 & 40.0 & 64.1 & 82.5 & \cellcolor{highlightColor} \textbf{93.4} & 16.7 & 17.7 & 17.8 & 34.1 & 53.2 & 78.3  \\ 
        & \textbf{SceneSayerODE (Ours)} & 59.5 & 62.0 & 62.0 & 71.1 & 86.1 & 93.0 & 24.6 & 26.5 & 26.5 & 32.2 & 48.5 & 78.3  \\
\hline
    \end{tabular}
    }
\end{table}

\newpage

\begin{table}[!h]
    \centering
    \captionsetup{font=small}
    \caption{Results for \textbf{SGA of GAGS}, when trained using anticipatory horizon of 5 future frames.}
    \label{tab:anticipation_results_sga_gags_future_frame_5}
    \setlength{\tabcolsep}{5pt} 
    \renewcommand{\arraystretch}{1.2} 
    \resizebox{\textwidth}{!}{
    \begin{tabular}{ll|cccccc|cccccc}
    \hline
         & & \multicolumn{6}{c|}{\textbf{Recall (R)}} & \multicolumn{6}{c}{\textbf{Mean Recall (mR)}} \\ 
        \cmidrule(lr){3-8} \cmidrule(lr){9-14} 
         \multicolumn{2}{c|}{\textbf{\textbf{SGA of GAGS}}} & \multicolumn{3}{c}{\textbf{With Constraint}} & \multicolumn{3}{c|}{\textbf{No Constraint}} & \multicolumn{3}{c}{\textbf{With Constraint}} & \multicolumn{3}{c}{\textbf{No Constraint}}\\ 
        \cmidrule(lr){1-2}\cmidrule(lr){3-5} \cmidrule(lr){6-8}\cmidrule(lr){9-11} \cmidrule(lr){12-14} 
         $\mathcal{F}$ & \textbf{Method} & \textbf{10} & \textbf{20} & \textbf{50} & \textbf{10} & \textbf{20} & \textbf{50} & \textbf{10} & \textbf{20} & \textbf{50}  & \textbf{10} & \textbf{20} & \textbf{50}   \\ \hline
        \multirow{4}{*}{1} & STTran+ \cite{cong_et_al_sttran_2021} & 44.9 & 45.8 & 45.8 & 56.2 & 78.0 & 92.6 & 10.5 & 10.9 & 10.9 & 18.6 & 37.0 & 73.6  \\ 
        & DSGDetr+ \cite{Feng_2021} & 38.7 & 39.7 & 39.7 & 55.7 & 77.4 & 92.3 & 10.5 & 10.9 & 10.9 & 18.5 & 36.5 & 74.3  \\ 
        & STTran++ \cite{cong_et_al_sttran_2021} & 44.4 & 45.6 & 45.6 & 64.0 & 82.5 & \cellcolor{highlightColor} \textbf{93.3} & 17.0 & 18.0 & 18.0 & 31.3 & 51.0 & 78.4  \\ 
        & DSGDetr++ \cite{Feng_2021} & 44.0 & 45.2 & 45.2 & 63.7 & 81.8 & 93.1 & 17.0 & 18.0 & 18.0 & 32.7 & 51.5 & 78.2  \\ 
        & \textbf{SceneSayerODE (Ours)} & 56.6 & 59.8 & 59.9 & 69.5 & 85.4 & 92.8 & 20.1 & 22.9 & 22.9 & 28.3 & 45.4 & 74.3  \\
\hline
    \end{tabular}
    }
\end{table}

\newpage



\bibliographystyle{splncs04}
\bibliography{main}

\end{document}